\documentclass[10pt]{article} 

\usepackage[accepted]{rlj}           

\usepackage{amssymb}            
\usepackage{mathtools}          
\usepackage{mathrsfs}           
\usepackage{graphicx}           
\usepackage{subcaption}         
\usepackage[space]{grffile}     
\usepackage{url}                
\usepackage{lipsum}             

\usepackage{comment}
\usepackage[utf8]{inputenc} 

\usepackage{booktabs}
\usepackage{amsfonts}     
\usepackage{nicefrac}
\usepackage{microtype}
\usepackage[dvipsnames]{xcolor}

\usepackage{helvet}
\usepackage{courier}
\usepackage{blkarray}
\usepackage{bbm}

\usepackage{amsthm}
\usepackage[font=small]{caption}
\usepackage[capitalize]{cleveref}
\usepackage{autonum}
\usepackage{siunitx}
\usepackage[colorinlistoftodos]{todonotes}
\usepackage{siunitx}
\usepackage{dsfont}
\usepackage[english]{babel}
\usepackage[parfill]{parskip}
\usepackage{bm}
\usepackage{multicol}
\usepackage{relsize}

\definecolor{teal}{RGB}{64, 143, 145}
\definecolor{aubergine}{RGB}{136, 36, 81}
\definecolor{burntorange}{RGB}{226,121,46}
\definecolor{darkcyan}{RGB}{60, 138, 138}
\definecolor{darkorange}{RGB}{240, 146, 54}
\definecolor{darkpurple}{RGB}{128, 23, 134}

\usepackage{pdflscape}
\usepackage{afterpage}
\usepackage{makecell}
\usepackage{array}
\usepackage{dblfloatfix}
\usepackage{tablefootnote}
\usepackage[flushleft]{threeparttable}
\usepackage{multirow}
\usepackage{wrapfig}
\graphicspath{{Figures/}}
\usepackage{adjustbox}

\usepackage{algorithm}
\usepackage{algpseudocode}
\usepackage{changepage}
\usepackage{bbm}

\newtheorem{assumption}{Assumption}

\newtheorem{theorem}{Theorem}
\newtheorem{lemma}{Lemma}

\newtheorem{proposition}{Proposition}

\DeclareMathOperator*{\argmax}{arg\,max}
\DeclareMathOperator*{\argmin}{arg\,min}
\DeclarePairedDelimiterX{\infdivx}[2]{(}{)}{%
	#1\;\delimsize\|\;#2%
}

\title{Fully Offline Reinforcement Learning}

\setrunningtitle{Fully Offline Reinforcement Learning}

\author{Mattie Fellows\textsuperscript{1,$\star$}, Clarisse Wibault\textsuperscript{1,2,$\star$},\\
\textbf{Uljad Berdica\textsuperscript{1}, Johannes Forkel\textsuperscript{1}, \\ Maike  Osborne\textsuperscript{2}, Jakob N. Foerster\textsuperscript{1}}} 

\emails{matthew.fellows@eng.ox.ac.uk, \,clarisse.wibault@eng.ox.ac.uk} 

\affiliations{
$^{1}$\textbf{Foerster Lab for AI Research (FLAIR), Department of Engineering Science, University of Oxford}\\
$^{2}$\textbf{Machine Learning Research Group, Department of Engineering Science, University of Oxford}\\
\par 
$^\star$ Equal Contribution
}

\contribution{
In \cref{sec:pil_derivation} we derive the posterior information loss (PIL), an information-theoretic metric that upper bounds true regret in offline Bayesian RL and characterises how regret decreases through the model’s information rate. Since the PIL is computable from offline data alone, it provides a principled metric for fully offline hyperparameter tuning.
    }
    {
    None
    }

\contribution{
In \cref{sec:frequentist_regret_analysis} we carry out a frequentist regret analysis of offline model-based Bayesian RL,  proving that its regret under the true data-generating distribution converges at the minimax-optimal parametric rate $\mathcal{O}\left(\nicefrac{1}{\sqrt{N}}\right)$ under standard local asymptotic normality assumptions.
    }
    {
    Research on Bayesian offline RL remains limited. To the authors' knowledge, only \citet{chen2024bayesadaptivemontecarlo} formulate offline model-based RL as a BAMDP, and no theoretical regret analysis currently exists.
    }
\contribution{
In \cref{sec:sorel}, we introduce SOReL, a fully offline RL framework that unifies policy learning, evaluation, and entirely offline hyperparameter tuning. Policy values are estimated via the Bayesian predictive median, and hyperparameters are selected using the PIL and the value estimate as principled criteria.
}
{Developing a fully offline pipeline for offline RL remains a largely under-explored open problem \citep{unifloral2025}. Off policy evaluation (OPE) methods \citep{fu2021benchmarksdeepoffpolicyevaluation} estimate policy value offline but are evaluators rather than planners, and thus do not account for offline policy improvement. Moreover, they typically require online tuning of their own hyperparameters. The closest method to SOReL is \citet{Paine20}, which likewise relies on online tuning.
}
        \contribution{
    In \cref{sec:torel} we introduce TOReL, which extends SOReL's hyperparameter tuning as a general method for fully offline tuning of model-based and model-free offline RL algorithms including state of the art methods like IQL \citep{Kostrikov21} and ReBRAC \citep{tarasov2023revisiting}.
    }
    {
    As contribution 3, but TOReL is designed for settings where offline value estimation is not required.
    }
    \contribution{In \cref{sec:experiments} we empirically confirm SOReL's ability to approximate online value and TOReL's ability to tune hyperparameters of general ORL algorithms, both entirely offline. }
     {
         SOReL's offline approximate of true online returns has an average absolute error that is less than a quarter of the absolute error of FQE (used by \citet{Paine20}) and TOReL matches or outperforms OPE-based selection methods while eliminating the online sample cost required by \citet{unifloral2025}'s online UCB tuning algorithm.
    }


\summary{ Offline RL (ORL) promises safe and sample-efficient deployment but existing methods rely on undocumented online interactions for hyperparameter tuning and lack reliable fully offline estimates of initial online performance. We introduce SOReL, a fully offline Bayesian model-based RL method that learns a posterior over dynamics, estimates policy value via predictive uncertainty, and enables complete offline hyperparameter selection. We further propose TOReL, which extends this tuning framework to arbitrary model-free and model-based ORL algorithms. We provide a regret analysis showing that Bayesian offline RL achieves the minimax-optimal parametric rate under standard regularity conditions. Together, our methods establish a practical and theoretically grounded framework for fully offline RL.
}

\begin{document}

\setlength{\abovedisplayskip}{6pt}
\setlength{\belowdisplayskip}{6pt}
\setlength{\abovedisplayshortskip}{4pt}
\setlength{\belowdisplayshortskip}{4pt}

\makeCover
\maketitle 

\begin{abstract}
 Sample efficiency remains a major barrier to the real-world deployment of Reinforcement Learning (RL), where large numbers of online interactions are often costly or unsafe. Offline RL (ORL) seeks to address this challenge by learning policies from static datasets, yet existing methods rely on undocumented online interactions for hyperparameter tuning. Moreover, current methods provide no reliable estimate of their initial online performance using offline data alone. To overcome these limitations, we introduce two complementary algorithms for fully offline RL. Firstly, SOReL (Safe Offline Reinforcement Learning) is a model-based Bayesian approach that infers a posterior over environment dynamics from offline data and learns an approximation to a Bayes-optimal policy entirely offline. By quantifying predictive uncertainty via model ensembling, SOReL enables reliable and tractable offline prediction of online policy value and fully offline hyperparameter selection. Secondly, TOReL (Tuning for Offline Reinforcement Learning) extends this principle to arbitrary model-free and non-Bayesian ORL algorithms, leveraging predictive uncertainty quantification to eliminate costly online tuning loops. We additionally provide a regret analysis of offline Bayesian RL, showing that Bayesian approaches achieve the optimal frequentist minimax regret rate, thereby offering a formal frequentist justification for offline Bayesian methods. Empirical results demonstrate that SOReL accurately predicts online performance and that, using only offline data, TOReL achieves competitive results with online hyperparameter tuning methods, advancing safe and reliable fully offline RL for real-world deployment. Our implementations are publicly available.

\end{abstract}

\section{Introduction}

\label{sec:intro}
\begin{figure}[t!] \vspace{-0.8cm}
  \centering
  \begin{subfigure}[t]{0.377\textwidth}
    \centering
   \includegraphics[width=\textwidth]{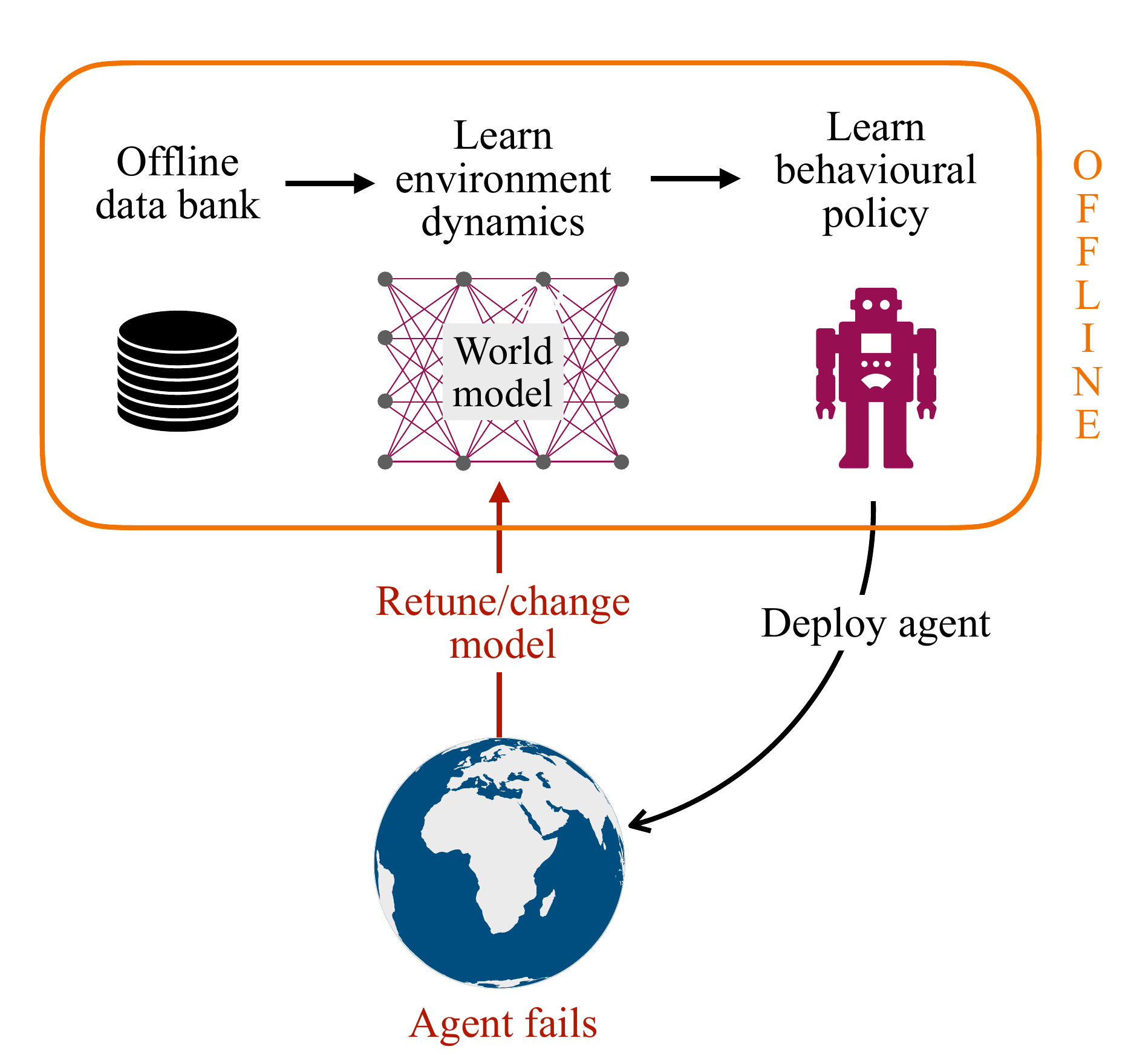}
    \vspace{-0.6cm}
    \caption{Existing Model-Based Offline RL}
    \label{fig:existing}
  \end{subfigure}
  \begin{subfigure}[t]{0.49\textwidth}
    \centering
    \includegraphics[width=\textwidth]{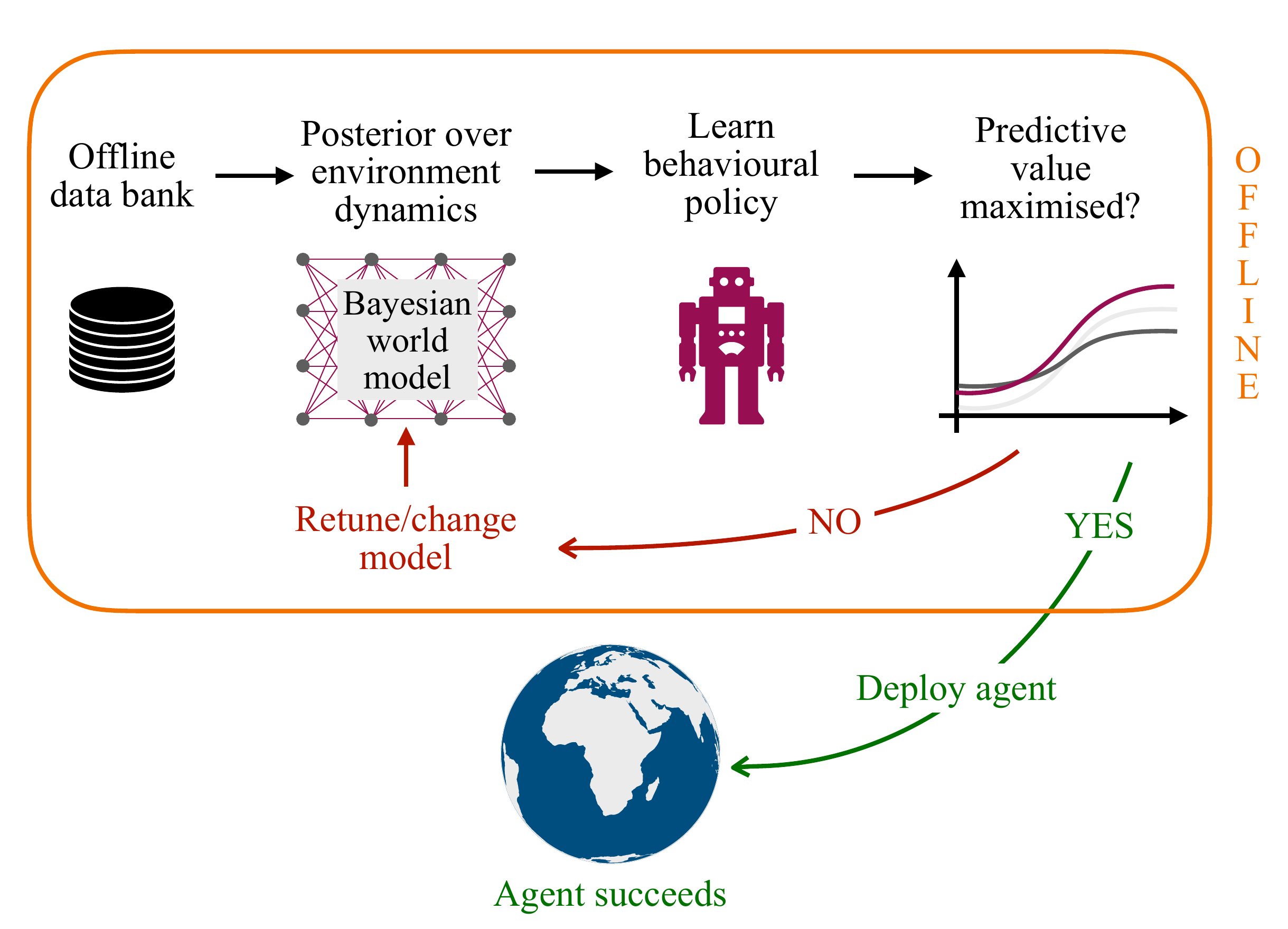}
     \vspace{-0.6cm}
    \caption{SOReL and TOReL (Our Approach)}
    \label{fig:sorel}
  \end{subfigure}
  \vspace{-0.2cm}
    \caption{Existing ORL algorithms rely on online interactions for hyperparameter tuning and validation, leading to poor online sample efficiency. In contrast, SOReL and TOReL carry out fully offline hyperparameter adjustment, using the predictive value and PIL as a tuning signal. Only then is the agent trained and deployed.}
  \label{fig:overview}
  \vspace{-0.5cm}
\end{figure}

Offline RL \citep{Lange2012,levine2020offlinereinforcementlearningtutorial,murphy2024reinforcementlearningoverview} aims to learn effective policies from static datasets, enabling agents to act autonomously and safely upon deployment. However, existing offline RL methods \citep{tarasov2023revisiting,Kostrikov21,Kidambi20,Yu21} fall short of this promise due to two largely overlooked issues. \textbf{Issue I: there is no \emph{fully offline} metric to tune all hyperparameters or select between approaches.} In practice, state-of-the-art methods rely on \emph{online} interactions to perform extensive hyperparameter tuning \citep{Zhang2021OnTI,unifloral2025}. As illustrated in \cref{fig:existing}, this leads to iterative cycles of offline training, online evaluation, hyperparameter adjustment, and retraining, incurring substantial online sample complexity — precisely what offline RL seeks to avoid.
For performance-critical problem settings, we also need reliable online performance guarantees \emph{before} the agent is deployed online. Precisely, \textbf{issue II: current methods offer no reliable \textit{fully offline} method to approximate true \textit{online} value}, i.e. the online expected returns of a policy trained using offline data. This is concerning from an AI safety perspective, as without a reliable performance estimate, we cannot deploy agents into the real world where agent failure presents a serious hazard to human life; for these settings it is essential that agents are deployed with near-optimal expected returns. For less safety-critical domains, users still require deployment-time guarantees, as suboptimal policies incur quantifiable costs in terms of reduced expected returns, such as financial losses or degraded operational performance.

We address both issues by leveraging model-based RL and exploiting predictive uncertainty from a posterior distribution over environment dynamics learned from offline data. We develop a fully offline RL pipeline (\cref{fig:sorel}) comprising two complementary algorithms. Firstly, for settings requiring accurate offline value prediction, we introduce \textbf{SOReL} (Safe Offline Reinforcement Learning), a theoretically grounded Bayesian model-based approach that resolves \textbf{Issues I \& II}.  The posterior and learned dynamics define a Bayesian RL problem whose solution, the Bayes-optimal policy, is robust to model uncertainty in environment dynamics when deployed online. For tractability, we approximate posterior sampling using ensembling and learn an amortised Bayes-optimal policy via meta-RL rollouts in posterior-sampled environments. The median predictive return across rollouts provides an offline estimate of online performance, while the predictive variance quantifies its reliability, directly addressing \textbf{Issue II}. 

To enable fully offline hyperparameter tuning, we introduce the \emph{posterior information loss} (PIL), defined as the expected KL divergence between the posterior model and the true dynamics. We show that the PIL upper-bounds regret and governs its convergence rate. Under standard regularity conditions, Bayesian offline RL achieves the optimal frequentist minimax rate, establishing a formal bridge between Bayesian uncertainty quantification and frequentist guarantees. Minimising and calibrating the PIL enables principled tuning of the environment dynamics model and inference hyperparameters, while the predictive median is used to tune planner hyperparameters, thereby resolving \textbf{Issue I}.

Empirically, we evaluate SOReL in a zero-shot setting on standard MuJoCo offline RL benchmarks \citep{Yu20,Kidambi20}. SOReL’s predictive value closely tracks realised online returns, with true performance consistently contained within the posterior predictive range once the PIL is calibrated. Compared to an OPE-based approach that still requires online tuning \citep{Paine20}, SOReL achieves less than one-quarter of the average absolute error, providing substantially more accurate value estimation, as well as calibrated uncertainty and conservative behaviour in low-data regimes. SOReL achieves this entirely without online interaction.

Secondly, for settings where exact offline value prediction is unnecessary, we introduce \textbf{TOReL} (Tuning for Offline Reinforcement Learning), which extends our fully offline tuning procedure to address \textbf{Issue I} in general non-Bayesian and model-free settings, enabling compatibility with existing methods. Many prior approaches \citep{Yu20,Kidambi20,ball21a,lu2022,sun23,tarasov2023revisiting,sims2024} rely on heuristics tailored to standard ORL benchmarks, which typically exhibit limited behavioural diversity and state-action coverage \citep{fu2020}; as a result, these methods can outperform more general Bayesian approaches like SOReL on such benchmarks. TOReL leverages the PIL and a predictive value metric correlated with true performance to tune hyperparameters entirely offline. Since both SOReL and TOReL rely only on ensembling, already standard in modern offline RL \citep{unifloral2025}, our approach introduces no additional computational overhead.

Empirically, TOReL matches or outperforms off-policy evaluation (OPE)-based tuning approaches that still rely on online calibration \citep{Paine20}. Crucially, unlike OPE methods, TOReL selects hyperparameters entirely offline using the PIL signal and predictive value metric. Compared to a state-of-the-art online bandit-based tuning method \citep{unifloral2025}, TOReL achieves comparable or superior performance while eliminating the substantial online sample cost required for hyperparameter search.

\vspace{-0.2cm}
\section{Preliminaries}\vspace{-0.1cm}
\label{sec:mat_prelim}
Let $X $ be a $\mathcal{X}\subseteq\mathbb{R}^n$-valued random variable. We denote its distribution as $P_X$ and its density (if it exists) as $p(x)$. We denote the set of all distributions over $\mathcal{X}$ as $\mathcal{P}(\mathcal{X})$. We introduce the notation $\mathcal{G}(p)$ to represent the geometric distribution and  $\mathcal{AG}(p)$ to represent the arithmetico-geometric distribution, with probability mass functions: $P_\mathcal{G}(i)\coloneqq (1-p)p^i$ and $ P_\mathcal{AG}(i)\coloneqq (1-p)^2p^i(i+1)$ respectively, for $i\in \mathbb{N}_0$ and parameter $p\in[0,1)$. We denote the uniform distribution over $\{0,1,\dots i\}$ as $\mathcal{U}_i$ and the multivariate normal distribution with mean vector $\mu$ and covariance matrix $\Sigma$ as $\mathcal{N}(\mu,\Sigma)$.
\vspace{-0.1cm}
\subsection{Offline Reinforcement Learning}\vspace{-0.1cm}
In our Reinforcement Learning (RL) setting, an agent is tasked with solving the learning problem in an infinite-horizon, discounted Markov decision process \citep{Bellman56,BELLMAN1958,Puterman94,Sutton18}: $\mathcal{M}^\star\coloneqq \langle \mathcal{S},\mathcal{A}, P_0, P_S^\star( s,a), P_R^\star( s,a), \gamma \rangle$, with state space $\mathcal{S}$, action space $\mathcal{A}$ and discount factor $\gamma$. At time $t=0$, an agent starts in an initial state allocated according the the initial state distribution: $s_0\sim P_0$. At every timestep $t$, an agent in state $s_t$ takes an action according to a policy $a_t\sim \pi(h_t)$, receives a scalar reward $r_t\sim P_R^\star(s_t,a_t)$ and transitions to a new state $s_{t+1}\sim P_S^\star(s_t,a_t)$ where $h_t\coloneqq \{s_0,a_0,r_0, s_1,a_1,r_1,\ldots a_{t-1},r_{t-1},s_t\}\in\mathcal{H}_t$ is the observed a history of interactions with the environment. Here $\mathcal{H}_t\coloneqq \mathcal{S}\times (\mathcal{A} \times \mathbb{R} \times \mathcal{S})^{t}$ denotes the corresponding product space. We assume rewards are bounded with $r_t\in [r_\textrm{min},r_\textrm{max}]\subset \mathbb{R}$ where $r_\textrm{min}$ and $r_\textrm{max}$ denote the minimum and maximum reward values respectively. For convenience, we often write the joint state transition-reward distribution as $P_{R,S}^\star(s,a)$. We denote the distribution over history $h_t$ as $P^{\star}_{t,\pi}$. The goal of an agent is to learn an optimal policy  $\pi^\star\in\Pi^\star$ where $\Pi^\star \coloneqq \argmax_{\pi} J^\pi(\mathcal{M}^\star)$ is the set of policies that maximise the expected discounted return: 
\begin{align}
    J^\pi(\mathcal{M}^\star) \coloneqq \mathbb{E}_{h_\infty \sim P^\star_{\infty,\pi}}\left[\sum_{i=0}^\infty \gamma^i r_i\right].
\end{align}
If the true transition and reward distributions $P_S^\star(s,a)$ and $P_R^\star(s,a)$ are known a priori, the problem reduces to planning, and (given sufficient compute) an optimal policy can be computed without further environment interaction; in this case, policies need only condition on the current state $s_t$. In contrast, we consider the learning setting where the dynamics are unknown. Under our Bayesian formulation, policies condition on the full history $h_t$, which carries the information required to update beliefs over the environment.

In offline RL \citep{Lange2012,levine2020offlinereinforcementlearningtutorial,murphy2024reinforcementlearningoverview}, an agent has access to a dataset of histories of various lengths collected from the true environment. The policies used to collect the data may vary and not be optimal. In this paper, we consider a zero-shot setting where the agent has no ability to interact with the target environment before deployment, as formalised by \citet{unifloral2025}. The agent is then deployed at test time $t=0$ and its performance evaluated. The goal of offline RL is to take advantage of offline data so that the deployed policy will be near-optimal from the outset. In model-based offline RL, a model of the environment dynamics is learned using the dataset. An offline policy is obtained by solving a planning problem using the learned dynamics model. In contrast, model-free offline RL methods forgo explicitly modelling environment dynamics and instead learn a policy or value function directly from the fixed dataset, without constructing an explicit dynamics model. 

\section{Related Work}
\label{sec:related}
\vspace{-0.1cm}
\subsection{Off-Policy Evaluation}
Off-policy evaluation (OPE) algorithms \citep{le2019batchpolicylearningconstraints, Nachum2019a, thomas2016dataefficientoffpolicypolicyevaluation, kostrikov2020statisticalbootstrappinguncertaintyestimation, fu2021benchmarksdeepoffpolicyevaluation} estimate the performance of a target policy using data collected by a different behaviour policy. In principle, OPE provides an offline mechanism for approximating online value. However, OPE methods are evaluators rather than planners: they estimate the value of a fixed policy without performing policy improvement \citep{Uehara2022review} and therefore do not constitute a complete offline RL pipeline. Although many OPE methods attach uncertainty measures or confidence intervals, these are typically conditional on modelling assumptions and do not provide a principled diagnostic for when insufficient data, support mismatch, or model misspecification render estimates unreliable. Moreover, OPE methods generally require calibration of their own hyperparameters using online data and methods based on importance sampling \citep{Kostrikov21} and double robustness \citep{thomas2016dataefficientoffpolicypolicyevaluation} require datasets containing trajectories rather than individual datapoints. 

OPE can also be used for offline hyperparameter selection by ranking candidate policies according to their estimated value. To our knowledge, \citet{Paine20} provide the primary study of this approach using FQE. However, their results show substantial value overestimation in several domains, provide no principled signal of estimate reliability, and require online tuning of FQE hyperparameters, so the overall pipeline is not fully offline. As noted by \citet{unifloral2025}, this framework is further limited to behavioural cloning and two critic-based methods that have since been surpassed by modern offline RL algorithms. Consequently, while OPE addresses offline value estimation, it does not fully resolve \textbf{Issues I \& II}: both policy-learning and OPE-specific hyperparameters typically require calibration beyond the offline dataset, and existing approaches lack a principled mechanism to determine whether the available data suffice for reliable performance estimation. In contrast, our method derives an information-theoretic signal (the PIL) tied to regret that governs both hyperparameter tuning and predictive value reliability entirely offline

\subsection{Model-Based Offline RL}
Developing reliable model-based offline RL remains challenging. Beyond \textbf{Issues I \& II}, performance depends critically on accurate dynamics modelling, as model errors can compound over long horizons. Additionally, benchmark datasets often lack coverage in important regions of state-action space, creating further generalisation challenges. Most existing approaches focus on mitigating this latter issue by introducing reward pessimism or uncertainty-based regularisation \citep{Yu20,Kidambi20,Kumar20,Yu21,Fujimoto21,Kostrikov21,an21,ball21a,lu2022,sun23,tarasov2023revisiting,sims2024}. Unifloral~\citep{unifloral2025} unifies these approaches within a common framework with streamlined implementations and benchmarking protocols, which we follow in our experiments.

Limited prior work directly addresses \textbf{Issues I \& II}. \citet{smith2023} propose offline hyperparameter tuning, but are limited to the model-free imitation learning setting and offer no value estimation. \citet{wang2022no} introduce offline pre-selection of hyperparameters for online methods, but do not learn optimal policies fully offline or approximate value. To our knowledge, our method is the first fully offline RL framework that integrates policy learning, value estimation (when required), and complete hyperparameter tuning using only offline data. Finally, Bayesian perspectives on offline RL remain underexplored. \citet{chen2024bayesadaptivemontecarlo} formulate offline model-based RL as solving a BAMDP, however no regret analysis of the Bayes-optimal policy is carried out, a continuous BAMDP \citep{Guez14} approximation is used to learn behavioural policies, does not propose a method for offline policy value estimation, and the algorithm still relies on online data for tuning.

\vspace{-0.2cm}

\section{Bayesian Offline RL}\label{sec:obrl}\vspace{-0.1cm}
We now introduce our Bayesian RL (BRL) framework, which constitutes learning an (approximate) posterior over environment dynamics from offline data, and then solving a BRL problem to find a Bayes-optimal policy with that posterior acting as a prior over the space of MDPs. Bayesian RL methods naturally minimise Bayes-regret, which maximises the average performance over MDPs as weighted by the posterior. We provide an introductory primer on Bayesian RL in \cref{app:brl_primer}. 
\vspace{-0.1cm}
\subsection{Learning a Posterior with Offline Data}\vspace{-0.1cm}
A Bayesian epistemology characterises the agent's uncertainty in the MDP through distributions over any unknown variable \citep{Martin67,Duff02}. For our model-based approach, we characterise uncertainty in the environment dynamics. We first specify a parametric model of the unknown joint reward-state transition distribution: $P_{R,S}(s_t,a_t, \theta)$, with each $\theta\in\Theta\subseteq \mathbb{R}^d$ representing a hypothesis about the underlying true MDP $\mathcal{M}^\star$.  A prior distribution over the parameter space $P_\Theta$ is specified, which represents the initial \emph{a priori} belief over hypothesis MDPs before the agent has observed any transitions. As we show in \cref{sec:regret_analysis}, our results can easily be generalised to non-parametric methods like Gaussian process regression \citep{Rasmussen2006Gaussian,Wiener23,Krige1951}. We denote an offline dataset of $N$ state-action-state-reward transition observations as: $\mathcal{D}_N=\{(s_i,a_i,s'_i,r_i)\}_{i=0}^{N-1}$, all collected from a single MDP $\mathcal{M}^\star$. Datapoints may be collected from several policies and non-Markovian sampling. Given the dataset $\mathcal{D}_N$, the prior $P_\Theta$ with density $p(\theta)$ is updated to posterior $P_\Theta(\mathcal{D}_N)$ with density $p(\theta | \mathcal{D}_N)$, using Bayes' rule:
\begin{align}
    p(\theta\vert \mathcal{D}_N)=\frac{p(\mathcal{D}_N\vert \theta)p(\theta)}{p(\mathcal{D}_N)}=\frac{\prod_{i=0}^{N-1}p(r_i,s_i'\vert s_i,a_i,\theta) p(\theta)}{\int_{\Theta} \prod_{i=0}^{N-1}p(r_i,s_{i}'\vert s_i,a_i,\theta)p(\theta)d\theta}.\label{eq:Bayes_rule}
\end{align}

The posterior represents the agent's updated belief in the unknown environment dynamics once $\mathcal{D}_N$ has been observed. We now detail how a Bayes-optimal policy is learned using the posterior as the initial belief in the environment dynamics.
\vspace{-0.1cm}
\subsection{Approximating a Bayes-Optimal Policy}\vspace{-0.1cm}

Solving a BRL problem exactly is intractable for all but the simplest models \citep{Martin67,Duff02,Guez12,Guez13,Zintgraf19,Fellows24}, and exact posterior inference in \cref{eq:Bayes_rule} is typically infeasible for dynamics models of interest (e.g., nonlinear Gaussian world models). We therefore learn an approximate posterior $\hat{P}_\Theta(\mathcal{D}_N)\approx P_\Theta(\mathcal{D}_N)$ using randomised prior (RP) ensembling \citep{Osband17a,Osband18,Ciosek2020} (details in \cref{app:world_model_implementation}). In the ORL setting, the prior $P_\Theta$ is replaced with the informative posterior $P_\Theta(\mathcal{D}_N)$, reducing the hypothesis space of the BRL problem. A Bayes-optimal policy is then learned using the posterior as the initial belief over environment dynamics, yielding the offline Bayesian RL objective:
\begin{align}
    J_\textrm{Bayes}^\pi(P_\Theta(\mathcal{D}_N))\coloneqq\mathbb{E}_{\theta\sim P_\Theta(\mathcal{D}_N)}\left[
    \mathbb{E}_{h_\infty \sim P_{\infty}^\pi(\theta)}\left[ \sum_{i=0}^\infty \gamma^{i} r_{i} \right]\right].\label{eq:obrl_objective}
\end{align}
Solving \cref{eq:obrl_objective} for $\pi$ to find a Bayes-optimal policy $\pi^\star_\textrm{Bayes}(h_t)\in\argmax_{\pi}J_\textrm{Bayes}^\pi(P_\Theta(\mathcal{D}_N))$ can be formulated as a meta-RL problem \citep{Zintgraf19,beck2024}, enabling an RL$^2$-style procedure \citep{Duan16}: environments are sampled from the posterior and the policy is trained via rollouts in these sampled models. An approximate Bayes-optimal policy is thus obtained by searching over history-conditioned policies, for example using recurrent PPO \citep{schulman2017ppo}. Crucially, posterior reasoning is amortised into the policy during training; deployment requires only a single forward pass of the recurrent network, incurring no additional overhead relative to standard model-based offline RL. Implementation details are provided in \cref{app:implementation}.

Beyond enabling exploration when deployed, the Bayesian formulation provides explicit epistemic uncertainty in the learned dynamics model, which underpins our fully offline pipeline. To address \textbf{Issue II}, we estimate policy value using posterior predictive rollouts: the median return across sampled models yields a Bayesian estimate of test-time performance, while the predictive variance quantifies its uncertainty. To address \textbf{Issue I}, this predictive estimate enables fully offline tuning of planner hyperparameters. In addition, as shown in \cref{sec:pil_derivation}, the Posterior Information Loss (PIL) links posterior uncertainty to regret and is used to tune model and inference hyperparameters. Together, the predictive variance and PIL provide principled diagnostics of model misspecification, data coverage, and uncertainty calibration prior to deployment.

Finally, we remark that a Bayesian approach is relatively simple and general compared to existing model-based approaches in \cref{sec:related} as it does not rely on hand-crafted heuristics tailored to specific problem settings. We emphasize that adopting a Bayesian perspective does not introduce additional computational overhead beyond what is already standard in modern model-based offline RL. We rely only on ensembling for approximate posterior sampling — already standard in modern model-based offline RL \citep{unifloral2025}. No expensive posterior inference or additional planning modules are required; the predictive median, variance, and PIL are computed from the same ensemble rollouts used for policy learning, effectively extracting additional information from computations that are already required. Hence our method does not introduce new costly components, but instead leverages existing tools more efficiently to obtain principled tuning signals and reliable offline value estimates.
\vspace{-0.5cm}
\section{Regret Analysis}\vspace{-0.1cm}
\label{sec:regret_analysis}

In this section, we provide a formal theoretical regret analysis of Bayesian offline RL. Our first result in \cref{sec:pil_derivation} proves that regret is tied to the posterior information loss (PIL), which can be used as a tuning metric and diagnostic tool during offline RL training. In \cref{sec:frequentist_regret_analysis}, we then use this result to prove that the Bayes-optimal policy convergences to zero regret at the minimax optimal rate under standard  assumptions, thereby providing a formal frequentist justification for Bayesian offline RL in general.  Proofs for all theorems can be found in \cref{app:proofs}.

\vspace{-0.1cm}
\subsection{Derivation of Posterior Information Loss (PIL)}
\label{sec:pil_derivation}

Given a dataset of datapoints $N$ collected from the true MDP $\mathcal{M}^\star$, we can measure how far the performance of the Bayes-optimal policy is from the true optimal policy using the \emph{true regret}, which, in this case, is the difference between the expected return $J^{\pi^\star}(\mathcal{M}^\star)$ of an optimal policy for the true MDP and the expected return of the Bayes-optimal policy $J^{\pi_\textrm{Bayes}^\star}(\mathcal{M}^\star,\mathcal{D}_N)$  given a posterior $P_\Theta(\mathcal{D}_N)$, all in the true MDP $\mathcal{M}^\star$: 
\begin{align}
    \textnormal{\textrm{Regret}}(\mathcal{M}^\star,\mathcal{D}_N)\coloneqq J^{\pi^\star}(\mathcal{M}^\star)- J^{\pi_\textrm{Bayes}^\star}(\mathcal{M}^\star,\mathcal{D}_N).
\end{align}
Our first result shows that true regret can be bounded using the PIL, defined as: 
\begin{align}
    \mathcal{I}_N^\pi\coloneqq\mathbb{E}_{\theta\sim P_\Theta(\mathcal{D}_N)}\left[\mathbb{E}_{s,a\sim \rho_\pi^\star}\left[ \textrm{KL}\left(P_{R,S}^\star(s,a)\Vert P_{R,S}(s,a,\theta) \right)\right]\right],
\end{align} 
where $\rho_\pi^\star\coloneqq\mathbb{E}_{i\sim \mathcal{AG(\gamma)}}\left[\mathbb{E}_{j\sim \mathcal{U}_i}\left[ P_{j,\pi}^\star\right]\right]$ is the arithmetico-geometric ergodic state-action distribution induced by a policy $\pi$ and recall $P_{j,\pi}^\star$ denotes the distribution over a $j$-length trajectory $h_j$. $\rho_\pi^\star$ places mass over state-action pairs according to how much errors in the model influence the regret at each state. Regions of the state-action space that require more timesteps to reach from initial states are weighted significantly less than those that are encountered earlier and more frequently, as state errors encountered early accumulate in each prediction from that timestep onwards.

The PIL has an intuitive information-geometric interpretation: the inner expectation $\mathbb{E}_{s,a\sim \rho_\pi^\star}\left[ \textrm{KL}\left(P_{R,S}^\star(s,a)\Vert P_{R,S}(s,a,\theta) \right)\right]$ measures the distance between the model and the true distribution in terms of the information lost when approximating $P_{R,S}^\star(s,a)$ with $P_{R,S}(s,a,\theta)$, averaged across all states. The PIL thus measures how close the posterior's belief is to the truth according to the average information lost under the posterior expectation. We now prove that regret is upper bounded by the PIL:

\begin{theorem} \label{proof:regret_bound_kl} Let $\mathcal{R}_{\max}\coloneqq \frac{(r_\textrm{max}-r_\textrm{min})}{1-\gamma}$ denote the maximum possible regret for the MDP. Using the PIL: $\mathcal{I}_N^\pi$, the true regret is bounded as: 
\begin{align}
    \textnormal{\textrm{Regret}}(\mathcal{M}^\star,\mathcal{D}_N)\le2\mathcal{R}_{\max}\cdot\sup_\pi\sqrt{1-\exp\left(-\frac{\mathcal{I}_N^\pi}{1-\gamma}\right)}.
    \label{eq:information_rate}
\end{align}\vspace{-0.2cm}
\end{theorem}
The key insight from \cref{proof:regret_bound_kl} is that the rate at which regret decreases with $N$ is governed by the rate at which the PIL decreases, which is known as the \emph{information rate}. This measures how much information the model has gained from an incremental amount of data. Fast information rates imply that highly informative posteriors can be learned from limited data, as regret decreases at least as quickly. How fast the PIL decreases depends on the model specification, the prior used, the data coverage and the underlying MDP. This makes it an ideal metric to hyperparameter associated with the dynamics model and approximate inference method. 

\subsection{Gaussian World Models}\label{sec:gauss_world}
Formulating our bound in terms of the PIL ties the regret to the KL divergence over the reward-state model: $\textrm{KL}\left(P_{R,S}^\star(s,a)\Vert P_{R,S}(s,a,\mathcal{D}_N) \right)$. Not only is this mathematically more convenient, but it means that the PIL is easy to estimate in practice. As an example, many methods use an ensemble of Gaussian reward and state transition models of the form introduced by \citet{Chua18}, where the individual reward and state transition models are given by: 
\begin{align}
    P_R(s,a,\theta)=\mathcal{N}(r_\theta(s,a),\sigma^2_r(s,a)),\quad  P_S(s,a,\theta)=\mathcal{N}(s'_\theta(s,a),I\sigma^2_s(s,a)),\label{eq:Gaussian_world_model}
\end{align}
with variances characterised by $\sigma^2_r$ and $\sigma^2_s$, mean reward function  $r_\theta(s,a)$ and mean state transition function $s'_\theta(s,a)$. 
Using an ensemble of Gaussian World Models admits a natural separation of aleatoric and epistemic uncertainty: the aleatoric uncertainty can be represented by the individual output variances $\sigma^2_r$ and $I\sigma^2_s(s,a)$, or mean of the learned variances, and the epistemic uncertainty by the ensemble disagreement, or variance of the means. Using a Gaussian world model, we find the PIL takes a convenient and intuitive form. Let $r(s,a,\mathcal{D}_N)\coloneqq \mathbb{E}_{\theta\sim P_\Theta (\mathcal{D}_N)}\left[r_\theta(s,a)\right]$  and $s'(s,a,\mathcal{D}_N)\coloneqq \mathbb{E}_{\theta\sim P_\Theta (\mathcal{D}_N)}\left[s_\theta(s,a)\right]$ denote the Bayesian mean reward and state transition functions and $r^\star(s,a)$ and ${s^\star}'(s,a)$ denote the true mean functions. We define the mean squared error between the true and Bayesian mean functions as:
\begin{align}
    \mathcal{E}(\mathcal{D}_N,\mathcal{M}^\star)\coloneqq  \mathbb{E}_{(s,a)\sim \rho_\pi^\star}\left[\frac{\lVert r(s,a,\mathcal{D}_N)-r^\star(s,a)\rVert^2_2}{2\sigma_r^2(s,a)}+\frac{\lVert s'(s,a,\mathcal{D}_N)-{s^\star}'(s,a)\rVert^2_2}{2\sigma_s^2(s,a)}\right],
    \label{eq:MSE_term}
\end{align}
and the predictive variance as:
\begin{align}
    \mathcal{V}(\mathcal{D}_N)\coloneqq  \mathbb{E}_{(s,a)\sim \rho_\pi^\star}\left[\mathbb{E}_{\theta\sim P_\Theta(\mathcal{D}_N)}\left[\frac{\lVert r(s,a,\mathcal{D}_N)-r_\theta(s,a)\rVert^2_2}{2\sigma_r^2(s,a)}+\frac{\lVert s'(s,a,\mathcal{D}_N)-s'_\theta(s,a)\rVert^2_2}{2\sigma_s^2(s,a)} \right]\right].
    \label{eq:predictive_variance_term}
\end{align}
We now re-write the PIL for the Gaussian world model using these two terms:

\begin{proposition} \label{proof:info_prop} 
Using the Gaussian world model in \cref{eq:Gaussian_world_model}, it follows that 
\begin{align}
    \mathcal{I}_N^\pi=\mathcal{E}(\mathcal{D}_N,\mathcal{M}^\star)+\mathcal{V}(\mathcal{D}_N).
\end{align}
\end{proposition}

\cref{proof:info_prop} shows that the PIL is governed by i) the MSE of the point estimate $ \mathcal{E}(\mathcal{D}_N,\mathcal{M}^\star)$, which characterises how quickly the Bayesian mean function converges to the true function; and ii) the predictive variance $\mathcal{V}(\mathcal{D}_N)$, which characterises the epistemic uncertainty in the model and is a purely Bayesian term. The PIL can easily be estimated by estimating $\mathcal{E}(\mathcal{D}_N,\mathcal{M}^\star)$ using the empirical MSE under the offline data distribution and estimating $\mathcal{V}(\mathcal{D}_N)$ using approximate posterior sampling. This decomposition provides a principled diagnostic mechanism based on the relative magnitudes and contraction behaviour of $\mathcal{E}(\mathcal{D}_N,\mathcal{M}^\star)$ and $\mathcal{V}(\mathcal{D}_N)$. In particular, imbalances between these terms can signal model misspecification, insufficient data coverage, or miscalibrated uncertainty. We exploit this property directly for hyperparameter tuning and model validation in \cref{sec:sorel}.
\vspace{-0.1cm}
\subsection{Frequentist Justification of Bayesian Offline RL}
\label{sec:frequentist_regret_analysis}
As discussed in \cref{sec:obrl} and derived in \cref{app:brl_primer}, all solutions to a Bayesian offline RL problem minimise Bayes regret, which provides a Bayesian justification for our method. We now provide a complementary frequentist justification. Specifically, we prove that the regret under the true data-generating distribution converges at the same  $\mathcal{O}\left(\nicefrac{1}{\sqrt{N}}\right)$ rate established in prior offline RL work \citep{Yu20,Kidambi20} under standard local asymptotic normality (LAN) assumptions \citep{LeCam53,barron1988exponential,Clarke1990,Komaki96,Hartigan1998,Barron1999,Aslan06}. These assumptions are formally stated and discussed in \cref{app:frequentist_justification}.

Frequentist analyses of offline RL span a broad spectrum of structural assumptions. At one extreme, minimax analyses evaluate performance uniformly over large classes of MDPs, effectively calibrating guarantees against the most challenging admissible environments \citep{Wang2020}. While such worst-case results sharply delineate statistical limits, they are agnostic to how representative those environments are in practice. Consequently, under minimal structure one obtains intrinsic hardness results \citep{Wang2020}, whereas under additional structural or parametric assumptions, classical statistical efficiency becomes attainable \citep{Agarwal2021,Li2022}.

The $\mathcal{O}\left(\nicefrac{1}{\sqrt{N}}\right)$ scaling we obtain is the canonical parametric asymptotic convergence rate and is known to be minimax-optimal in regular finite-dimensional models for offline RL \citep{Agarwal2021,Li2022}. Modern model-based offline RL methods operate precisely in such structured regimes, positing smooth parametric dynamics models and relying on likelihood-based estimation \citep{Yu20,Kidambi20}. From a Bayesian standpoint, LAN formalises this modelling commitment: a smooth, identifiable parametric family in which posterior uncertainty contracts at the canonical 
$\mathcal{O}\left(\nicefrac{1}{\sqrt{N}}\right)$ parametric 
 rate. Consequently, under the conditions of Assumption~\ref{ass:model_regularity}, our bound is asymptotically rate-tight within this nonlinear parametric function class.

\begin{theorem} \label{proof:info_rate} Let the data be drawn from the underlying true distribution $\mathcal{D}_N\sim P_\textrm{Data}^\star$.
Under standard local asymptotic normality assumptions (see \cref{ass:model_regularity} in \cref{app:frequentist_justification}), there exists some constant $0<C<\infty$ such that for sufficiently large $N$: 
\begin{align}
\mathbb{E}_{\mathcal{D}_N\sim P_\textrm{Data}^\star}\left[\textnormal{\textrm{Regret}}(\mathcal{M}^\star,\mathcal{D}_N)\right] \le 2\mathcal{R}_{\max}\cdot\sqrt{1-\exp\left(-\frac{Cd}{(1-\gamma)N}\right)}=\mathcal{O}\left(\sqrt{\frac{d}{(1-\gamma)N}}\right)
\label{eq:frequentist}
\end{align} 
\end{theorem}
We also extend this analysis to include the case of model-misspecification and sub-optimal policy learning in \cref {app:miss_proof_info_rate}, obtaining tighter bounds than obtained previous in literature \citep{Kidambi20} as $\gamma\rightarrow 1$. Finally, we remark that asymptotic frequentist regret bounds like \cref{eq:frequentist} are primarily rate characterisations: they describe how regret scales in the large-sample limit but do not provide calibrated, dataset-specific diagnostics. Bounds of the form $\textrm{Regret}_N\le \nicefrac{C}{\sqrt{N}}+m$  depend on problem-specific constants $C,m$ and unspecified large-sample regimes that are  inaccessible in practice. Consequently, while such results are valuable for theoretical comparison, they offer limited use as practical metrics for assessing dataset sufficiency, posterior calibration, or hyperparameter selection in finite-sample settings. By contrast, the PIL metric yields directly observable quantities that can be monitored and acted upon during training.

\section{Methods}
\label{sec:regret_approximation}
\subsection{SOReL}\label{sec:sorel}
\begin{wrapfigure}{r}{0.5\textwidth}
\vspace{-2.4cm}
 \begin{minipage}{0.49\textwidth}
\begin{algorithm}[H]
\caption{SOReL$(P_\Theta,\mathcal{D}_N,\phi,\pi)$}  \label{alg:sorel}
\footnotesize
\begin{algorithmic}
\State \textit{Hyperparameter tuning:}
\State $\textrm{\textcolor{RedViolet}{$\phi_{I}$}},\textrm{\textcolor{OliveGreen}{$\phi_{II}$}}\gets \argmin_{\textrm{\textcolor{RedViolet}{$\phi_{I}$}},\textrm{\textcolor{OliveGreen}{$\phi_{II}$}}} \textrm{PIL}(\textrm{\textcolor{RedViolet}{$\phi_{I}$}},\textrm{\textcolor{OliveGreen}{$\phi_{II}$}},\mathcal{D}_N)$ \State \hskip1em $\text{ s.t. } \mathcal{E}(\mathcal{D}_N,\mathcal{M}^\star)\approx \mathcal{V}(\mathcal{D}_N)$ 

\State $\textrm{\textcolor{RedOrange}{$\phi_{III}$}}\gets 
 \argmax_{\textrm{\textcolor{RedOrange}{$\phi_{III}$}}} \text{PredValue} 
(\textrm{\textcolor{RedViolet}{$\phi_{I}$}}, \textrm{\textcolor{OliveGreen}{$\phi_{II}$}},\textrm{\textcolor{RedOrange}{$\phi_{III}$}},\mathcal{D}_N,\pi)$ 

\State \textit{Policy Learning and Value Estimation:}
       \State  $\pi \gets \textrm{BayesPolicyImprovement}(\textrm{\textcolor{RedViolet}{$\phi_{I}$}},\textrm{\textcolor{OliveGreen}{$\phi_{II}$}},\textrm{\textcolor{RedOrange}{$\phi_{III}$}},\mathcal{D}_N)$
\State $J_N \gets \text{PredValue} (\textrm{\textcolor{RedViolet}{$\phi_{I}$}}, \textrm{\textcolor{OliveGreen}{$\phi_{II}$}},\textrm{\textcolor{RedOrange}{$\phi_{III}$}},\mathcal{D}_N,\pi)$
\State $\textrm{metrics}\leftarrow \textrm{GetMetrics}(\textrm{\textcolor{RedViolet}{$\phi_{I}$}}, \textrm{\textcolor{OliveGreen}{$\phi_{II}$}},\textrm{\textcolor{RedOrange}{$\phi_{III}$}},\mathcal{D}_N,\pi)$
\State \Return $\phi$, $\pi$, $J_N$, metrics
\end{algorithmic}
\end{algorithm}
\end{minipage}\vspace{-0.2cm}
\end{wrapfigure}
We now introduce SOReL in \cref{alg:sorel}, our fully offline Bayesian offline RL algorithm with value estimation and hyperparameter tuning. The algorithm shows a single step in an iterative process that updates the policy $\pi$ and hyperparameters $\phi$ whilst returning a value estimate after each iteration and a set of metrics that include the PIL values and predictive variance. The metrics can be used for diagnostics and assessing when the desired value has been achieved depending upon the setting. In our SOReL framework, there are three sets of hyperparameters:  \textcolor{RedViolet}{$\phi_{I}$ the model} (such as the architecture and hyperparameters for learning a neural-network function approximator); \textcolor{OliveGreen}{$\phi_{II}$ the approximate inference method} (such as the number of ensemble members for RP); and \textcolor{RedOrange}{$\phi_{III}$ the BAMDP solver} (the hyperparameters of a Bayesian meta-learning algorithm like RNN-PPO). Sets \textcolor{RedViolet}{$\phi_{I}$} and \textcolor{OliveGreen}{$\phi_{II}$} are tuned jointly to both minimise the PIL and ensure a roughly even split between the predictive variance and MSE loss terms. Set \textcolor{RedOrange}{$\phi_{III}$} is then tuned to minimise approximate regret based on the now-fixed model and approximate posterior: for each combination of hyperparameters, we update the policy using the approximate BAMDP solver (BayesPolicyImprovement($\cdot$)), and choose the combination whose policy leads to the highest approximate value. 
 
\textbf{Fixing Issue I - PIL Monitoring:} To tune sets \textcolor{RedViolet}{$\phi_{I}$} and \textcolor{OliveGreen}{$\phi_{II}$}, we monitor the change in the PIL. Our objective is to select hyperparameters that minimise the PIL whilst ensuring that the MSE term (c.f. \cref{eq:MSE_term}) closely matches the predictive variance term (c.f. \cref{eq:predictive_variance_term}), i.e. $\mathcal{E}(\mathcal{D}_N,\mathcal{M}^\star)\approx \mathcal{V}(\mathcal{D}_N)$. The PIL alignment provides a powerful bias–variance diagnostic during training: $\mathcal{E}$ measures systematic model error (bias), whereas $\mathcal{V}$ captures epistemic uncertainty (posterior variance). If $\mathcal{E}\gg\mathcal{V}$ and fails to contract as $N$ increases, the posterior is confident but biased, suggesting overfitting due to under-dispersed posterior uncertainty. Under a well-specified and well-calibrated model, both terms contract at comparable rates as $N$ increases; persistent imbalance signals hyperparameter misconfiguration in \textcolor{RedViolet}{$\phi_{I}$} and/or \textcolor{OliveGreen}{$\phi_{II}$}. Since miscalibrated uncertainty leads to unreliable value estimates, we tune \textcolor{RedViolet}{$\phi_{I}$} and \textcolor{OliveGreen}{$\phi_{II}$} first in \cref{alg:sorel}. Empirically, as shown in \cref{sec:sorel_validation}, when $\mathcal{E}\approx \mathcal{V}$ and both are small, the approximate value closely tracks the true value. Likewise, in regimes with limited data coverage, policies must be regularised to remain close to the support of the training distribution (e.g. via a pessimistic prior in SOReL, or pessimistic mechanisms such as MOPO’s reward penalty and MOReL’s uncertainty-aware dynamics in TOReL). In this regime, value estimates are inherently biased, which is unavoidable in offline RL approaches with limited data coverage. However, this failure mode is reflected in an increase in ensemble variance due to the large epistemic uncertainty over regions with limited data, and consequently in the variance term of the PIL, which provides a practical signal that the policy is operating in poorly supported regions. We note that standard OPE methods do not expose this uncertainty.

\textbf{Fixing Issues I and II - Bayesian Value Estimation:} 
We estimate the value of the Bayes-optimal policy using the posterior predictive median:
\begin{align}
\hat{J}^{\pi_\textrm{Bayes}^\star}(\mathcal{M}^\star) 
&= \mathbb{M}_{\theta\sim P_\Theta(\mathcal{D}_N),h_\infty\sim P_\infty^\pi(\theta)}\left[ R(h_\infty)\right],
\label{eq:sorel_approximate_regret}
\end{align}
where $\mathbb{M}_{\theta\sim P_\Theta(\mathcal{D}_N),h_\infty\sim P_\infty^\pi(\theta)}\left[ R(h_\infty)\right]$ denotes the median predictive return. In \cref{alg:sorel}, $\hat{J}^{\pi_\textrm{Bayes}^\star}(\mathcal{M}^\star)$ is estimated in $\text{PredValue}(\cdot)$ via Monte Carlo sampling: ensemble members are drawn from the approximate posterior, the Bayes-optimal policy is rolled out in each sampled model, and the empirical median of returns is computed. Alternative Bayesian estimators, such as the predictive mean or lower quantiles, are also valid and correspond to different loss functions or levels of conservatism (see \cref{sec:regret_approximators}). We adopt the predictive median because it is the Bayes estimator under absolute error loss and is robust to skewed or heavy-tailed predictive return distributions. Under symmetric predictive distributions the median and mean coincide, but in skewed posterior predictive settings the median provides greater stability. In particular, it avoids being overly conservative while remaining insensitive to outlier posterior samples corresponding to rare but highly optimistic or pessimistic dynamics. Our empirical evaluations in \cref{sec:sorel_validation} support this choice. The posterior predictive median further enables fully offline tuning of hyperparameter set \textcolor{RedOrange}{$\phi_{III}$}, by selecting the policy configuration that maximises the predictive value in \cref{alg:sorel}. Moreover, the predictive variance across rollouts characterises the epistemic and aleatoric uncertainty in the value estimate (returned in the `metrics' set in \cref{alg:sorel}) and provides a quantitative signal for the practitioner to assess whether the learned policy is sufficiently reliable for deployment.
\subsection{TOReL} 
\label{sec:torel}
\begin{wrapfigure}{r}{0.5\textwidth}
\vspace{-1.7cm}
 \begin{minipage}{0.49\textwidth}
\begin{algorithm}[H]
\footnotesize
\caption{TOReL$(P_\Theta,\mathcal{D}_N,\phi)$}  \label{alg:torel}
\begin{algorithmic}
\State $\textrm{\textcolor{RedViolet}{$\phi_{I}$}}, \textrm{\textcolor{OliveGreen}{$\phi_{II}$}}\gets \argmin_{\textrm{\textcolor{RedViolet}{$\phi_{I}$}},\textrm{\textcolor{OliveGreen}{$\phi_{II}$}}} \textrm{PIL}(\textrm{\textcolor{RedViolet}{$\phi_{I}$}},\textrm{\textcolor{OliveGreen}{$\phi_{II}$}},\mathcal{D}_N)$
\State \hskip1em $[\text{ s.t. } \mathcal{E}(\mathcal{D}_N,\mathcal{M}^\star)\approx \mathcal{V}(\mathcal{D}_N), (\text{model-based})]$ 
\State $\textrm{\textcolor{RedOrange}{$\phi_{III}$}}\gets \argmax_{\textrm{\textcolor{RedOrange}{$\phi_{III}$}}} \text{ValueMetric} (\textrm{\textcolor{RedViolet}{$\phi_{I}$}},\textrm{\textcolor{OliveGreen}{$\phi_{II}$}},\textrm{\textcolor{RedOrange}{$\phi_{III}$}}, \mathcal{D}_N)$ 
\State $\pi^\star_\textrm{TOReL} \gets \begin{cases}
    \textrm{ORL}(\textrm{\textcolor{RedOrange}{$\phi_{III}$}}, \mathcal{D}_N),& \textrm{(model-free)}\\
    \textrm{ORL}(\textrm{\textcolor{RedViolet}{$\phi_{I}$}}, \textrm{\textcolor{OliveGreen}{$\phi_{II}$}},\textrm{\textcolor{RedOrange}{$\phi_{III}$}}, \mathcal{D}_N),& \textrm{(model-based)}\\
\end{cases}$
\State \Return $\phi$, $\pi^\star_\textrm{TOReL}$
\end{algorithmic}
\end{algorithm}
\end{minipage}\vspace{-0.2cm}
\end{wrapfigure}
SOReL's offline hyperparameter tuning methods are directly applicable to general offline RL approaches, allowing us to address \textbf{Issue I} for existing offline methods. We now adapt these methods to derive a general tuning for offline reinforcement learning approach called TOReL, shown in \cref{alg:torel}. A policy is learned offline using a planning algorithm, denoted by ORL. There thus exists a corresponding set of hyperparameters associated with \textcolor{RedOrange}{$\phi_{III}$ the offline planner}. For model-based methods with uncertainty estimation like MOReL \cite{Kidambi20} and MOPO \cite{Yu20}, we can exactly adapt SOREL's PIL tuning method to the parameters associated with: \textcolor{RedViolet}{$\phi_{I}$ the dynamics model} and  \textcolor{OliveGreen}{$\phi_{II}$ uncertainty estimation}. For all other methods, we introduce and learn a dynamics model and an approximate inference method like in SOReL and jointly tune the corresponding hyperparameters \textcolor{RedViolet}{$\phi_{I}$} and \textcolor{OliveGreen}{$\phi_{II}$} to minimise the PIL. Since the policy learned with ORL is typically neither Bayes-optimal nor robust to model uncertainty, we expect that applying SOReL's value estimation method to more general methods in  TOReL will not yield an accurate estimate of the value in terms of its absolute value. Instead, we treat the predictive value in \cref{eq:sorel_approximate_regret} as a \emph{value metric}  that is positively correlated with true expected returns, and use this to tune ORL parameters \textcolor{RedOrange}{$\phi_{III}$}. Our empirical evaluations in \cref{sec:torel_validation} support this hypothesis. We note that in model-free methods, the dynamics model and an approximate inference method are not used in policy learning, only to choose hyperparameters.
\vspace{-0.2cm}
\section{Experiments}\vspace{-0.1cm}
\label{sec:experiments} 
The goal of our experiments is to confirm our methods provide a solution to \textbf{Issues I \& II}: first, to evaluate SOReL's ability to approximate online performance entirely offline; and second, to evaluate TOReL's ability to tune hyperparameters of general ORL algorithms entirely offline. 
\vspace{-0.1cm}
\subsection{Zero-Shot Fully Offline RL with Value Estimation}
\label{sec:sorel_validation}
Our first experiment evaluates SOReL in the strict zero-shot setting: policy learning, hyperparameter tuning, and value estimation are performed using only offline data, with no interaction with the true environment prior to deployment. We assess whether the resulting offline value estimates reliably predict realised online performance. We test SOReL in five environments — two Gymnax \citep{gymnax2022github} and three Brax \citep{freeman2021braxdifferentiablephysics} — using datasets we collect to ensure broad behavioural diversity and state-action coverage (see \cref{app:sorel_datasets}). To emulate practical usage, we progressively increase dataset size and estimate value entirely offline. We deploy each policy in the true environment to validate the estimated value, but in practice the policy would only be deployed once the estimated value is high enough and uncertainty small enough. 

\begin{figure}
    \vspace{-0.8cm}
    \centering
    \includegraphics[width=0.85\linewidth]{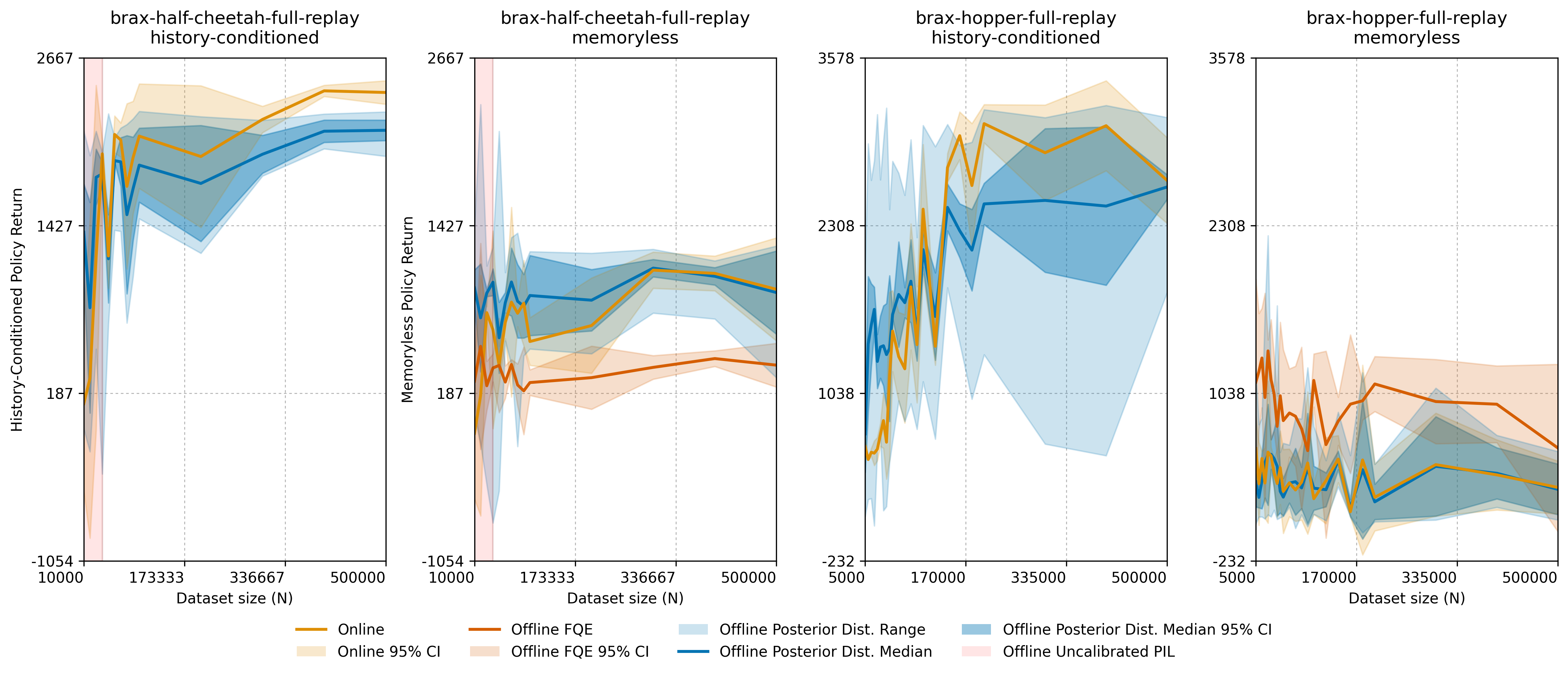}
    \vspace{-0.3cm}\caption{Evaluation of SOReL in zero-shot setting over 3 seeds. Policy hyperparameters are tuned \textbf{entirely offline} using the predictive value estimate. History is set to zero (memoryless) for  comparison with FQE.}
    \label{fig:sorel_overall}
    \vspace{-0.6cm}
\end{figure}

Results (\cref{fig:sorel_overall} and \cref{fig:sorel_5_panel}) show that the posterior predictive median closely tracks realised online returns, which consistently lie within the predictive range induced by sampling dynamics from the posterior. Regions shaded in red (e.g., small datasets in Half-Cheetah) correspond to miscalibrated PIL, signalling unreliable estimates. Once the PIL is calibrated and predictive variance sufficiently reduced, deployment can be undertaken with confidence. The diagnostics are actionable. In Half-Cheetah, small datasets yield wide predictive ranges and poorly calibrated PIL, correctly warning against deployment; with more data, calibration improves and the median aligns tightly with online returns. In Hopper, although the PIL remains calibrated, predictive variance stays large, signalling persistent uncertainty. Crucially, this uncertainty is visible \emph{offline}, enabling informed deployment decisions without online interaction.
All SOReL hyperparameters are tuned entirely offline. As shown in \cref{fig:sorel_hyper_sweep} in \cref{app:sorel_further_results}, predictive-value-based tuning substantially improves performance, with both Half-Cheetah ($r = 0.93, p = 0.00$) and Hopper ($r = 0.70, p = 0.00$) having statistically significant ($p < 0.05$), strong positive ($r > 0.5$) Pearson correlation between offline predictive value and realised online return. Moreover, SOReL learns policies that outperform the behaviour policies used to generate the datasets. For example, on a subset of the Half-Cheetah dataset, SOReL achieves nearly $1.5 \times$ the return of the best data-collecting policy (\cref{fig:brax_datasets}), demonstrating its ability to improve beyond suboptimal data.
\vspace{-0.1cm}
\paragraph{Benchmarking Value Estimation} We benchmark SOReL's ability to approximate online performance with fitted Q-evaluation (FQE) as used in \citet{Paine20}, which is the closest existing algorithm to SOReL. We cannot benchmark against OPE metrics based on importance sampling  \citep{kostrikov2020statisticalbootstrappinguncertaintyestimation} or double robustness \citep{thomas2016dataefficientoffpolicypolicyevaluation}, which require datasets containing trajectories rather than individual datapoints. As SOReL is a Bayesian approach, policies condition on histories to represent belief states. In contrast, FQE is designed for state-conditioned policies. For fair comparison, we reset the hidden states of all trajectories on each step to zero to project the history-conditioned policy to a state-conditioned policy. We compare SOReL’s predictive value for the history-zeroed policy (as defined in \cref{eq:sorel_approximate_regret}) with FQE’s value estimate by evaluating both against the ground-truth returns of this policy.

Results (\cref{fig:sorel_overall} in \cref{app:sorel_further_results}) show that SOReL’s predictive median tracks the realised online return more closely than FQE, with an average absolute error of 120 versus 470 for Half-Cheetah (over the points where the PIL is well-calibrated) and 70 versus 550 for Hopper. While FQE can approximate value in some regimes, it frequently exhibits overestimation or unstable behaviour for small datasets. In contrast, SOReL’s posterior predictive range consistently contains the true return once the PIL is calibrated, and contracts as dataset size increases, reflecting improved data sufficiency. Moreover, when SOReL’s predictive estimate deviates from the realised return, it tends to do so conservatively, underestimating rather than overestimating performance in low-data regimes. This pessimistic bias is desirable in safety-critical settings, where optimistic errors may lead to premature deployment. Unlike FQE, SOReL provides a calibrated uncertainty quantification alongside its value estimate; note that standard deviations are across policies, rather than for a single policy, like SOReL's return range is. The posterior predictive variance meaningfully reflects epistemic uncertainty: wide ranges for small datasets signal unreliable estimates, while contraction with additional data indicates increasing confidence. Finally, SOReL's PIL approach affords a principled diagnostic for detecting model misspecification, support mismatch, or insufficient data, while FQE does not provide an analogue of the PIL that links uncertainty calibration to regret or guides hyperparameter tuning.

\begin{figure}[t]
\vspace{-0.8cm}
    \begin{subfigure}[b]{\textwidth}
        \centering
        \includegraphics[width=0.89\linewidth]{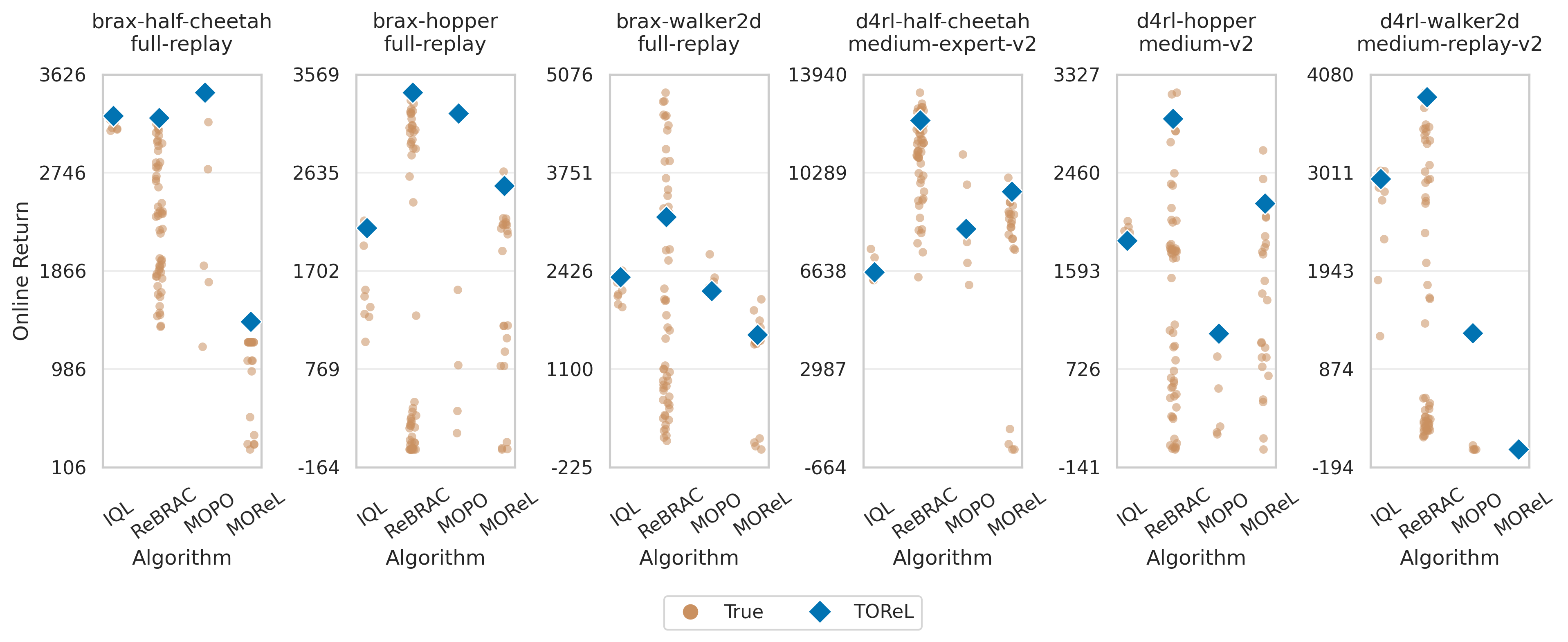}
        \label{fig:torel_barplot}
    \end{subfigure}
    \hfill
    \begin{subfigure}[b]{\textwidth}
        \centering
        \includegraphics[width=0.89\linewidth]{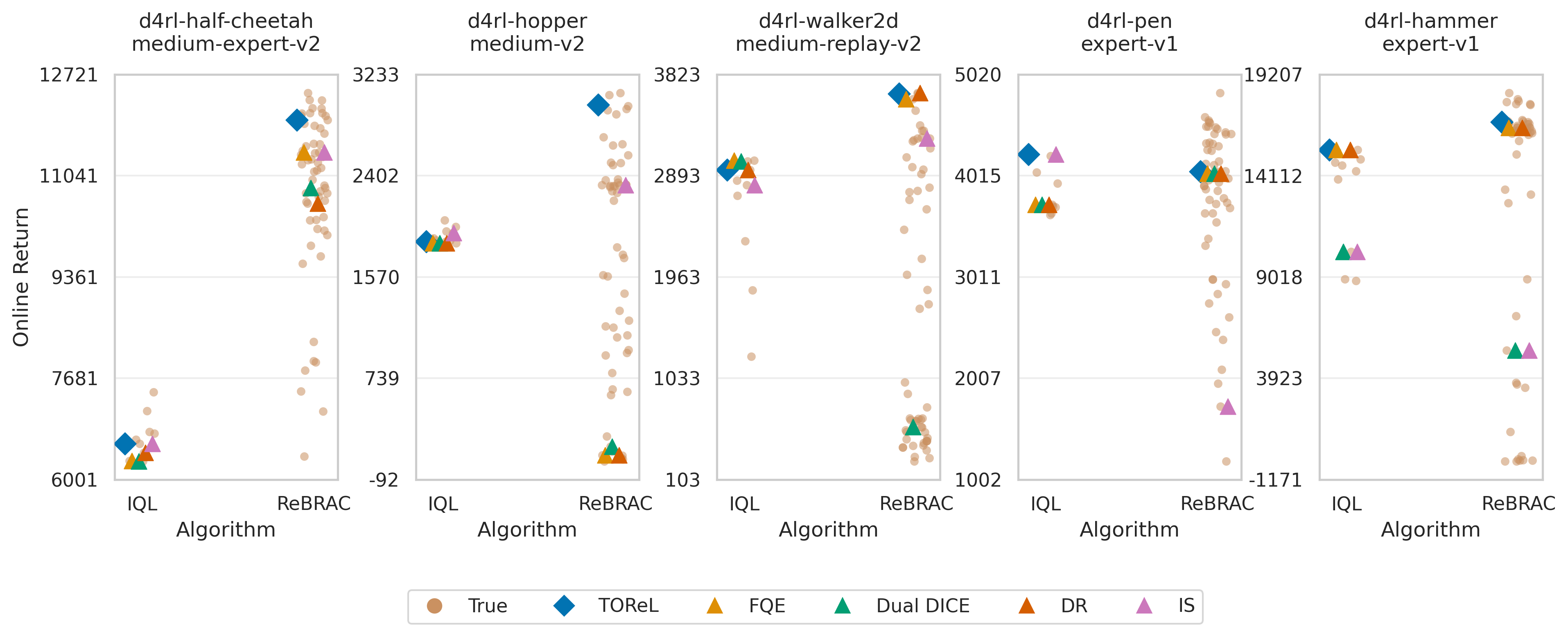}
        \label{fig:d4rl_torel_barchart}
    \end{subfigure}
    \vspace{-0.7cm}
    \caption{True online returns achieved by different hyperparameter combinations for different ORL algorithms and datasets. \textbf{Top}: hyperparameters selected by \textbf{TOReL} for IQL, ReBRAC, MOPO and MOReL. \textbf{Bottom:} hyperparameters selected by \textbf{TOReL} and 4 OPE metrics (\textbf{FQE}, \textbf{Dual DICE}, \textbf{DR} and \textbf{IS}) for IQL and ReBRAC. Error bars omitted due to the prohibitive cost of repeated runs ($\sim$2000 experiments).}
    \label{fig:overall_return_barplot}
    \vspace{-0.6cm}
\end{figure}
These results highlight three key differences. Firstly, SOReL's predictive median approach yields more reliable value estimates in the zero-shot setting, with conservative behaviour when uncertainty is high. Secondly, SOReL provides actionable uncertainty diagnostics that allow practitioners to determine \emph{offline} whether a value estimate is trustworthy. Thirdly, we remark that \citet{Paine20}'s approach FQE still requires online hyperparameter tuning for its own hyperparameter set, whereas SOReL's approach is fully offline.
\vspace{-0.1cm}
\subsection{ORL Offline Hyperparameter Tuning}
\begin{figure}[t]
    \vspace{-0.8cm}
    \centering
    \includegraphics[width=0.89\linewidth]{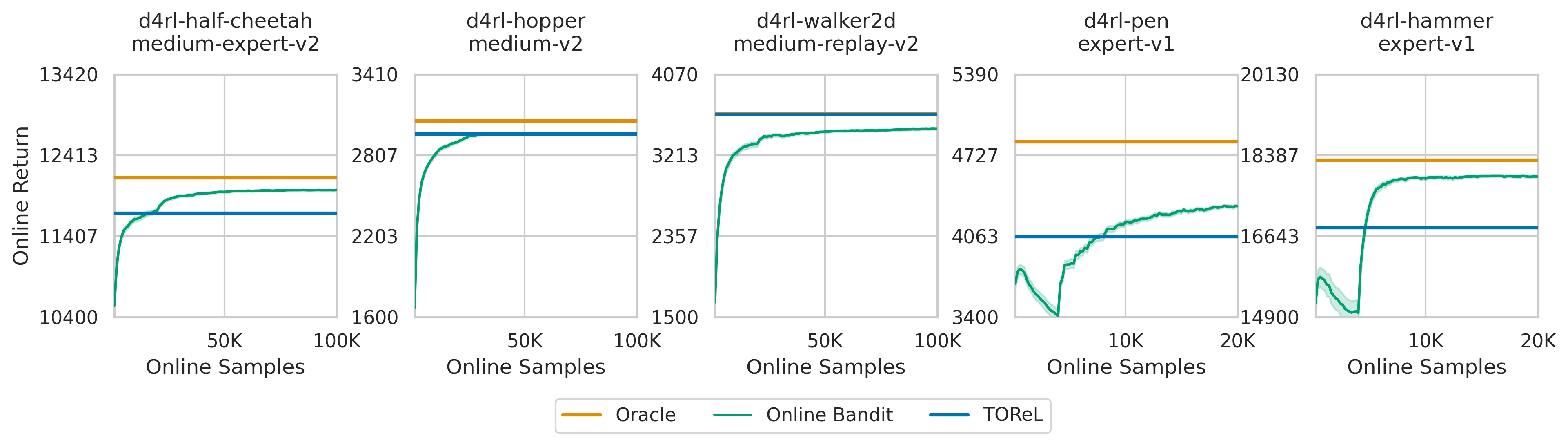}\vspace{-0.3cm}
    \caption{\textbf{TOReL}'s performance compared to an \textbf{online} bandit-based hyperparameter tuning algorithm. Shaded region indicates 95\% bootstrap CI.}
\label{fig:d4rl_torel_regret_sample_efficiency}
\vspace{-0.6cm}
\end{figure}

\vspace{-0.1cm}
\label{sec:torel_validation}
We carry out three experiments to test TOReL's ability as a fully offline hyperparameter tuning method for general ORL. 
Firstly, we apply TOReL to four existing ORL algorithms; two model-free, IQL \citep{Kostrikov21} and ReBRAC \citep{tarasov2023revisiting}, and two model-based, MOPO \citep{Yu20} and MOReL \cite{Kidambi20}. We assess whether TOReL can identify the hyperparameter configuration yielding the highest true online return (\cref{fig:torel_barplot}). Second, we compare TOReL against OPE-based hyperparameter selection methods following \citet{Paine20}. Using five D4RL datasets (which contain full trajectories required by OPE methods), we evaluate IQL and ReBRAC under hyperparameters selected by TOReL and by FQE, Dual DICE, DR, and IS. As OPE algorithms have hyperparameters that are tuned online, we use the values given in \citet{fu2021benchmarksdeepoffpolicyevaluation}.  We learn policies for the hyperparameter combinations suggested by \citet{unifloral2025}. Results are shown in \cref{fig:d4rl_torel_barchart}. Finally, we compare TOReL against \citet{unifloral2025}'s state-of-the-art online UCB-based hyperparameter tuning algorithm, measuring the online sample complexity required to replicate the performance obtained by TOReL’s fully offline tuning (\cref{fig:d4rl_torel_regret_sample_efficiency}).
\cref{fig:torel_barplot} and \cref{fig:d4rl_torel_barchart} shows TOReL identifes hyperparameters with the high online returns across both different algorithms and datasets. Across all evaluated datasets, TOReL matches or exceeds the best online return achieved by hyperparameters selected via FQE, Dual DICE, DR, and IS. Moreover, TOReL’s offline predictive metric exhibits strong positive correlation with true performance; in particular, it achieves statistically significant strong positive correlation on 18 of 28 hyperparameter tuning tasks (see \cref{tab:fpe_main} for Pearson correlations), and on 6 of 10 tasks where we benchmark against standard OPE metrics (compared to 4 for FQE, 0 for Dual DICE, and 1 each for DR and IS). Learning the ORL policy with ReBRAC and tuning hyperparameters with TOReL is a particularly strong combination. 
Finally, \cref{fig:d4rl_torel_regret_sample_efficiency} highlights the sample-efficiency advantage of TOReL's fully offline tuning. From the spread of returns for different ReBRAC hyperparameter combinations, \cref{fig:d4rl_torel_regret_sample_efficiency} demonstrates that online bandit-based tuning requires significant interaction to reliably identify high-performing settings. An additional 16000, 36000, 7700 and 5000 online samples are required to match TOReL's performance on Half-Cheetah, Hopper, Pen and Hammer, respectively. For Walker2d, TOReL’s performance remains unmatched within the evaluated interaction budget. 

\vspace{-0.2cm}
\section{Conclusion}
\vspace{-0.1cm}
High online sample complexity and lack of performance guarantees of existing methods present a major barrier to the widespread adoption of offline RL. In this paper, we introduce SOReL and TOReL, two theoretically grounded approaches to tackle these core issues. SOReL is a model-based Bayesian method that exploits predictive uncertainty to approximate online value entirely offline. Hyperparameters of the dynamics model and posterior are tuned by minimising and calibrating the PIL, while hyperparameters of the BAMDP solver are selected using predictive performance from posterior-sampled rollouts. TOReL extends this fully offline tuning procedure to general offline RL algorithms. Empirically, SOReL provides reliable offline performance estimates, and TOReL enables fully offline hyperparameter selection, substantially reducing online tuning requirements without sacrificing performance. 

\textbf{Broader impact:} This paper presents work whose goal is to improve the safety and efficacy of  offline RL. Our work therefore takes a step towards the development of safe offline RL methods with accurate value estimate that act predictably once deployed.


\bibliography{SOReL}
\bibliographystyle{rlj}


\beginSupplementaryMaterials
\appendix

\section{Primer on Bayesian RL}
\label{app:brl_primer}
A Bayesian epistemology characterises the agent's uncertainty in the MDP through distributions over any unknown variable \citep{Martin67,Duff02}. In our learning problem, a Bayesian first specifies a model $P_{R,S}(s_t,a_t)$ over the unknown state transition and reward distribution, representing a hypothesis space of possible environment dynamics. We focus on a parametric model:  $p(r_t,s_{t+1} \vert s_t,a_t,\theta)$ with each $\theta\in\Theta\subseteq \mathbb{R}^d$ representing a hypothesis about the MDP $\mathcal{M}^\star$, however our results can easily be generalised to non-parametric methods like Gaussian process regression \citep{Rasmussen2006Gaussian,Wiener23,Krige1951}. A prior distribution over the parameter space $P_\Theta$ is specified, which represents the initial \emph{a priori} belief in the true value of $ P_{R,S}^\star( s,a)$ before the agent has observed any transitions. Priors are a powerful aspect of Bayesian RL, allowing practitioners to provide the agent with any information about the MDP and transfer knowledge between agents and domains. Given a history $h_t$, the prior is updated to a posterior $P_\Theta(h_t)$, representing the agent's beliefs in the MDP's dynamics once $h_t$ has been observed. For each history, the posterior is used to \emph{marginalise} across all hypotheses according to the agent's uncertainty, yielding the predictive state transition-reward distribution $P_{R,S}(h_t,a_t)=\mathbb{E}_{\theta\sim P_\Theta( h _t)} \left[P_{R,S}(s_{t},a_{t},\theta)\right] $ which characterise the epistemic and aleatoric uncertainty in $P_{R,S}^\star(s_t,a_t)$. Given $P_{R,S}(h_t,a_t)$, we  reason over counterfactual future trajectories using the predictive distribution over trajectories $P_{t}^\pi$
and define the BRL objective as:
\begin{align}
    J^\pi_\textrm{Bayes}(P_\Theta)\coloneqq \mathbb{E}_{h_\infty \sim P_{\infty}^\pi}\left[ \sum_{i=0}^\infty \gamma^{i} r_{i} \right]=\mathbb{E}_{\theta\sim P_\Theta}\left[
    \mathbb{E}_{h_\infty \sim P_{\infty}^\pi(\theta)}\left[ \sum_{i=0}^\infty \gamma^{i} r_{i} \right]\right].\label{eq:brl_objective}
\end{align}
Let $\Pi_\mathcal{H}$ denote the space of all history-conditioned policies. A corresponding optimal policy is known as a Bayes-optimal policy, which we denote as $\pi^\star_{\textrm{Bayes}}(\cdot)\in \Pi^\star_{\textrm{Bayes}}(P_\Theta)\coloneqq \argmax_{\pi\in\Pi_\mathcal{H}} J^\pi_\textrm{Bayes}(P_\Theta)$. Unlike in frequentist RL, Bayesian variables depend on histories obtained through posterior marginalisation; hence the posterior is often known as the \emph{belief state}, which augments each ground state $s_t$ like in a partially observable Markov decision process \citep{Drake62, Astrom65,Smallwood73,Kaelbling98}. Analogously to the state-transition distribution in frequentist RL, we define a \emph{belief transition} distribution $P_\mathcal{H}(h_t,a_t)$ using the predictive state transition-reward distribution, which yields \emph{Bayes-adaptive MDP} (BAMDP) \citep{Duff02}:
$
 \mathcal{M}_\textrm{Bayes}(P_\Theta)\coloneqq \langle\mathcal{H}, \mathcal{A}, P_0, P_\mathcal{H}(h,a), \gamma \rangle$. The BAMDP is solved using planning methods to obtain a Bayes-optimal policy, which naturally balances exploration with exploitation: after every timestep, the agent's uncertainty is characterised via the posterior conditioned on the history $h_{t}$, which includes all future trajectories to marginalise over. Via the belief transition, the BRL objective accounts for how the posterior evolves on every timestep, and hence any Bayes-optimal policy $\pi^\star_\textrm{Bayes}$ is optimal not only according to the epistemic uncertainty of a fixed belief but accounts for how the epistemic uncertainty evolves at every future timestep, decaying according to the discount factor. 

Let 
\begin{align}
    J^\pi(\theta)\coloneqq\mathbb{E}_{h_\infty \sim P_{\infty}^\pi(\theta)}\left[ \sum_{i=0}^\infty \gamma^{i} r_{i} \right],
\end{align}
denote the expected returns under policy $\pi$ for an MDP parametrised by $\theta$ and $J^\star(\theta)\coloneqq \sup_{\pi}J^\pi(\theta)$ denote the expected returns for under an optimal policy or an MDP parametrised by $\theta$. The Bayes regret of a policy $\pi$ is defined as:
\begin{align}
    \textrm{Regret}_\textrm{Bayes}^\pi(P_\Theta)&\coloneqq\mathbb{E}_{\theta \sim P_\Theta}\left[J^\star(\theta)-J^\pi(\theta)\right],\\
    &=\mathbb{E}_{\theta \sim P_\Theta}\left[J^\star(\theta)\right]-J^\pi_\textrm{Bayes}(P_\Theta). \label{eq:Bayes_regret}
\end{align}
which characterises how far a policy $\pi$ is from the true optimal policy taking an average over all possible MDPs using the prior $P_\Theta$. It is clear from \cref{eq:Bayes_regret} that any policy that minimises Bayes regret maximises the Bayesian RL objective $J^\pi_\textrm{Bayes}(P_\Theta)$.

\section{Derivation of Negative Log Likelihood Loss Function with Approximate Inference}
\label{app:log_likelihood_derivation}
Assume a dataset of $N$ input-output pairs:
\begin{align}
    \mathcal{D}_N\coloneqq \{(x_0,y_0),(x_1,y_1),\dots(x_{N-1},y_{N-1})\},
\end{align}
and a multivariate Gaussian regression model:
\begin{align}
    p(y\vert x,\theta)= \frac{1}{(2\pi)^\frac{D}{2}}\lvert \Sigma_\theta(x)\rvert^\frac{1}{2} \exp\left(-\frac{1}{2}(y-\mu_\theta(x))\Sigma_\theta^{-1}(y-\mu_\theta(x))^T\right),
\end{align}
where $D$ is the number of dimensions. Here our model $\text{NN}_\theta:\mathcal{X}\rightarrow \mathcal{P(Y)}$ is a neural network parametrised by $\theta\in\Theta$ that outputs a Gaussian distribution $\mathcal{N}(\mu_\theta,\Sigma_\theta)$ over $\mathcal{Y}$.
Assuming independent dimensions, such that the covariance matrix is diagonal:  
\begin{align}
    p(y\vert x,\theta)= \prod_{d=0}^{D-1} \frac{1}{\sqrt{2\pi\sigma_{\theta_d}^2(x)}}\exp\left(-\frac{1}{2\sigma_{\theta_d}^2(x)}(y_d-\mu_{\theta_d}(x))^2\right).
\end{align}
We can then fit our model by minimising the negative log likelihood loss: 
\begin{align}
    \mathcal{L} (\textrm{NLL}(\theta))\\
    \coloneqq& -\log p(\mathcal{D}_N|\theta),\\
    =& \sum_{i=0}^{N-1}\left(\frac{D}{2}\log(2\pi)+\frac{1}{2}\sum_{d=0}^{D-1}\left(\log \sigma_{\theta_d}^2(x_i) + \frac{(y_{i_d}-\mu_{\theta_d}(x_i))^2}{\sigma_{\theta_d}^2(x_i)}\right) \right),\\
    =&  \left[ \sum_{i=0}^{N-1}\frac{1}{N}\left(\frac{D}{2}\log(2\pi)+\frac{1}{2}\sum_{d=0}^{D-1}\left(\log \sigma_{\theta_d}^2(x_i) + \frac{(y_{i_d}-\mu_{\theta_d}(x_i))^2}{\sigma_{\theta_d}^2(x_i)}\right) \right) \right],\\
    =&  \mathbb{E}_{i\sim\mathcal{U}_N} \left[\frac{D}{2}\log(2\pi)+\frac{1}{2}\sum_{d=0}^{D-1}\left(\log \sigma_{\theta_d}^2(x_i) + \frac{(y_{i_d}-\mu_{\theta_d}(x_i))^2}{\sigma_{\theta_d}^2(x_i)}\right) \right],\\
    \overset{c}{=}& \mathbb{E}_{i\sim\mathcal{U}_N} \left[\sum_{d=0}^{D-1}\left(\log \sigma_{\theta_d}^2(x_i) + \frac{(y_{i_d}-\mu_{\theta_d}(x_i))^2}{\sigma_{\theta_d}^2(x_i)}\right) \right],
\end{align}
where recall $\mathcal{U}_N$ is the uniform distribution over $\{0,1,\dots N-1\}$. Our final line means equality up to a constant, as we can ignore the $\frac{D}{2}\log(2\pi)$ term for optimisation because it is independent of $\theta$. 

We use RP ensembles for our approximate posterior \citep{Osband18,Ciosek2020}; here an ensemble of $M$ separate model weights $\{\theta_0,\theta_i,\dots\theta_{M-1}\}$ are randomly initialised and are optimised in parallel, summing over the corresponding negative log likelihoods. When training, we optimise the log-variance rather than the variance for numerical stability and to ensure that the variance remains positive. This allows us to simultaneously optimise maximum and minimum log-variance parameters for each dimension across the ensemble, which we use to soft-clamp the log-variances output by individual models, preventing any individual model becoming overly confident or too uncertain in one dimension. Our final loss function is then given by:
\begin{align}
    \mathcal{L} (\theta, \mathcal{D}_N)& = \sum_{j=0}^{M-1}\left( \mathbb{E}_{i\sim\mathcal{U}_N} \left[\sum_{d=0}^{D-1}\left(\xi_{\theta_{d_j}}(x_i) + \frac{(y_{i_d}-\mu_{\theta_{d_j}}(x_i))^2}{\exp \left( \xi_{\theta_{d_j}}(x_i) \right)} \right)\right]\right) \\
    &\qquad+ c\cdot\sum_{d=0}^{D-1} \left( \xi_{\theta_{d_{\text{max}}}} - \xi_{\theta_{d_{\text{min}}}}\right),
\end{align}
where $\xi_{\theta_{d_j}} = \xi_{\theta_{d_{\text{min}}}} + \left[1+\exp \left(\xi_{\theta_{d_{\text{max}}}} - \left[1+\exp \left(\log\sigma_{\theta_{d_j}}^2 - \xi_{\theta_{d_{\text{max}}}}\right) \right] -\xi_{\theta_{d_{\text{min}}}}\right)\right]$ and $\xi_{\theta_{d_{\text{min}}}}$ and $\xi_{\theta_{d_{\text{max}}}}$ are respectively the minimum and maximum log-variance parameters optimised across the ensemble, $c$ is the log-variance difference coefficient used to control the clamping term, and $M$ is the number of models in the ensemble. 
$\mathcal{L}(\theta, \mathcal{D}_N)$ can be minimised by using Monte Carlo minibatch gradient descent with a minibatch $\mathcal{M}_n$ of $n<N$ samples drawn uniformly from $\mathcal{D}_N$.

\section{Value Estimators}
\label{sec:regret_approximators}

To estimate the value of the Bayes-optimal policy using only offline data, we sample dynamics models from the posterior (approximated via an ensemble), roll out $\pi_\textrm{Bayes}^\star$ in each sampled model, and compute the corresponding discounted returns. This yields predictive returns
\begin{align}
    \{ R_\theta^{\pi_\textrm{Bayes}^\star} \}_{\theta \sim P_\Theta(\mathcal{D}_N)},
\end{align}
which approximate the posterior predictive distribution over the policy value. From these samples, we construct estimators of $J^{\pi_\textrm{Bayes}^\star}(\mathcal{M}^\star)$.

Specifically, we define the predictive mean:
\begin{align}
    \hat{J}^{\pi_\textrm{Bayes}^\star}_{\mathrm{mean}}(\mathcal{M}^\star)
    = \mathbb{E}_{\theta \sim P_\Theta(\mathcal{D}_N)}
    \left[ R_\theta^{\pi_\textrm{Bayes}^\star} \right],
\end{align}
the predictive median:
\begin{align}
    \hat{J}^{\pi_\textrm{Bayes}^\star}_{\mathrm{median}}(\mathcal{M}^\star)
    = \mathbb{M}_{\theta \sim P_\Theta(\mathcal{D}_N)}
    \left[ R_\theta^{\pi_\textrm{Bayes}^\star} \right],
\end{align}
the predictive maximum:
\begin{align}
    \hat{J}^{\pi_\textrm{Bayes}^\star}_{\max}(\mathcal{M}^\star)
    = \max_{\theta \sim P_\Theta(\mathcal{D}_N)}
    R_\theta^{\pi_\textrm{Bayes}^\star},
\end{align}
and the predictive minimum
\begin{align}
    \hat{J}^{\pi_\textrm{Bayes}^\star}_{\min}(\mathcal{M}^\star)
    = \min_{\theta \sim P_\Theta(\mathcal{D}_N)}
    R_\theta^{\pi_\textrm{Bayes}^\star}.
\end{align}

These estimators reflect different attitudes toward posterior uncertainty. The maximum is optimistic and can overestimate performance if even a single posterior sample is favourable, while the minimum is conservative and may underestimate performance when a subset of posterior samples are pessimistic. The predictive mean aggregates all samples but can be sensitive to extreme values. Empirically, we find that the predictive median provides the most stable trade-off: it is robust to outliers, avoids undue optimism driven by a small number of favourable models, and is less conservative than the minimum. In practice, 
\(
\hat{J}^{\pi_\textrm{Bayes}^\star}_{\mathrm{median}}(\mathcal{M}^\star)
\)
tracks the true online value closely across domains, making it a reliable fully offline estimator.

\section{Proofs}
\label{app:proofs}
\subsection{Primer on Total Variational Distance}\label{sec:math_prelim_app}
We measure distance between two probability distributions $P_X$ and $Q_X$ using the total variational (TV) distance, defined as:
\begin{align}
\textrm{TV}(P_X\Vert Q_X)&\coloneqq \sup_{E}\left\lvert P_X(E)- Q_X(E)\right\rvert.
\end{align}
The TV distance takes the supremum over all events $E$ to find the event that gives rise to the maximum difference in probability between two distributions. A key property of the TV distance is:  $0\le\textrm{TV}(P_X\Vert Q_X)\le1$. If $\textrm{TV}(P_X\Vert Q_X)=0$, then $P_X = Q_X$ as there is no event that both distributions don't assign the same probability mass to. If $\textrm{TV}(P_X\Vert Q_X)=1$, then the distributions assign completely different mass to at least one event. The TV distance can be related to the Kullback-Leibler (KL) divergence using the  Bretagnolle-Huber \citep{Bretagnolle1978} inequality: $\textrm{TV}(P_X\Vert Q_X)\le \sqrt{1-\exp\left(- \textrm{KL}(P_X\Vert Q_X)\right)}\le 1$, which preserves the property that  $0\le\textrm{TV}(P_X\Vert Q_X)\le1$. The TV distance can be shown \citep{Sriperumbudur09} to be equivalent to the integral probability metric under the $\infty$-norm, which we will make use of in our theorems:
\begin{align}
\textrm{TV}(P_X\Vert Q_X)&=\frac{1}{2}\sup_{f\in\mathcal{F}:\mathcal{X}\rightarrow[-1,1]}\left\lvert\mathbb{E}_{x\sim P_X}\left[f(x)\right]- \mathbb{E}_{x\sim Q_X}\left[f(x)\right]\right\rvert,\label{eq:ipm_form}
\end{align}
In this form, the supremum is taken over the space of all functions that are bounded by unity, that is $\lVert f\rVert_\infty=1$.

\setcounter{theorem}{0}
\setcounter{proposition}{0}
\subsection{Proof of \cref{proof:regret_bound_kl}}
\label{app:proof_regret_kl}
Let the predictive distribution  over history $h_t$ using the posterior $P_\Theta(\mathcal{D}_N)$ be $P_{t,\pi}(\mathcal{D}_N)$, which has density:
\begin{align}
    p_\pi(h_t,\mathcal{D}_N)\coloneqq p_0(s_0) \prod_{i=0}^{t-1}\pi(a_i\vert h_i) p(r_i\vert h_i,a_i,\mathcal{D}_N)p(s_{i+1}\vert h_i,a_i,\mathcal{D}_N).
\end{align}
According to the Bernstein-von Mises theorem \citep{doob49,LeCam53, vaart98}, as the posterior becomes more informative it concentrates around a smaller (and more tractable) subset of the hypothesis space. Not only does this ease the computational burden of solving the BRL objective, but in the limit $N\rightarrow \infty$, the Bayesian RL objective using the true posterior will approach the true expected discounted return for the MDP: $J^{\pi}(P_\Theta(\mathcal{D}_N))\xrightarrow[N\rightarrow\infty]{}J^\pi(\mathcal{M}^\star)$. In this limit, any Bayes-optimal policy will be an optimal policy for the true MDP, achieving the highest expected returns once deployed. Our first lemma bounds the difference between the expected true rewards after time $i$ and the predictive expected rewards in terms of total variational distance:

\begin{lemma} \label{proof:reward_bound_lemma}For bounded reward functions:
\begin{align}
    \left\lvert\mathbb{E}_{h_{i+1}\sim P^\star_{i+1,\pi}}\left[r_i\right]-\mathbb{E}_{h_{i+1}\sim P^\pi_{i+1}(\mathcal{D}_N)}\left[r_i\right]\right\rvert\le (r_{\max}-r_{\min})\cdot\textnormal{\textrm{TV}}\left(P^\star_{i+1,\pi}\Vert P^\pi_{i+1}(\mathcal{D}_N) \right).\label{eq:TV_reward_bound}
\end{align}
    \begin{proof}
We start by subtracting and adding $\frac{r_{\max}+r_{\min}}{2}$ to the left hand side of Ineq.~\ref{eq:TV_reward_bound}:
\begin{align}
     &\lvert\mathbb{E}_{h_{i+1}\sim P^\star_{i+1,\pi}}\left[r_i\right]-\mathbb{E}_{h_{i+1}\sim P^\pi_{i+1}(\mathcal{D}_N)}\left[r_i\right]\rvert\\
     &=\left\lvert\mathbb{E}_{h_{i+1}\sim P^\star_{i+1,\pi}}\left[r_i-\frac{r_{\max}+r_{\min}}{2}\right]-\mathbb{E}_{h_{i+1}\sim P^\pi_{i+1}(\mathcal{D}_N)}\left[r_i-\frac{r_{\max}+r_{\min}}{2}\right]\right\rvert,\\
     &=\left\lvert\mathbb{E}_{h_{i+1}\sim P^\star_{i+1,\pi}}\left[\frac{2r_i-(r_{\max}+r_{\min})}{2}\right]-\mathbb{E}_{h_{i+1}\sim P^\pi_{i+1}(\mathcal{D}_N)}\left[\frac{2r_i-(r_{\max}+r_{\min})}{2}\right]\right\rvert,\\
     &=\frac{(r_{\max}-r_{\min})}{2}\cdot\left\lvert\mathbb{E}_{h_{i+1}\sim P^\star_{i+1,\pi}}\left[\frac{2r_i-(r_{\max}+r_{\min})}{r_{\max}-r_{\min}}\right]-\mathbb{E}_{h_{i+1}\sim P^\pi_{i+1}(\mathcal{D}_N)}\left[\frac{2r_i-(r_{\max}+r_{\min})}{r_{\max}-r_{\min}}\right]\right\rvert,\\
     &=\frac{(r_{\max}-r_{\min})}{2}\cdot\left\lvert\mathbb{E}_{h_{i+1}\sim P^\star_{i+1,\pi}}\left[r_\textrm{norm}(h_{i+1})\right]-\mathbb{E}_{h_{i+1}\sim P^\pi_{i+1}(\mathcal{D}_N)}\left[r_\textrm{norm}(h_{i+1})\right]\right\rvert,\label{eq:function_normalised_form}
\end{align}
 where:
  \begin{align}
      r_\textrm{norm}(h_{i+1})\coloneqq\frac{2r_i-(r_{\max}+r_{\min})}{r_{\max}-r_{\min}}.
  \end{align}
Now, as $r_\textrm{norm}:\mathcal{H}_{i+1}\rightarrow[-1,1]$, we can bound  \cref{eq:function_normalised_form} using the integral probability metric form of the TV distance (see \cref{eq:ipm_form}), yielding our desired result:
\begin{align}
    &\lvert\mathbb{E}_{h_{i+1}\sim P^\star_{i+1,\pi}}\left[r_i\right]-\mathbb{E}_{h_{i+1}\sim P^\pi_{i+1}(\mathcal{D}_N)}\left[r_i\right]\rvert\\
     &\le\frac{r_{\max}-r_{\min}}{2}\cdot\sup_{f\in\mathcal{F}:\mathcal{H}_{i+1}\rightarrow[-1,1]}\left\lvert\mathbb{E}_{h_{i+1}\sim P^\star_{i+1,\pi}}\left[f(h_{i+1})\right]-\mathbb{E}_{h_{i+1}\sim P^\pi_{i+1}(\mathcal{D}_N)}\left[f(h_{i+1})\right]\right\rvert,\\
     &=(r_{\max}-r_{\min})\cdot\textnormal{\textrm{TV}}\left(P^\star_{i+1,\pi}\Vert P^\pi_{i+1}(\mathcal{D}_N) \right).
\end{align}
    \end{proof}                             
\end{lemma}

Using this result, we bound  the true regret $\textnormal{\textrm{Regret}}(\mathcal{M}^\star,\mathcal{D}_N)$ using the TV distance between the true $P_{i,\pi}^\star$ and predictive $P_{i,\pi}(\mathcal{D}_N)$ history distributions:

\begin{lemma} \label{proof:regret_bound_app} Let $\mathcal{R}_{\max}\coloneqq \frac{(r_\textrm{max}-r_\textrm{min})}{1-\gamma}$ denote the maximum possible regret for the MDP. For a prior $P_\Theta(\mathcal{D}_N)$, the true regret can be bounded as:
\begin{align}
\textnormal{\textrm{Regret}}(\mathcal{M}^\star,\mathcal{D}_N)&\le2\mathcal{R}_{\max}\cdot \sup_\pi \mathbb{E}_{i\sim\mathcal{G}(\gamma)}\left[\textrm{TV}\left(P_{i+1,\pi}^\star\Vert P_{i+1}^\pi( \mathcal{D}_N)\right)\right].\label{eq:true_regret_bound}
\end{align}
\begin{proof}
We start from the definition of the true regret:
\begin{align}
    \textnormal{\textrm{Regret}}(\mathcal{M}^\star,\mathcal{D}_N)\coloneqq &J^{\pi^\star}(\mathcal{M}^\star)- J^{\pi_\textrm{Bayes}^\star}(\mathcal{M}^\star,\mathcal{D}_N).
    \end{align}
We now bound the difference between $J^{\pi^\star}(\mathcal{M}^\star)$ and $J^{\pi_\textrm{Bayes}^\star}(\mathcal{M}^\star,\mathcal{D}_N)$ in terms of the difference between $J^{\pi}(\mathcal{M}^\star)$ and $J^{\pi}_\textrm{Bayes}(\mathcal{D}_N)$:
        \begin{align}
 &\textnormal{\textrm{Regret}}(\mathcal{M}^\star,\mathcal{D}_N)\\
 &=J^{\pi^\star}(\mathcal{M}^\star)-J^{\pi^\star}_\textrm{Bayes}(P_\Phi(\mathcal{D}_N))+J^{\pi^\star}_\textrm{Bayes}(P_\Phi(\mathcal{D}_N))-J^{\pi_\textrm{Bayes}^\star}(\mathcal{M}^\star,\mathcal{D}_N),\\
   &\le J^{\pi^\star}(\mathcal{M}^\star)-J^{\pi^\star}_\textrm{Bayes}(P_\Phi(\mathcal{D}_N))+J^{\pi^\star_\textrm{Bayes}}_\textrm{Bayes}(P_\Phi(\mathcal{D}_N))-J^{\pi_\textrm{Bayes}^\star}(\mathcal{M}^\star,\mathcal{D}_N),\\
    &\le \sup_\pi \left\lvert J^{\pi}(\mathcal{M}^\star)-J^\pi_\textrm{Bayes}(P_\Theta(\mathcal{D}_N))\right\rvert+\sup_\pi \left\lvert J^\pi_\textrm{Bayes}(P_\Theta(\mathcal{D}_N)) -J^{\pi}(\mathcal{M}^\star)\right\rvert,\\
    &=2 \sup_\pi \left\lvert  J^{\pi}(\mathcal{M}^\star)-J^\pi_\textrm{Bayes}(P_\Theta(\mathcal{D}_N))\right\rvert, \label{eq:true_regret_bayes}
\end{align}
where the second line follows from $J^{\pi^\star}_\textrm{Bayes}(P_\Phi(\mathcal{D}_N))\le J^{\pi^\star_\textrm{Bayes}}_\textrm{Bayes}(P_\Phi(\mathcal{D}_N))$ by definition. Now our goal is to bound $\left\lvert J^{\pi}(\mathcal{M}^\star)-J^\pi_\textrm{Bayes}(P_\Theta(\mathcal{D}_N))\right\rvert$:
\begin{align}
\left\lvert J^{\pi}(\mathcal{M}^\star)-J^\pi_\textrm{Bayes}(P_\Theta(\mathcal{D}_N))\right\rvert&=\left\lvert\mathbb{E}_{h_\infty\sim P^\star_{\infty,\pi}}\left[\sum_{i=0}^\infty \gamma^ir_i\right]-\mathbb{E}_{h_\infty\sim P^\pi_\infty(\mathcal{D}_N)}\left[\sum_{i=0}^\infty \gamma^ir_i\right]\right\rvert,\\
&=\left\lvert\sum_{i=0}^\infty \gamma^i\mathbb{E}_{h_{i+1}\sim P^\star_{i+1,\pi}}\left[r_i\right]-\sum_{i=0}^\infty \gamma^i\mathbb{E}_{h_{i+1}\sim P^\pi_{i+1}(\mathcal{D}_N)}\left[r_i\right]\right\rvert,\\
&=\left\lvert\sum_{i=0}^\infty \gamma^i\left(\mathbb{E}_{h_{i+1}\sim P^\star_{i+1,\pi}}\left[r_i\right]-\mathbb{E}_{h_{i+1}\sim P^\pi_{i+1}(\mathcal{D}_N)}\left[r_i\right]\right)\right\rvert,\\
&\le\sum_{i=0}^\infty \gamma^i\left\lvert\mathbb{E}_{h_{i+1}\sim P^\star_{i+1,\pi}}\left[r_i\right]-\mathbb{E}_{h_{i+1}\sim P^\pi_{i+1}(\mathcal{D}_N)}\left[r_i\right]\right\rvert.
\end{align}
Using Ineq.~\ref{eq:TV_reward_bound} from \cref{proof:reward_bound_lemma}, we now bound each difference $\left\lvert\mathbb{E}_{h_{i+1}\sim P^\star_{i+1,\pi}}\left[r_i\right]-\mathbb{E}_{h_{i+1}\sim P^\pi_{i+1}(\mathcal{D}_N)}\left[r_i\right]\right\rvert$ in terms of total variational distance between $P^\star_{i+1,\pi}$ and $P^\pi_{i+1}(\mathcal{D}_N)$:
\begin{align}
    \left\lvert J^{\pi}(\mathcal{M}^\star)-J^\pi_\textrm{Bayes}(P_\Theta(\mathcal{D}_N))\right\rvert&\le(r_{\max}-r_{\min})\cdot\sum_{i=0}^\infty \gamma^i\textnormal{\textrm{TV}}\left(P^\star_{i+1,\pi}\Vert P^\pi_{i+1}(\mathcal{D}_N) \right),\\
    &=\frac{r_{\max}-r_{\min}}{1-\gamma}\cdot\sum_{i=0}^\infty (1-\gamma)\gamma^i\textnormal{\textrm{TV}}\left(P^\star_{i+1,\pi}\Vert P^\pi_{i+1}(\mathcal{D}_N) \right),\\
    &=\frac{r_{\max}-r_{\min}}{1-\gamma}\cdot\mathbb{E}_{i\sim \mathcal{G}(\gamma)}\left[\textnormal{\textrm{TV}}\left(P^\star_{i+1,\pi}\Vert P^\pi_{i+1}(\mathcal{D}_N) \right)\right],\\
    &=\mathcal{R}_{\max}\cdot \mathbb{E}_{i\sim \mathcal{G}(\gamma)}\left[\textnormal{\textrm{TV}}\left(P^\star_{i+1,\pi}\Vert P^\pi_{i+1}(\mathcal{D}_N) \right)\right],
\end{align}
where $\mathcal{G}(\gamma)$ is the geometric distribution. Finally, substituting into Ineq.~\ref{eq:true_regret_bayes} yields our desired result:
\begin{align}
    \textnormal{\textrm{Regret}}(\mathcal{M}^\star,\mathcal{D}_N)&\le 2 \sup_\pi \left\lvert (\mathcal{R}_{\max}\cdot\mathbb{E}_{i\sim\mathcal{G}(\gamma)}\left[ \textrm{TV}\left(P^\star_{i+1,\pi}\Vert P^\pi_{i+1}(\mathcal{D}_N))\right)\right]\right\rvert,\\
     &=2\mathcal{R}_{\max}\cdot \sup_\pi \mathbb{E}_{i\sim\mathcal{G}(\gamma)}\left[ \textrm{TV}\left(P^\star_{i+1,\pi}\Vert P^\pi_{i+1}(\mathcal{D}_N))\right)\right].
\end{align}

\end{proof}
\end{lemma}
We remark that \cref{proof:regret_bound_app} holds for any general reward-transition model given bounded rewards. The bound in Ineq.~\ref{eq:true_regret_bound} proves the true regret is governed by the geometric average of TV distances: $\mathbb{E}_{i\sim \mathcal{G}(\gamma)}\left[\textnormal{TV}\left(P^\star_{i+1,\pi}\Vert P^\pi_{i+1}(\mathcal{D}_N)) \right)\right]$. As each term $\textnormal{TV}\left(P^\star_{i+1,\pi}\Vert P^\pi_{i+1}(\mathcal{D}_N)) \right)$ measures the distance between the true  and predictive distributions over history $h_i$ of length $i$, the discounting factor $\gamma$ determines how much long term histories contribute to regret.

Intuitively, the more mass the posterior places close to the true value $\theta^\star\in \Theta^\star$, the smaller each TV distance becomes, with regret tending to zero for $P_{i+1,\pi}(\mathcal{D}_N)\approx  P_{i+1,\pi}^\star\implies \textnormal{TV}\left(P^\star_{i+1,\pi}\Vert P^\pi_{i+1}(\mathcal{D}_N)) \right)\approx 0$. Conversely, a strong but highly incorrect prior will concentrate mass around MDPs whose dynamics oppose the true dynamics, yielding $\textnormal{TV}\left(P^\star_{i+1,\pi}\Vert P^\pi_{i+1}(\mathcal{D}_N))\right)\approx 1$ for all $i$, achieving the highest possible regret: $\mathcal{R}_{\max}\coloneqq \nicefrac{(r_\textrm{max}-r_\textrm{min})}{1-\gamma}$. The resulting Bayes-optimal policy would choose actions that encourage negative reward-seeking behaviour, being farthest from optimal in terms of expected returns.

Using the Bretagnolle-Huber inequality (see \cref{sec:math_prelim_app}), we now relate the sum of discounted TV distances to a sum of KL divergences, allowing us to control the expected regret using the PIL.

\begin{theorem} \label{app:proof_PIL} Using the PIL: $\mathcal{I}_N^\pi$, the true regret is bounded as:
\begin{align}
\textnormal{\textrm{Regret}}(\mathcal{M}^\star,\mathcal{D}_N)\le2 \mathcal{R}_{\max}\cdot\sup_\pi\sqrt{1-\exp\left(-\frac{\mathcal{I}_N^\pi}{1-\gamma}\right)}\label{eq:kl_regret_bound}
\end{align}
    \begin{proof}
    Starting with the bounded derived in Ineq.~\ref{eq:true_regret_bound} of \cref{proof:regret_bound_app}, we apply the Bretagnolle-Huber inequality \citep{Bretagnolle1978} (see \cref{sec:math_prelim_app}) to bound the TV distance terms using the KL divergence:
        \begin{align}
\textnormal{\textrm{Regret}}(\mathcal{M}^\star,\mathcal{D}_N)&\le2\mathcal{R}_{\max}\cdot \sup_\pi \mathbb{E}_{i\sim\mathcal{G}(\gamma)}\left[ \textrm{TV}\left(P^\star_{i+1,\pi}\Vert P^\pi_{i+1}(\mathcal{D}_N))\right)\right],\\
&\le2\mathcal{R}_{\max}\cdot \sup_\pi \mathbb{E}_{i\sim\mathcal{G}(\gamma)}\left[\sqrt{1-\exp\left(-\textrm{KL}\left(P^\star_{i+1,\pi}\Vert P^\pi_{i+1}(\mathcal{D}_N))\right)\right)}\right].\label{eq:bh_bound}
        \end{align}
        We make two observations. Firstly, as the KL divergence is convex in its second argument and $P_{i+1,\pi}(\mathcal{D}_N)=\mathbb{E}_{\theta\sim P_\Theta(\mathcal{D}_N)}\left[P_{i+1}^\pi(\theta)\right]$, we can bound each KL divergence term using Jensen's inequality as:
        \begin{align}
            \textrm{KL}\left(P^\star_{i+1,\pi}\Vert P^\pi_{i+1}(\mathcal{D}_N))\right)\le \mathbb{E}_{\theta\sim P_\Theta(\mathcal{D}_N)}\left[\textrm{KL}\left(P_{i+1,\pi}^\star\Vert P_{i+1}^\pi(\theta)\right)\right].
        \end{align}
    Secondly, as the function $f(x)=\sqrt{1-\exp(-x)}$ is monotonically increasing in $x$, it follows that  $f(x)\le f(x')$ for any $x\le x'$, hence:
        \begin{align}
            \sqrt{1-\exp\left(-\textrm{KL}\left(P^\star_{i+1,\pi}\Vert P^\pi_{i+1}(\mathcal{D}_N))\right)\right)}\le \sqrt{1-\exp\left(-\mathbb{E}_{\theta\sim P_\Theta(\mathcal{D}_N)}\left[\textrm{KL}\left(P_{i+1,\pi}^\star\Vert P_{i+1}^\pi(\theta)\right)\right]\right)}
        \end{align}
    Applying this bound to Ineq.~\ref{eq:bh_bound} yields: 
    \begin{align}
        \mathbb{E}_{i\sim\mathcal{G}(\gamma)}&\left[ \sqrt{1-\exp\left(-\textrm{KL}\left(P^\star_{i+1,\pi}\Vert P^\pi_{i+1}(\mathcal{D}_N))\right)\right)}\right]\\
        &\le\mathbb{E}_{i\sim\mathcal{G}(\gamma)}\left[\sqrt{1-\exp\left(-\mathbb{E}_{\theta\sim P_\Theta(\mathcal{D}_N)}\left[\textrm{KL}\left(P_{i+1,\pi}^\star\Vert P_{i+1}^\pi(\theta)\right)\right]\right)} \right].
    \end{align}
As the function $f(x)=\sqrt{1-\exp(-x)}$ is concave in $x$, we can apply Jensen's inequality, yielding: 
        \begin{align}
             \mathbb{E}_{i\sim\mathcal{G}(\gamma)}&\left[ \sqrt{1-\exp\left(-\textrm{KL}\left(P^\star_{i+1,\pi}\Vert P^\pi_{i+1}(\mathcal{D}_N))\right)\right)}\right]\\
            &\le\sqrt{1-\exp\left(-\mathbb{E}_{i\sim\mathcal{G}(\gamma)}\left[\mathbb{E}_{\theta\sim P_\Theta(\mathcal{D}_N)}\left[\textrm{KL}\left(P_{i+1,\pi}^\star\Vert P_{i+1}^\pi(\theta)\right)\right]\right]\right)} .\label{eq:jensens}
        \end{align}
        Examining the KL divergence term: 
        \begin{align}
        \textrm{KL}&\left(P^\star_{i+1,\pi}\Vert P^\pi_{i+1}(\theta) \right)\\
        &=\mathbb{E}_{h_{i+1}\sim P_{i+1,\pi}^\star}\left[\log\left( \frac{d_0(s_0)\prod_{j=0}^{i}\pi(a_j\vert h_j) p^\star(r_j\vert s_j,a_j)p^\star(s_{j+1}\vert s_j,a_j)}{d_0(s_0)\prod_{j=0}^{i}\pi(a_j\vert h_j) p(r_j\vert s_j,a_j,\theta)p(s_{j+1}\vert s_j,a_j,\theta)} \right)\right],\\
        &=\mathbb{E}_{h_{i+1}\sim P_{i+1,\pi}^\star}\left[\log\left( \frac{\prod_{j=0}^{i} p^\star(r_j\vert s_j,a_j)p^\star(s_{j+1}\vert s_j,a_j)}{\prod_{j=0}^{i} p(r_j\vert s_j,a_j,\theta)p(s_{j+1}\vert s_j,a_j,\theta)} \right)\right],\\
        &=\mathbb{E}_{h_{i+1}\sim P_{i+1,\pi}^\star}\Bigg[\sum_{j=0}^{i}\Big(\log p^\star(r_j\vert s_j,a_j) -\log p(r_j\vert s_j,a_j,\theta) \\
        &\qquad\qquad\qquad\quad +\log p^\star(s_{j+1}\vert s_j,a_j)-\log p(s_{j+1}\vert s_j,a_j,\theta)\Big)\Bigg],\\
        &=\sum_{j=0}^{i}\mathbb{E}_{h_j\sim P_{j,\pi}^\star}\Bigg[\Big(\log p^\star(r_j\vert s_j,a_j) -\log p(r_j\vert s _j,a_j,\theta) \\
        &\qquad\qquad\qquad\quad +\log p^\star(s_{j+1}\vert s_j,a_j)-\log p(s_{j+1}\vert s_j,a_j,\theta)\Big)\Bigg],\\
        &=\sum_{j=0}^{i}\mathbb{E}_{s_j,a_j\sim P^\star_{j,\pi}}\Bigg[\mathbb{E}_{r_j,s_{j+1}\sim P_{R,S}^\star(s_j,a_j)}\Bigg[\Big(\log p^\star(r_j\vert s_j,a_j) -\log p(r_j\vert s_j,a_j,\theta) \\
        &\qquad\qquad\qquad\quad +\log p^\star(s_{j+1}\vert s_j,a_j)-\log p(s_{j+1}\vert s_j,a_j,\theta)\Big)\Bigg]\Bigg],\\
        &=\sum_{j=0}^{i}\mathbb{E}_{s_j,a_j\sim P^\star_{j,\pi}}\Bigg[\mathbb{E}_{r_j,s_{j+1}\sim P_{R,S}^\star(s_j,a_j)}\Bigg[\Big(\log p^
        \star(r_j,s_{j+1}\vert s_j,a_j)\\
        &\qquad\qquad\qquad\quad-\log p(r_j,s_{j+1}\vert s_j,a_j,\theta) \Bigg]\Bigg],\\
        &=\sum_{j=0}^{i}\mathbb{E}_{s,a\sim P^\star_{j,\pi}}\left[\textrm{KL}( P_{R,S}^\star(s,a)\Vert P_{R,S}(s,a,\theta))\right],
        \end{align}
        hence:
        \begin{align}
            &\mathbb{E}_{i\sim\mathcal{G}(\gamma)}\left[\mathbb{E}_{\theta\sim P_\Theta(\mathcal{D}_N)}\left[\textrm{KL}\left(P_{i+1,\pi}^\star\Vert P_{i+1}^\pi(\theta)\right)\right]\right]\\
            &=\mathbb{E}_{\theta\sim P_\Theta(\mathcal{D}_N)}\left[\mathbb{E}_{i\sim\mathcal{G}(\gamma)}\left[\sum_{j=0}^{i}\mathbb{E}_{s,a\sim P^\star_{j,\pi}}\left[\textrm{KL}( P_{R,S}^\star(s,a)\Vert P_{R,S}(s,a,\theta))\right]\right]\right],\\
            &=\mathbb{E}_{\theta\sim P_\Theta(\mathcal{D}_N)}\left[\sum_{i=0}^\infty  (1-\gamma)\gamma^i\sum_{j=0}^{i}\mathbb{E}_{s,a\sim P^\star_{j,\pi}}\left[\textrm{KL}( P_{R,S}^\star(s,a)\Vert P_{R,S}(s,a,\theta))\right]\right],\\
            &=\mathbb{E}_{\theta\sim P_\Theta(\mathcal{D}_N)}\left[\sum_{i=0}^\infty  (1-\gamma)\gamma^i(i+1)\sum_{j=0}^{i}\frac{1}{i+1}\mathbb{E}_{s,a\sim P^\star_{j,\pi}}\left[\textrm{KL}( P_{R,S}^\star(s,a)\Vert P_{R,S}(s,a,\theta))\right]\right],\\
            &=\frac{1}{1-\gamma}\mathbb{E}_{\theta\sim P_\Theta(\mathcal{D}_N)}\left[\sum_{i=0}^\infty  (1-\gamma)^2\gamma^i(i+1)\mathbb{E}_{j\sim \mathcal{U}_i}\left[\mathbb{E}_{s,a\sim P^\star_{j,\pi}}\left[\textrm{KL}( P_{R,S}^\star(s,a)\Vert P_{R,S}(s,a,\theta))\right]\right]\right],\\
            &=\frac{1}{1-\gamma}\mathbb{E}_{\theta\sim P_\Theta(\mathcal{D}_N)}\left[\mathbb{E}_{i\sim \mathcal{AG(\gamma)}}\left[\mathbb{E}_{j\sim \mathcal{U}_i}\left[\mathbb{E}_{s,a\sim P^\star_{j,\pi}}\left[\textrm{KL}( P_{R,S}^\star(s,a)\Vert P_{R,S}(s,a,\theta))\right]\right]\right]\right].\label{eq:KL_sum_form}
        \end{align}
        Now, as $\rho_\pi^\star=\mathbb{E}_{i\sim \mathcal{AG(\gamma)}}\left[\mathbb{E}_{j\sim \mathcal{U}_i}\left[ P_{j,\pi}^\star\right]\right]$ is the arithemetico-geometric ergodic state-action distribution, we can simplify \cref{eq:KL_sum_form} to yield:
        \begin{align}
            &\mathbb{E}_{i\sim\mathcal{G}(\gamma)}\left[\mathbb{E}_{\theta\sim P_\Theta(\mathcal{D}_N)}\left[\textrm{KL}\left(P_{i+1,\pi}^\star\Vert P_{i+1}^\pi(\theta)\right)\right]\right] \\
            &=\frac{1}{1-\gamma}\mathbb{E}_{\mathcal{D}_N\sim P_{\textnormal{\textrm{Data}}}}\left[\mathbb{E}_{\theta\sim P_\Theta(\mathcal{D}_N)}\left[\mathbb{E}_{s,a\sim \rho_\pi^\star}\left[\textrm{KL}( P_{R,S}^\star(s,a)\Vert P_{R,S}(s,a,\theta))\right]\right]\right],\\
            &=\frac{1}{1-\gamma}\mathbb{E}_{\mathcal{D}_N\sim P_{\textnormal{\textrm{Data}}}}\left[\mathbb{E}_{s,a\sim \rho_\pi^\star}\left[\mathbb{E}_{\theta\sim P_\Theta(\mathcal{D}_N)}\left[\textrm{KL}( P_{R,S}^\star(s,a)\Vert P_{R,S}(s,a,\theta))\right]\right]\right],\\
            &=\frac{1}{1-\gamma}\mathcal{I}^\pi_N,
        \end{align}
        hence, substituting into Ineq.~\ref{eq:jensens}, we obtain: 
        \begin{align}
            \mathbb{E}_{i\sim\mathcal{G}(\gamma)}&\left[ \sqrt{1-\exp\left(-\textrm{KL}\left(P^\star_{i+1,\pi}\Vert P^\pi_{i+1}(\mathcal{D}_N))\right)\right)}\right]\le \sqrt{1-\exp\left(-\frac{\mathcal{I}^\pi_N}{1-\gamma}\right)}.
        \end{align}
        Finally, substituting into \cref{eq:bh_bound} yields our desired result:
        \begin{align}
        \textnormal{\textrm{Regret}}(\mathcal{M}^\star,\mathcal{D}_N)&\le2\mathcal{R}_{\max}\cdot \sup_\pi\left\lvert \mathbb{E}_{i\sim\mathcal{G}(\gamma)}\left[\sqrt{1-\exp\left(-\textrm{KL}\left(P^\star_{i+1,\pi}\Vert P^\pi_{i+1}(\mathcal{D}_N))\right)\right)}\right]\right\rvert,\\
&\le 2\mathcal{R}_{\max}\cdot \sup_\pi\sqrt{1-\exp\left(-\frac{\mathcal{I}^\pi_N}{1-\gamma}\right)}.
        \end{align}
    \end{proof}
\end{theorem}
The PIL has an intuitive information-geometric interpretation: the inner expectation $\mathbb{E}_{s,a\sim \rho_\pi^\star}\left[ \textrm{KL}\left(P_{R,S}^\star(s,a)\Vert P_{R,S}(s,a,\theta) \right)\right]$ measures the distance between the model and the true distribution in terms of the information lost when approximating $P_{R,S}^\star(s,a)$ with $P_{R,S}(s,a,\theta)$, averaged across all states. The PIL thus measures how close the posterior's belief is to the truth according to the average information lost under the posterior expectation. We observe that via Jensen's inequality, the PIL is an upper bound on the classic KL risk (sometimes known as expected relative entropy) from Bayesian asymptotics and regret analysis \citep{Aitchson75, Clarke1990, Komaki96,Hartigan1998,barron1988exponential,Barron1999,yang99,vanderVaart11,Aslan06,Alaa18,Bilodeau2021}.

By substituting in our definition of the Gaussian world model, we now find a convenient form for the PIL:

\begin{proposition}  Using the Gaussian world model in \cref{eq:Gaussian_world_model}, it follows:
\begin{align}
    \mathcal{I}_N^\pi=\mathcal{E}(\mathcal{D}_N,\mathcal{M}^\star)+ \mathcal{V}(\mathcal{D}_N).
\end{align}
\begin{proof}
    We substitute the Gaussian world model into the KL divergence to yield: 
    \begin{align}
        \textrm{KL}&\left(P_{R,S}^\star(s,a)\Vert P_{R,S}(s,a,\theta) \right)\\
           =&\mathbb{E}_{r,s'\sim P_{R,S}^\star(s,a)}\left[\log \left(\exp\left(-\frac{\lVert r^\star(s,a) - r\rVert^2_2}{2\sigma^2_r}\right)\exp\left(-\frac{\lVert {s^\star}'(s,a) - s'\rVert^2_2}{2\sigma^2_s}\right)\right)\right]\\
        &   -\mathbb{E}_{r,s'\sim P_{R,S}^\star(s,a)}\left[\log\left(\exp\left(-\frac{\lVert r_\theta(s,a) - r\rVert^2_2}{2\sigma^2_r}\right)\exp\left(-\frac{\lVert s_{\theta}'(s,a) - s'\rVert^2_2}{2\sigma^2_s}\right)\right)\right],\\
        =&\mathbb{E}_{r,s'\sim P_{R,S}^\star(s,a)}\Bigg[\frac{\lVert r_\theta(s,a) - r\rVert^2_2-\lVert r^\star(s,a) - r\rVert^2_2}{2\sigma^2_r}\\
        &\qquad\qquad\qquad\qquad\qquad+\frac{\lVert s_{\theta}'(s,a) - s'\rVert^2_2-\lVert {s^\star}'(s,a) - s'\rVert^2_2}{2\sigma_s^2}\Bigg],\\
        =&\mathbb{E}_{r,s'\sim P_{R,S}^\star(s,a)}\Bigg[\frac{ r_\theta(s,a)^2 - 2rr_\theta(s,a)-r^\star(s,a)^2 +2rr^\star(s,a)}{2\sigma_r^2}\\
        &\qquad\qquad\qquad\qquad\qquad+\frac{\lVert s_{\theta}'(s,a)\rVert^2_2 - 2s'^\top s_{\theta}'(s,a)-\lVert {s^\star}'(s,a)\rVert^2_2 + 2s'^\top {s^\star}'(s,a)}{2\sigma^2_s}\Bigg],\\
        =&\frac{r_\theta(s,a)^2 - 2r^\star(s,a)r_\theta(s,a)-r^\star(s,a)^2 +2r^\star(s,a)^2}{2\sigma_r^2}\\
        &\qquad+\frac{\lVert s_{\theta}'(s,a)\rVert^2_2 - 2{s^\star}'(s,a)^\top s_{\theta}'(s,a)-\lVert {s^\star}'(s,a)\rVert^2_2 + 2\lVert {s^\star}'(s,a)\rVert^2_2}{2\sigma_s^2},\\
        =&\frac{ r_\theta(s,a)^2 - 2r^\star(s,a)r_\theta(s,a)+r^\star(s,a)^2}{2\sigma_r^2}\\
        &\qquad+\frac{\lVert s_{\theta}'(s,a)\rVert^2_2 - 2{s^\star}'(s,a)^\top s_{\theta}'(s,a)+ \lVert {s^\star}'(s,a)\rVert^2_2}{2\sigma_s^2}.
        \end{align}
        Now, taking expectations with respect to the posterior:
        \begin{align}
           \mathbb{E}_{\theta\sim P_\Theta(\mathcal{D}_N)}&\left[ \textrm{KL}\left(P_{R,S}^\star(s,a)\Vert P_{R,S}(s,a,\theta) \right)\right]\\
           &\quad=\mathbb{E}_{\theta\sim P_\Theta(\mathcal{D}_N)}\left[ \frac{r_\theta(s,a)^2 - 2r^\star(s,a)r_\theta(s,a)+r^\star(s,a)^2}{2\sigma^2_r}\right] \\
        &\qquad+\mathbb{E}_{\theta\sim P_\Theta(\mathcal{D}_N)}\left[ \frac{\lVert s_{\theta}'(s,a)\rVert^2_2 - 2{s^\star}'(s,a)^\top s_{\theta}'(s,a)+ \lVert {s^\star}'(s,a)\rVert^2_2}{2\sigma_s^2}\right],\\
        &\quad= \frac{\mathbb{E}_{\theta\sim P_\Theta(\mathcal{D}_N)}\left[r_\theta(s,a)^2\right] - 2r^\star(s,a)r(s,a,\mathcal{D}_N)+r^\star(s,a)^2}{2\sigma^2_r}\\
        &\qquad+ \frac{\mathbb{E}_{\theta\sim P_\Theta(\mathcal{D}_N)}\left[\lVert s_{\theta}'(s,a)\rVert^2_2\right] - 2{s^\star}'(s,a)^\top s'(s,a,\mathcal{D}_N)+ \lVert {s^\star}'(s,a)\rVert^2_2}{2\sigma_s^2}.
        \end{align}
Now, we use the variance identity for both the reward and state functions: $\mathbb{E}_{\theta\sim P_\Theta(\mathcal{D}_N)}\left[ r_{\theta}(s,a)^2\right]=\mathbb{V}_{\theta\sim P_\Theta(\mathcal{D}_N)}\left[r_{\theta}(s,a)\right]+r(s,a,\mathcal{D}_N)^2$ and $\mathbb{E}_{\theta\sim P_\Theta(\mathcal{D}_N)}\left[\lVert s_{\theta}'(s,a)\rVert^2_2\right]=\mathbb{V}_{\theta\sim P_\Theta(\mathcal{D}_N)}\left[\lVert s_{\theta}'(s,a)\rVert_2\right]+\lVert s'(s,a,\mathcal{D}_N)\rVert^2_2$ yielding:
\begin{align}
     &\mathbb{E}_{\theta\sim P_\Theta(\mathcal{D}_N)}\left[ \textrm{KL}\left(P_{R,S}^\star(s,a)\Vert P_{R,S}(s,a,\theta) \right)\right]\\
       &\quad= \frac{\mathbb{V}_{\theta\sim P_\Theta(\mathcal{D}_N)}\left[r_{\theta}(s,a)\right]+r(s,a,\mathcal{D}_N)^2 - 2r^\star(s,a)r(s,a,\mathcal{D}_N)+r^\star(s,a)^2}{2\sigma^2_r}\\
        &\qquad+ \frac{\mathbb{V}_{\theta\sim P_\Theta(\mathcal{D}_N)}\left[\lVert s_{\theta}'(s,a)\rVert_2\right]+\lVert s'(s,a,\mathcal{D}_N)\rVert^2_2 - 2{s^\star}'(s,a)^\top s'(s,a,\mathcal{D}_N)+ \lVert {s^\star}'(s,a)\rVert^2_2}{2\sigma_s^2},\\
        &\quad= \frac{\mathbb{V}_{\theta\sim P_\Theta(\mathcal{D}_N)}\left[r_{\theta}(s,a)\right]+(r(s,a,\mathcal{D}_N)^2 - r^\star(s,a))^2}{2\sigma^2_r}\\
        &\qquad+ \frac{\mathbb{V}_{\theta\sim P_\Theta(\mathcal{D}_N)}\left[\lVert s_{\theta}'(s,a)\rVert_2\right]+\lVert s'(s,a,\mathcal{D}_N) -{s^\star}'(s,a)\rVert^2_2}{2\sigma_s^2},\\
         &\quad= \frac{\mathbb{V}_{\theta\sim P_\Theta(\mathcal{D}_N)}\left[r_{\theta}(s,a)\right]}{2\sigma^2_r}+\frac{\mathbb{V}_{\theta\sim P_\Theta(\mathcal{D}_N)}\left[\lVert s_{\theta}'(s,a)\rVert_2\right]}{2\sigma_s^2}\\
        &\qquad+\frac{(r(s,a,\mathcal{D}_N) - r^\star(s,a))^2}{2\sigma^2_r}+ \frac{\lVert s'(s,a,\mathcal{D}_N) -{s^\star}'(s,a)\rVert^2_2}{2\sigma_s^2},\\
        &\quad= \mathcal{E}(\mathcal{D}_N,\mathcal{M}^\star)+ \mathcal{V}(\mathcal{D}_N),
\end{align}
and hence:
\begin{align}
    \mathcal{I}_N^\pi&=\mathbb{E}_{\theta\sim P_\Theta(\mathcal{D}_N)}\left[ \textrm{KL}\left(P_{R,S}^\star(s,a)\Vert P_{R,S}(s,a,\theta) \right)\right],\\
    &=\mathcal{E}(\mathcal{D}_N,\mathcal{M}^\star)+ \mathcal{V}(\mathcal{D}_N).
\end{align}
\end{proof}
\end{proposition}

\subsection{Proof of \cref{proof:info_rate}}
	
\label{app:frequentist_justification}

We first introduce some simplifying notation for the expected cross entropy, log likelihood and corresponding gradients and Hessian:
\begin{align}
	\ell(\theta)&\coloneqq \mathbb{E}_{s,a\sim \rho_\pi^\star,r,s'\sim P_{R,S}^\star(s,a)}\left[\log p(r,s'\vert s,a,\theta)\right],\\
	\ell^\star &\coloneqq  \max_{\theta\in\Theta}\ell(\theta)=\mathbb{E}_{s,a\sim \rho_\pi^\star,r,s'\sim P_{R,S}^\star(s,a)}\left[\log p^\star(r,s'\vert s,a)\right],\\
	\ell_N(\theta)&\coloneqq \frac{1}{N}\sum_{i=0}^{N-1}\log p(r_i,s'_i\vert s_i,a_i,\theta),\\
    g^\star_{i,N}&\coloneqq\sqrt{N}\nabla_\theta\ell_N(\theta)\big\vert_{\theta=\theta^\star_i},\\
    H^\star_i&\coloneqq\nabla_\theta^2 \ell(\theta)\big\vert_{\theta=\theta^\star_i}.
\end{align}

We now introduce key regularity assumptions for our parametric model that are required to derive the convergence rate for PIL. They are relatively mild and commonplace in the asymptotic statistics literature \citep{LeCam53,barron1988exponential,Clarke1990, Komaki96,Hartigan1998, Barron1999,Aslan06}. 

\begin{assumption}\label{ass:model_regularity}
    We assume that:
  \begin{enumerate}[label=\Roman*.]
    \item There exists at least one parametrisation that corresponds to the true environment dynamics with:
    \begin{align}
      \left\lvert\mathbb{E}_{s,a\sim \rho_\pi^\star,r,s'\sim P_{R,S}^\star(s,a)}\left[\log p^\star(r,s'\vert s,a)\right]\right\rvert<\infty 
    \end{align}
    and $\left\lvert\ell^\star-\ell(\theta) \right\rvert$ is bounded $P_\Theta$-almost surely.
        \item $\ell_N(\theta)$ and $\ell(\theta)$ are $C^2$-continuous in $\theta$.
    \item There are $K<\infty$  maximising points $\theta_i^\star$:
    \begin{align}
    	\{\theta_1^\star,\theta_2^\star,\dots \theta_K^\star\}=\argmax_{\theta\in\Theta}\ell(\theta).
    \end{align}
    For each maximiser $\theta^\star_i$, there exists a small region $\Theta^\star_i\coloneqq \{\theta\in\Theta\vert \lVert \theta^\star_i-\theta\rVert\le\epsilon \}  $ for some $\epsilon>0$ such that $\theta^\star_i$ is the unique maximiser in $\Theta_i^\star$,  $\theta^\star_i$ is in the interior of $\Theta^\star_i$, $\nabla_\theta^2\ell(\theta^\star_i)$ is negative definite, invertible and the regions are disjoint: $\bigcap_{i=1}^K \Theta^\star_i=\varnothing$.

    \item The  prior $p(\theta)$ is Lipschitz continuous in $\theta$ with support over $\Theta$.
    \item The sampling regime ensures that the strong law of large numbers holds for all maximisers $\theta_i^\star$ for the Hessian, and uniformly for $\theta\in\Theta$ for the likelihood, that is:
    \begin{align}
      \ell_N(\theta) \xrightarrow{\textrm{Unif. } a.s.} \ell(\theta),\quad \nabla^2_\theta\ell_N(\theta^\star_i) \xrightarrow{a.s.} \nabla^2_\theta\ell(\theta^\star_i).
    \end{align}
    The central limit theorem applies to the gradient at each $\theta^\star_i$, that is:
    \begin{align}
    	\sqrt{N}\nabla_\theta\ell_N(\theta^\star_i)\xrightarrow{d}\mathcal{N}(0,\Sigma_i^g),
    \end{align}
    where $\Sigma_i^g=\mathbb{E}_{s,a\sim \rho_\pi^\star,r,s'\sim P_{R,S}^\star(s,a)}\left[\nabla_\theta\log p(r,s'\vert s,a,\theta^\star_i)\nabla_\theta\log p(r,s'\vert s,a,\theta^\star_i)^\top\right]$ with $\lVert\Sigma_i^g\rVert<\infty$.
    \end{enumerate}
\end{assumption}
 \cref{ass:model_regularity}i is our strictest assumption and is included for ease of exposition. We generalise our theory in \cref{app:generalisation} to relax this assumption and also account for incomplete Bayes-optimal policy learning. \cref{ass:model_regularity}ii ensures that a second order Taylor series expansion can be applied to obtain an asymptotic expansion around the maximising points.
\cref{ass:model_regularity}iii is much more general than most settings, which only consider problems with a single maximiser. The invertibility of the matrix can easily be guaranteed in Bayesian methods by the use of a prior that can re-condition a low rank matrix that  may results from linearly dependent data.
\cref{ass:model_regularity}iv ensures that the prior places sufficient mass on the true parametrisation.
The sampling and model would need to be very irregular for \cref{ass:model_regularity}v not to hold; stochastic optimisation methods used to find statistics like the MAP will fail if this assumption did not hold. \cref{ass:model_regularity}v holds automatically if sampling is either i.i.d. from $s,a\sim\rho^\star_\pi $ (see e.g. \citet{Bass13}) or from an aperiodic and irreducible Markov chain with stationary distribution $\rho^\star_\pi$ (see e.g.  \citet{Roberts04}). In both sample regimes,  noting that $\mathbb{E}_{s,a\sim \rho_\pi^\star,r,s'\sim P_{R,S}^\star(s,a)}\left[\nabla_\theta\log p(r,s'\vert s,a,\theta^\star_i)\right]=0$, it's clear the (long run) covariance of $\nabla_\theta\log p(r,s'\vert s,a,\theta^\star_i)$ is $\Sigma_i^g$.

Our first lemma borrows techniques from \citet[Chapter 10]{vaart98}. This approach is similar to asymptotic integral expansion approaches that apply Laplace's method \citep{Lindley80, Tierney86,Tierney89,kass90} except we expand around the global maximising values of $\ell(\theta)$ rather than the maximising values of the likelihood $\ell_N(\theta)$ to obtain an asymptotic expression for the posterior: 

\begin{lemma}\label{proof:contraction_maxima}
    Under \cref{ass:model_regularity} and using the notation introduced at the start of \cref{app:frequentist_justification}:
    \begin{align}
      \frac{\int_{\Theta^\star_i}\left( \ell^\star-\ell(\theta) \exp\left(N\ell_N(\theta)\right)\right)p(\theta)d\theta}{\int_{\Theta^\star_i}  \exp\left(N\ell_N(\theta)\right)p(\theta)d\theta}=\mathcal{O}\left(\frac{d-{g^\star_{i,N}}^\top{H^\star_i}^{-1}g^\star_{i,N}
        }{N}\right),
    \end{align}
    almost surely.
    \begin{proof}
        We start by applying the transformation of variables $\theta'=f(\theta)\coloneqq\sqrt{N}(\theta-\theta^\star_i)$ to integrals in the numerator and denominator with:
        \begin{align}
            \theta=f^{-1}(\theta')=\theta^\star_i+\frac{1}{\sqrt{N}}\theta',\quad \left\lvert \det\nabla_\theta f^{-1}(\theta')\right\rvert=N^{-\frac{d}{2}},\quad \Theta'\coloneqq f(\Theta^\star_i),
        \end{align}
        yielding:
        \begin{align}
            &\frac{\int_{\Theta^\star_i}(\ell^\star- \ell(\theta) )\exp\left(N\ell_N(\theta)\right)p(\theta)d\theta}{\int_{\Theta^\star}  \exp\left(N\ell_N(\theta)\right)p(\theta)d\theta}\\
            &\qquad=\frac{\int_{\Theta'}\left(\ell^\star- \ell\left(\theta=\theta^\star+\frac{1}{\sqrt{N}}\theta'\right) \right)\exp\left(N\ell_N\left(\theta=\theta^\star_i+\frac{1}{\sqrt{N}}\theta'\right)\right)p'(\theta')d\theta'}{\int_{\Theta'}  \exp\left(N\ell_N\left(\theta=\theta^\star_i+\frac{1}{\sqrt{N}}\theta'\right)\right)p'(\theta')d\theta'},\label{eq:cov}
        \end{align} 
        where $p'(\theta')\coloneqq p\left(\theta=\theta^\star_i+\frac{1}{\sqrt{N}}\theta'\right)$. Now and making a Taylor series expansion of $\ell(\theta)$ about $\theta^\star_i$:
        \begin{align}
            \ell(\theta)&=\ell^\star+{\underbrace{\nabla_\theta \ell(\theta^\star_i)}_{=0}}^\top(\theta-\theta^\star_i)+(\theta-\theta^\star_i)^\top H^\star_i(\theta-\theta^\star_i)+\mathcal{O}\left(\lVert \theta-\theta^\star_i\rVert^3\right),\\
            &=\ell^\star+(\theta-\theta^\star_i)^\top H^\star_i(\theta-\theta^\star_i)+\mathcal{O}\left(\lVert \theta-\theta^\star_i\rVert^3\right),
        \end{align}
        hence:
        \begin{align}
            \ell\left(\theta=\theta^\star_i+\frac{1}{\sqrt{N}}\theta'\right)
            &=\ell^\star+\frac{1}{N}\theta'^\top H^\star_i \theta'+\mathcal{O}\left(N^{-\frac{3}{2}}\right).
        \end{align}
        Using the notation $H^\star_N \coloneqq\nabla_\theta^2 \ell_N(\theta)\big\vert_{\theta=\theta^\star_i}$ and making a Taylor series expansion of $\ell_N(\theta)$ about $\theta^\star_i$:
        \begin{align}
            \ell_N(\theta)=\ell_N(\theta^\star_i)+\nabla_\theta \ell_N(\theta^\star_i)^
        \top(\theta-\theta^\star_i)+(\theta-\theta^\star_i)^\top\nabla_\theta^2\ell_N(\theta^\star_i)(\theta-\theta^\star_i)+ \mathcal{O}\left(\lVert \theta-\theta^\star_i\rVert^3\right),
        \end{align}
        hence:
        \begin{align}
            N\ell_N\left(\theta=\theta^\star_i+\frac{1}{\sqrt{N}}\theta'\right)=N\ell_N(\theta^\star_i)+\sqrt{N}\nabla_\theta \ell_N(\theta^\star_i)^\top\theta'+\theta'^\top\nabla_\theta^2\ell_N(\theta^\star_i)\theta'+\mathcal{O}\left(\frac{1}{\sqrt{N}}\right).
        \end{align}
          Substituting into \cref{eq:cov} yields:
        \begin{align}
            &\frac{\int_{\Theta^\star} \left(\ell^\star-\ell(\theta) \right)\exp\left(N\ell(\theta)\right)p(\theta)d\theta}{\int_{\Theta^\star}  \exp\left(N\ell(\theta)\right)p(\theta)d\theta}\\
            &=-\frac{\int_{\Theta'} \theta'^\top H^\star_i\theta'\exp\left(N\ell_N(\theta^\star_i)+{g^\star_{i,N}
        }^\top\theta'+\theta'^\top H^\star_N\theta'+\mathcal{O}\left(\frac{1}{\sqrt{N}}\right)\right)p'(\theta')d\theta'}{\int_{\Theta'}  \exp\left(N\ell_N(\theta^\star_i)+{g^\star_{i,N}
        }^\top\theta'+\theta'^\top H^\star_N\theta'+\mathcal{O}\left(\frac{1}{\sqrt{N}}\right)\right)p'(\theta')d\theta'}\mathcal{O}\left(\frac{1}{N}\right),\\
            &=-\frac{\int_{\Theta'} \theta'^\top H^\star_i\theta'\exp\left({g^\star_{i,N}
        }^\top\theta'+\theta'^\top H^\star_N\theta'+\mathcal{O}\left(\frac{1}{\sqrt{N}}\right)\right)p'(\theta')d\theta'}{\int_{\Theta'}  \exp\left(N{g^\star_{i,N}
        }^\top\theta'+\theta'^\top H^\star_N\theta'+\mathcal{O}\left(\frac{1}{\sqrt{N}}\right)\right)p'(\theta')d\theta'}\mathcal{O}\left(\frac{1}{N}\right),\\
           &=-\frac{\int_{\Theta'} \theta'^\top H^\star_i\theta'\exp\left({g^\star_{i,N}
        }^\top\theta'+\theta'^\top H^\star_N\theta'\right)\exp\left(\mathcal{O}\left(\frac{1}{\sqrt{N}}\right)\right)p'(\theta')d\theta'}{\int_{\Theta'}  \exp\left(N{g^\star_{i,N}
        }^\top\theta'+\theta'^\top H^\star_N\theta'\right)\exp\left(\mathcal{O}\left(\frac{1}{\sqrt{N}}\right)\right)p'(\theta')d\theta'}\mathcal{O}\left(\frac{1}{N}\right),\\
   &=\mathcal{O}\left(-\frac{1}{N}\frac{\int_{\Theta'} \theta'^\top H^\star_i\theta'\exp\left({g^\star_{i,N}
   	}^\top\theta'+\theta'^\top H^\star_N\theta'\right)p'(\theta')d\theta'}{\int_{\Theta'}  \exp\left(N{g^\star_{i,N}
   	}^\top\theta'+\theta'^\top H^\star_N\theta'\right)p'(\theta')d\theta'}\right)\label{eq:Gaussian_form}
        \end{align}
       where we have multiplied top and bottom by $\exp\left(-N\ell_N(\theta^\star_i)\right)$ to derive the second equality and used the fact that  $0<\exp\left(\mathcal{O}\left(\frac{1}{\sqrt{N}}\right)\right)<\infty$ to derive the final line. Now, as the prior is Lipschitz, we make a Taylor series expansion about $\theta^\star_i$:
       \begin{align}
           p(\theta)=p(\theta^\star_i)+\mathcal{O}\left(\lVert\theta-\theta^\star_i\rVert\right),
       \end{align}
       hence:
       \begin{align}
           p'(\theta')=p\left(\theta=\theta^\star_i+\frac{1}{\sqrt{N}} \theta'\right)=p(\theta^\star_i)+\mathcal{O}\left(\frac{1}{\sqrt{N}}\right).
       \end{align}
       This allows us to find an asymptotic expression for \cref{eq:Gaussian_form}:
       \begin{align}
           &\frac{\int_{\Theta'} \theta'^\top H^\star_i\theta' \exp\left({g^\star_{i,N}
        }^\top\theta'+\theta'^\top H^\star_N\theta'\right)p'(\theta')d\theta'}{\int_{\Theta'}  \exp\left({g^\star_{i,N}
        }^\top\theta'+\theta'^\top H^\star_N\theta'\right)p'(\theta')d\theta'}\\
           &\qquad=\frac{\int_{\Theta'} \theta'^\top H^\star_i\theta' \exp\left({g^\star_{i,N}
        }^\top\theta'+\theta'^\top H^\star_N\theta'\right)p'(\theta^\star_i)d\theta'\left(1+\mathcal{O}\left(\frac{1}{\sqrt{N}}\right)\right)}{\int_{\Theta'}  \exp\left({g^\star_{i,N}
        }^\top\theta'+\theta'^\top H^\star_N\theta'\right)p'(\theta^\star_i)d\theta'\left(1+\mathcal{O}\left(\frac{1}{\sqrt{N}}\right)\right)}\\
           &\qquad=\frac{\int_{\Theta'} \theta'^\top H^\star_i\theta' \exp\left({g^\star_{i,N}
        }^\top\theta'+\theta'^\top H^\star_N\theta'\right)p'(\theta^\star_i)d\theta'}{\int_{\Theta'}  \exp\left({g^\star_{i,N}
        }^\top\theta'+\theta'^\top H^\star_N\theta'\right)p'(\theta^\star_i)d\theta'}\left(1+\mathcal{O}\left(\frac{1}{\sqrt{N}}\right)\right)\\
            &\qquad=\frac{\int_{\Theta'} \theta'^\top H^\star_i\theta' \exp\left({g^\star_{i,N}
        }^\top\theta'+\theta'^\top H^\star_N\theta'\right)d\theta'}{\int_{\Theta'}  \exp\left({g^\star_{i,N}
        }^\top\theta'+\theta'^\top H^\star_N\theta'\right)d\theta'}\left(1+\mathcal{O}\left(\frac{1}{\sqrt{N}}\right)\right).
       \end{align}
       We re-write the exponential term to recover a  quadratic form:
       \begin{align}
       	 &\exp\left({g^\star_{i,N}
       	}^\top\theta'+\theta'^\top H^\star_N\theta'\right)\\
        &=\exp\left(\left(\frac{1}{2}{H^\star_N}^{-1}g^\star_{i,N}+  \theta'\right)^\top H^\star_N\left(\frac{1}{2}{H^\star_N}^{-1}g^\star_{i,N}+  \theta'\right)-\frac{1}{4}{g^\star_{i,N}}^\top{H^\star_N}^{-1}g^\star_{i,N}\right).
       \end{align}
       Substituting yields:
       \begin{align}
           &\frac{\int_{\Theta'} \theta'^\top H^\star_i\theta' \exp\left({g^\star_{i,N}
        }^\top\theta'+\theta'^\top H^\star_N\theta'\right)d\theta'}{\int_{\Theta'}  \exp\left({g^\star_{i,N}
        }^\top\theta'+\theta'^\top H^\star_N\theta'\right)d\theta'}\\
        &=\frac{\int_{\Theta'} \theta'^\top H^\star_i\theta' \exp\left(\left(\frac{1}{2}{H^\star_N}^{-1}g^\star_{i,N}+  \theta'\right)^\top H^\star_N\left(\frac{1}{2}{H^\star_N}^{-1}g^\star_{i,N}+  \theta'\right)\right)d\theta'}{\int_{\Theta'}  \exp\left(\left(\frac{1}{2}{H^\star_N}^{-1}g^\star_{i,N}+  \theta'\right)^\top H^\star_N\left(\frac{1}{2}{H^\star_N}^{-1}g^\star_{i,N}+  \theta'\right)\right)d\theta'}.
       \end{align}
        In this form, we notice the expectation is that of a Gaussian $\mathcal{N}\left(\mu=-\frac{1}{2}{H^\star_N}^{-1}g^\star_{i,N},\Sigma=-{H^\star_i}^{-1}\right)$ restricted to $\Theta'$. Noting that in the limit $\Theta'\xrightarrow[N\rightarrow\infty]{} \mathbb{R}^d$, hence:
       \begin{align}
       &\frac{\int_{\Theta'} \theta'^\top H^\star_i\theta'\exp\left( \left(\theta'+\frac{1}{2}{H^\star_N}^{-1}g^\star_{i,N}
       	\right)^\top H^\star_N\left(\theta'+\frac{1}{2}{H^\star_N}^{-1}g^\star_{i,N}
       	\right)\right)d\theta'}{\int_{\Theta'}  \exp\left( \left(\theta'+\frac{1}{2}{H^\star_N}^{-1}g^\star_{i,N}
       	\right)^\top H^\star_N\left(\theta'+\frac{1}{2}{H^\star_N}^{-1}g^\star_{i,N}
       	\right)\right)d\theta'}\\
       	&=\mathcal{O}\left(\frac{\int_{\mathbb{R}^d} \theta'^\top H^\star_i\theta'\exp\left( \left(\theta'+\frac{1}{2}{H^\star_N}^{-1}g^\star_{i,N}
       	\right)^\top H^\star_N\left(\theta'+\frac{1}{2}{H^\star_N}^{-1}g^\star_{i,N}
       	\right)\right)d\theta'}{\int_{\mathbb{R}^d}  \exp\left( \left(\theta'+\frac{1}{2}{H^\star_N}^{-1}g^\star_{i,N}
       	\right)^\top H^\star_N\left(\theta'+\frac{1}{2}{H^\star_N}^{-1}g^\star_{i,N}
       	\right)\right)d\theta'}\right),\\
       	&=\mathcal{O}\left(\mathbb{E}_{\theta'\sim\mathcal{N}\left(-\frac{1}{2}{H^\star_N}^{-1}g^\star_{i,N},-{H^\star_N}^{-1}\right)}\left[\theta'^\top H^\star_i\theta'\right]\right).\label{eq:mvg_variance}
       \end{align}
        Putting everything together, we have:
       \begin{align}
       	 &\frac{\int_{\Theta^\star_i}\left( \ell^\star-\ell(\theta) \exp\left(N\ell_N(\theta)\right)\right)p(\theta)d\theta}{\int_{\Theta^\star_i}  \exp\left(N\ell_N(\theta)\right)p(\theta)d\theta}\\
       	&\qquad=\mathcal{O}\left(-\frac{1}{N}\frac{\int_{\Theta'} \theta'^\top H^\star_i\theta'\exp\left({g^\star_{i,N}
       		}^\top\theta'+\theta'^\top H^\star_N\theta'\right)p'(\theta')d\theta'}{\int_{\Theta'}  \exp\left({g^\star_{i,N}
       		}^\top\theta'+\theta'^\top H^\star_N\theta'\right)p'(\theta')d\theta'}\right),\quad\textrm{\cref{eq:Gaussian_form}}\\
       	&\qquad=\mathcal{O}\left(-\frac{1}{N}\mathbb{E}_{\theta'\sim\mathcal{N}\left(-\frac{1}{2}{H^\star_N}^{-1}g^\star_{i,N},-{H^\star_i}^{-1}\right)}\left[\theta'^\top H^\star_i\theta'\right]\right),\quad\textrm{\cref{eq:mvg_variance}}
       \end{align}
       
       Using standard results for the multivariate Gaussian  \citep{Petersen12} yields our desired result: 
       \begin{align}
       	-\frac{1}{N}\mathbb{E}_{\theta'\sim\mathcal{N}\left(-\frac{1}{2}{H^\star_N}^{-1}g^\star_{i,N},-{H^\star_i}^{-1}\right)}\left[\theta'^\top H^\star_i\theta'\right]&=\frac{\textrm{Tr}\left(H^\star_i{H^\star_N}^{-1}\right)-\frac{1}{4}{g^\star_{i,N}}^\top {{H^\star_N}^{-1}}^\top H^\star_i{H^\star_N}^{-1} g^\star_{i,N}}{N},\\
 &=\mathcal{O}\left(\frac{\textrm{Tr}\left(I\right)-{g^\star_{i,N}}^\top  {H^\star_i}^{-1} g^\star_{i,N}}{N}\right),\\
       	&=\mathcal{O}\left(\frac{d-{g^\star_{i,N}}^\top{H^\star_i}^{-1}g^\star_{i,N}
        }{N}\right),
       \end{align}
      almost surely, where we have used the strong law of large numbers on the empirical Hessian from \cref{ass:model_regularity} to derive the second line. 
    \end{proof}
\end{lemma}
In our final Lemma, we show that regions that are not close to the maximising points diminish exponentially in posterior probability as $N$ grows large.
\begin{lemma}\label{proof:posterior_tail_contraction_rate}
	Under \cref{ass:model_regularity}, $\mathbb{E}_{\mathcal{D}_N\sim P_\textrm{Data}^\star}\left[ P(\bar{\Theta}\vert \mathcal{D}_N)\right]=\mathcal{O}\left(\exp(-N)\right)$.
	\begin{proof}
We start by splitting the posterior expectation into integrals over $\bar{\Theta}$ and $\Theta\setminus \bar{\Theta}$:
\begin{align}
    P(\bar{\Theta}\vert \mathcal{D}_N)&=\frac{\int_{\bar{\Theta}}\exp\left(N\ell_N(\theta)\right)p(\theta)d\theta}{\int_{\Theta}\exp\left(N\ell_N(\theta)\right)p(\theta)d\theta},\\
    &=\frac{\int_{\bar{\Theta}}\exp\left(N\ell_N(\theta)\right)p(\theta)d\theta}{\int_{\bar{\Theta}}\exp\left(N\ell_N(\theta)\right)p(\theta)d\theta+\int_{\Theta\setminus\bar{\Theta}}\exp\left(N\ell_N(\theta)\right)p(\theta)d\theta}.
\end{align}
Dividing top and bottom by $\int_{\bar{\Theta}}\exp\left(N\ell_N(\theta)\right)p(\theta)d\theta$:
  \begin{align}
            P(\bar{\Theta}\vert \mathcal{D}_N)= \frac{1}{1+\frac{\int_{\Theta\setminus \bar{\Theta}} \exp\left(N\ell_N(\theta)\right)p(\theta)d\theta}{\int_{\bar{\Theta}} \exp\left(N\ell_N(\theta)\right)p(\theta)d\theta}}.
        \end{align}
        Hence if we can show there exists some $N'<\infty$ and a function $C \exp(c N)$ with positive constants $c$ and $C$ that lower bounds the ratio:
        \begin{align}
            C \exp(c N)\le \frac{\int_{\Theta\setminus \bar{\Theta}} \exp\left(N\ell_N(\theta)\right)p(\theta)d\theta}{\int_{\bar{\Theta}} \exp\left(N\ell_N(\theta)\right)p(\theta)d\theta}
        \end{align}
        almost surely for all $N\ge N'$, then it follows:
        \begin{align}
            \mathbb{E}_{\mathcal{D}_N\sim P_\textrm{Data}^\star}\left[ P(\bar{\Theta}\vert \mathcal{D}_N)\right]&=\mathcal{O}\left(\frac{1}{1+\exp(N)}\right),\\
            &=\mathcal{O}\left(\exp(-N)\right).
        \end{align}
        
	From \cref{ass:model_regularity}, each $\theta^\star_i$ maximises $\ell(\theta)$ with $\sup_{\theta\in\bar{\Theta}} \ell(\theta')<\ell(\theta^\star_i) $. As $\ell(\theta)$ is continuous, there thus exists a small, closed ball $B(\theta^\star_j,r)\coloneqq \{\theta\vert \| \theta^\star_i-\theta\| \le r\}$ of radius $r>0$ centred on some $\theta^\star_j$ such that $\sup_{\theta'\in\bar{\Theta}}\ell(\theta')<\min_{\theta''\in B(\theta^\star_j,r)}\ell(\theta'')$. From \cref{ass:model_regularity}, the uniform strong law of large numbers holds with $\ell_N(\theta)\xrightarrow[]{\textrm{Unif. } a.s}\ell(\theta)$. By the definition of the limit and continuity of $\ell_N(\theta)$, there thus exists some finite $N'$ such that $\sup_{\theta'\in\bar{\Theta}}\ell_N(\theta')<\min_{\theta''\in B(\theta^\star_j,\frac{r}{2})}\ell_N(\theta'')$ for all $N\ge N'$ almost surely, where $B(\theta^\star_j,\frac{r}{2})$ is a ball of half radius $\frac{r}{2}$. Noting that $B(\theta^\star_j,\frac{r}{2})\subset \Theta\setminus\bar{\Theta}$ and $0\le\exp\left(N\ell_N(\theta)\right)$, this allows us to lower bound the integral:
		\begin{align}
			\int_{\Theta\setminus \bar{\Theta}} \exp\left(N\ell_N(\theta)\right)p(\theta)d\theta&\ge \int_{B(\theta^\star_j,\frac{r}{2})} \exp\left(N\ell_N(\theta)\right)p(\theta)d\theta,\\
            &\ge\exp\left(N\min_{\theta'' \in B(\theta^\star_j,\frac{r}{2})}\ell_N(\theta'')\right)\int_{B(\theta^\star_j,\frac{r}{2})} p(\theta)d\theta,\\
            &=\exp\left(N\min_{\theta'' \in B(\theta^\star_j,\frac{r}{2})}\ell_N(\theta'')\right)P\left(B\left(\theta^\star_j,\frac{r}{2}\right)\right).
		\end{align}
         We can also upper bound the integral:
        \begin{align}
            \int_{\bar{\Theta}} \exp\left(N\ell_N(\theta)\right)p(\theta)d\theta&\le\exp\left(N\sup_{\theta'\in\bar{\Theta}}\ell_N(\theta')\right)\int_{\bar{\Theta}}p(\theta)d\theta,\\
            &=\exp\left(N\sup_{\theta'\in\bar{\Theta}}\ell_N(\theta')\right)P\left(\bar{\Theta}\right).
        \end{align}
        Using these results, we lower bound the ratio as:
        \begin{align}
            \frac{\int_{\Theta\setminus \bar{\Theta}} \exp\left(N\ell_N(\theta)\right)p(\theta)d\theta}{\int_{\bar{\Theta}} \exp\left(N\ell_N(\theta)\right)p(\theta)d\theta}\ge &\frac{\exp\left(N\min_{\theta'' \in B(\theta^\star_j,\frac{r}{2})}\ell_N(\theta'')\right)P\left(B\left(\theta^\star_j,\frac{r}{2}\right)\right)}{\exp\left(N\sup_{\theta'\in\bar{\Theta}}\ell_N(\theta')\right)P\left(\bar{\Theta}\right)},\\
            =&\exp\left(N\left(\min_{\theta'' \in B(\theta^\star_j,\frac{r}{2})}\ell_N(\theta'')-\sup_{\theta'\in\bar{\Theta}}\ell_N(\theta')\right)\right)\frac{P\left(B\left(\theta^\star_j,\frac{r}{2}\right)\right)}{P\left(\bar{\Theta}\right)}.
        \end{align}
       Let $\frac{P\left(B\left(\theta^\star_j,\frac{r}{2}\right)\right)}{P\left(\bar{\Theta}\right)}=C>0$ from \cref{ass:model_regularity}. As there exists some $N'$ such that $\min_{\theta'' \in B(\theta^\star_j,\frac{r}{2})}\ell_N(\theta'')>\sup_{\theta'\in\bar{\Theta}}\ell_N(\theta')$ for all $N>N'$, we have shown exists some positive constants $c>0$ and $C>0$ such that
        \begin{align}
            C\exp\left(cN\right)\le\frac{\int_{\Theta\setminus \bar{\Theta}} \exp\left(N\ell_N(\theta)\right)p(\theta)d\theta}{\int_{\bar{\Theta}} \exp\left(N\ell_N(\theta)\right)p(\theta)d\theta},
        \end{align}
        for all $N>N'$ almost surely, as required. 
	\end{proof}
\end{lemma}
We now present our proof of \cref{proof:info_rate}. Here we split the posterior expectation up into small regions close to maximising points and regions away from maximising. We then apply our two lemmas to each region. Our result then follows by an application the central limit theorem under \cref{ass:model_regularity}.
\begin{theorem} \label{app:proof_info_rate} Let the data be drawn from the underlying true distribution $\mathcal{D}_N\sim P_\textrm{Data}^\star$.
    Under \cref{ass:model_regularity}, there exists some constant $0<C<\infty$ such that for sufficiently large $N$: 
    \begin{align}
    \mathbb{E}_{\mathcal{D}_N\sim P_\textrm{Data}^\star}\left[\textnormal{\textrm{Regret}}(\mathcal{M}^\star,\mathcal{D}_N)\right] \le 2\mathcal{R}_{\max}\cdot\sqrt{1-\exp\left(-\frac{Cd}{(1-\gamma)N}\right)}.     \label{eq_app:parametric_regret}
    \end{align}

    \begin{proof}
  Using the notation introduced at the start of \cref{app:frequentist_justification}, we write the PIL as:
    \begin{align}
 \mathcal{I}_N^\pi\coloneqq&\mathbb{E}_{s,a\sim \rho_\pi^\star}\left[ \mathbb{E}_{\theta\sim P_\Theta(\mathcal{D}_N)}\left[\textrm{KL}\left(P_{R,S}^\star(s,a)\Vert P_{R,S}(s,a,\theta) \right)\right]\right],\\
 =&\mathbb{E}_{\theta\sim P_\Theta(\mathcal{D}_N)}\left[\mathbb{E}_{s,a\sim \rho_\pi^\star,r,s'\sim P_{R,S}^\star(s,a)}\left[\log p(r,s'\vert s,a,\theta^\star)-\log p(r,s'\vert s,a,\theta)\right]\right],\\
  =&\mathbb{E}_{\theta\sim P_\Theta(\mathcal{D}_N)}\left[\ell^\star-\ell(\theta)\right],
    \end{align}
Under this same notation, we write the posterior density as:
    \begin{align}
        p(\theta\vert \mathcal{D}_N)&=\frac{\exp\left(N\ell_N(\theta)\right)p(\theta)}{\int_\Theta\exp\left(N\ell_N(\theta)\right)p(\theta)d\theta}.\label{eq:posterior_notation}
    \end{align}
   Now, under \cref{ass:model_regularity}, we split the inner expectation into small regions $\Theta_i^\star$ around each maximising point $\theta_i^\star$ and the remainder of the parameter space $\bar{\Theta}\coloneqq \Theta\setminus \bigcup_{i=1}^K \Theta_i^\star$ :
   \begin{align}
   	\mathbb{E}_{\theta\sim P_\Theta(\mathcal{D}_N)}\left[\ell^\star-\ell(\theta)\right]=\sum_{i=1}^K \int_{\Theta_i^\star}  \left(\ell^\star -\ell(\theta)\right) p(\theta\vert \mathcal{D}_N) d\theta+\int_{\bar{\Theta}}   \left(\ell^\star -\ell(\theta)\right) p(\theta\vert \mathcal{D}_N) d\theta.\label{eq:split_integral}
   \end{align}
   Using \cref{eq:posterior_notation}, we now re-write each integral in the summation term of \cref{eq:split_integral} as:
   \begin{align}
    \int_{\Theta_i^\star}  \left(\ell^\star -\ell(\theta)\right) p(\theta\vert \mathcal{D}_N) d\theta&=\frac{\int_{\Theta_i^\star}\left(\ell^\star -\ell(\theta)\right) \exp\left(N\ell_N(\theta)\right)p(\theta)d\theta }{\int_\Theta\exp\left(N\ell_N(\theta)\right)p(\theta)d\theta},\\
    &=\frac{\int_{\Theta_i^\star}\left(\ell^\star -\ell(\theta)\right) \exp\left(N\ell_N(\theta)\right)p(\theta)d\theta }{\int_\Theta\exp\left(N\ell_N(\theta)\right)p(\theta)d\theta}\cdot\frac{\int_{\Theta^\star_i}\exp\left(N\ell_N(\theta)\right)p(\theta)d\theta}{\int_{\Theta^\star_i}\exp\left(N\ell_N(\theta)\right)p(\theta)d\theta},\\
    &=\frac{\int_{\Theta_i^\star}\left(\ell^\star -\ell(\theta)\right) \exp\left(N\ell_N(\theta)\right)p(\theta)d\theta }{\int_{\Theta^\star_i}\exp\left(N\ell_N(\theta)\right)p(\theta)d\theta}\cdot\frac{\int_{\Theta^\star_i}\exp\left(N\ell_N(\theta)\right)p(\theta)d\theta}{\int_\Theta\exp\left(N\ell_N(\theta)\right)p(\theta)d\theta},\\
     &=\frac{\int_{\Theta_i^\star}\left(\ell^\star -\ell(\theta)\right) \exp\left(N\ell_N(\theta)\right)p(\theta)d\theta }{\int_{\Theta^\star_i}\exp\left(N\ell_N(\theta)\right)p(\theta)d\theta}\cdot P(\Theta^\star_i\vert \mathcal{D}_N),\\
     &\le \frac{\int_{\Theta_i^\star}\left(\ell^\star -\ell(\theta)\right) \exp\left(N\ell_N(\theta)\right)p(\theta)d\theta }{\int_{\Theta^\star_i}\exp\left(N\ell_N(\theta)\right)p(\theta)d\theta}.\label{eq:maximiser_bound}
   \end{align}
   where we have used $ 0\le P(\Theta^\star_i\vert \mathcal{D}_N)\le1$ from Kolmogorov's axioms to bound the final line.

For the last term in \cref{eq:split_integral}, we note that $\ell^\star-\ell(\theta)$ is bounded $P_\Theta$-almost surely from \cref{ass:model_regularity}, hence there exists some $\ell^\dagger<\infty$ such that:
\begin{align}
	\int_{\bar{\Theta}}   \left(\ell^\star -\ell(\theta)\right) p(\theta\vert \mathcal{D}_N) d\theta\le &	\int_{\bar{\Theta}}  \ell^\dagger p(\theta\vert \mathcal{D}_N) d\theta,\\
	=&\ell^\dagger P(\bar{\Theta}\vert \mathcal{D}_N). \label{eq:remainder_bound}
\end{align}
Using Ineqs.~\ref{eq:maximiser_bound} and~\ref{eq:remainder_bound}, we bound \cref{eq:split_integral} as:
\begin{align}
	\mathbb{E}_{\theta\sim P_\Theta(\mathcal{D}_N)}\left[\ell^\star-\ell(\theta)\right]\le \sum_{i=1}^K \frac{\int_{\Theta_i^\star}\left(\ell^\star -\ell(\theta)\right) \exp\left(N\ell_N(\theta)\right)p(\theta)d\theta }{\int_{\Theta^\star_i}\exp\left(N\ell_N(\theta)\right)p(\theta)d\theta}+\ell^\dagger P(\bar{\Theta}\vert \mathcal{D}_N),
\end{align}
and hence the PIL can be bounded as:
\begin{align}
	 \mathcal{I}_N^\pi\le  \sum_{i=1}^K \frac{\int_{\Theta_i^\star}\left(\ell^\star -\ell(\theta)\right) \exp\left(N\ell_N(\theta)\right)p(\theta)d\theta }{\int_{\Theta^\star_i}\exp\left(N\ell_N(\theta)\right)p(\theta)d\theta}+\ell^\dagger  P(\bar{\Theta}\vert \mathcal{D}_N).
\end{align}
Applying \cref{proof:contraction_maxima} and \cref{proof:posterior_tail_contraction_rate} under \cref{ass:model_regularity} yields:
\begin{align}
	\mathcal{I}_N^\pi&=  \sum_{i=1}^K \mathcal{O}\left(\frac{{d-g^\star_{i,N}}^\top{H^\star_i}^{-1}g^\star_{i,N}}{N}\right)+\ell^\dagger \mathcal{O}\left(\exp(-N)\right),\\
	&= \mathcal{O}\left(\frac{d- \sum_{i=1}^K{g^\star_{i,N}}^\top{H^\star_i}^{-1}g^\star_{i,N}}{N}\right).\label{eq:pil_asymptotic_bound}
\end{align}
almost surely. As $f(x  )\coloneqq2\mathcal{R}_{\max}\cdot\sqrt{1-\exp\left(-\frac{x}{(1-\gamma)}\right)}$ is monotonic in $x$ and $\frac{d- \sum_{i=1}^K{g^\star_{i,N}}^\top{H^\star_i}^{-1}g^\star_{i,N}}{N}\ge0$, \cref{eq:pil_asymptotic_bound} implies there exists some positive $0<C<\infty$ such that:
\begin{align}
    \textnormal{\textrm{Regret}}(\mathcal{M}^\star,\mathcal{D}_N)\le 2\mathcal{R}_{\max}\cdot\sqrt{1-\exp\left(-C\frac{ d- \sum_{i=1}^K{g^\star_{i,N}}^\top{H^\star_i}^{-1}g^\star_{i,N}}{(1-\gamma)N}\right)},
\end{align}
almost surely for large enough $N$. Under \cref{ass:model_regularity}, $g^\star_{i,N}\xrightarrow{d}\mathcal{N}(0,\Sigma_i^g)$. As $f(x)$ is also a bounded, continuous function and concave, we can apply the Portmanteau Theorem (see for example \citet[Chapter 21.7]{Bass13}) followed by Jensen's inequality to yield:
       \begin{align}
       	\mathbb{E}_{\mathcal{D}_N\sim P_\textrm{Data}^\star}\left[\textnormal{\textrm{Regret}}(\mathcal{M}^\star,\mathcal{D}_N)\right]\le&2\mathcal{R}_{\max}\cdot\mathbb{E}_{g_i\sim \mathcal{N}(0,\Sigma_i^g)}\left[\sqrt{1-\exp\left(-C\frac{d- \sum_{i=1}^K{g_i}^\top{H^\star_i}^{-1}g_i}{(1-\gamma)N}\right)}\right],\\
        \le&2\mathcal{R}_{\max}\cdot\sqrt{1-\exp\left(-C\frac{ d-\sum_{i=1}^K\mathbb{E}_{g_i\sim \mathcal{N}(0,\Sigma_i^g)}\left[{g}^\top{H^\star_i}^{-1}g\right]}{(1-\gamma)N}\right)},\\
        =&2\mathcal{R}_{\max}\cdot\sqrt{1-\exp\left(-C\frac{ d-\sum_{i=1}^K\textrm{Tr}\left(\Sigma_i^g{H^\star_i}^{-1}\right)}{(1-\gamma)N}\right)}.\label{eq:raw_hessian_form}
       \end{align}
        Now, examining the Hessian:
        \begin{align}
            H(\theta)&=\nabla_\theta^2 \mathbb{E}_{s,a\sim \rho_\pi^\star,r,s'\sim P_{R,S}^\star(s,a)}\left[\log p(r,s'\vert s,a,\theta)\right] \\
            &=\nabla_\theta \mathbb{E}_{s,a\sim \rho_\pi^\star,r,s'\sim P_{R,S}^\star(s,a)}\left[\nabla_\theta \log p(r,s'\vert s,a,\theta)\right],\\
            &=\nabla_\theta \mathbb{E}_{s,a\sim \rho_\pi^\star,r,s'\sim P_{R,S}^\star(s,a)}\left[\frac{\nabla_\theta p(r,s'\vert s,a,\theta)}{p(r,s'\vert s,a,\theta)}\right],\\
            &=\mathbb{E}_{s,a\sim \rho_\pi^\star,r,s'\sim P_{R,S}^\star(s,a)}\left[\nabla_\theta \frac{ \nabla_\theta p(r,s'\vert s,a,\theta)}{p(r,s'\vert s,a,\theta)}\right],\\
           &=\mathbb{E}_{s,a\sim \rho_\pi^\star,r,s'\sim P_{R,S}^\star(s,a)}\left[ \frac{ \nabla_\theta^2 p(r,s'\vert s,a,\theta)}{p(r,s'\vert s,a,\theta)}\right]\\
           &\qquad-\mathbb{E}_{s,a\sim \rho_\pi^\star,r,s'\sim P_{R,S}^\star(s,a)}\left[\frac{ \nabla_\theta p(r,s'\vert s,a,\theta)}{p(r,s'\vert s,a,\theta)}\frac{ \nabla_\theta p(r,s'\vert s,a,\theta)^\top}{p(r,s'\vert s,a,\theta)}\right],\label{eq:expanded_hessian}
\end{align}
Hence at $\theta=\theta^\star_i$, the first term of \cref{eq:expanded_hessian} is:
\begin{align}
           \mathbb{E}_{s,a\sim \rho_\pi^\star,r,s'\sim P_{R,S}^\star(s,a)}\left[ \frac{ \nabla_\theta^2 p(r,s'\vert s,a,\theta)}{p(r,s'\vert s,a,\theta^\star_i)}\right]&=\mathbb{E}_{s,a\sim \rho_\pi^\star,r,s'\sim P_{R,S}^\star(s,a)}\left[  \frac{\nabla_\theta^2 p(r,s'\vert s,a,\theta)\vert_{\theta=\theta^\star_i}}{p^\star(r,s'\vert s,a)}\right],\\
           &=\mathbb{E}_{s,a\sim \rho_\pi^\star}\left[\int_{\mathbb{R}\times \mathcal{S}} \nabla_\theta^2 p(r,s'\vert s,a,\theta)\vert_{\theta=\theta^\star_i}d(r,s')\right],\\
           &=\mathbb{E}_{s,a\sim \rho_\pi^\star}\left[\nabla_\theta^2\int_{\mathbb{R}\times \mathcal{S}}  p(r,s'\vert s,a,\theta)d(r,s')\vert_{\theta=\theta^\star_i}\right],\\
           &=\nabla_\theta^21\vert_{\theta=\theta^\star_i},\\
           &=0,
\end{align}
hence:
\begin{align}
           H(\theta^\star_i)&=0-\mathbb{E}_{s,a\sim \rho_\pi^\star,r,s'\sim P_{R,S}^\star(s,a)}\left[\frac{ \nabla_\theta p(r,s'\vert s,a,\theta_i^\star)}{p(r,s'\vert s,a,\theta_i^\star)}\frac{ \nabla_\theta p(r,s'\vert s,a,\theta_i^\star)^\top}{p(r,s'\vert s,a,\theta_i^\star)}\right],\\
           &=-\mathbb{E}_{s,a\sim \rho_\pi^\star,r,s'\sim P_{R,S}^\star(s,a)}\left[\nabla_\theta\log p(r,s'\vert s,a,\theta^\star_i) \nabla_\theta\log p(r,s'\vert s,a,\theta^\star_i)^\top\right],\\
           &=-\Sigma_i^g.
       \end{align}
        Using this result, each $\textrm{Tr}\left(\Sigma_i^g{H^\star_i}^{-1}\right)=\textrm{Tr}\left(-I\right)=-d$. Substituting y ields:
        \begin{align}
            \mathbb{E}_{\mathcal{D}_N\sim P_\textrm{Data}^\star}\left[\textnormal{\textrm{Regret}}(\mathcal{M}^\star,\mathcal{D}_N)\right]\le2\mathcal{R}_{\max}\cdot\sqrt{1-\exp\left(-C\frac{ (k+1)d}{(1-\gamma)N}\right)},\\
            \le2\mathcal{R}_{\max}\cdot\sqrt{1-\exp\left(-C'\frac{ d}{(1-\gamma)N}\right)},
        \end{align}
        for some $0<C'<\infty$ and sufficiently large $N$, as required.
    \end{proof}
\end{theorem}

We note that \cref{proof:info_rate} applies to the Gaussian world model introduced in \cref{sec:gauss_world} with neural network mean functions with $C^2$-continuous activations (tanh, identity, sigmoid, softplus, SiLU, SELU, GELU...) using a Gaussian or uniform prior truncated to a compact parameter space and similarly well-behaved parametric models. The resulting differences in performance only arise from the choice of prior, model representability and coverage of dataset, which affect Bayesian and frequentist methods equally. The information rate coincides with the optimal `minimax' convergence rate of frequentist parametric density estimators \citep{yang99, Bilodeau2021}. Similar results for the information rate have been found for nonparametric models such a Gaussian processes \citep{vanderVaart11}, which converge at slower rates. 

\begin{wrapfigure}{r}{0.5\textwidth} \vspace{-0.6cm}
  \begin{center}
    \includegraphics[width=0.5\textwidth]{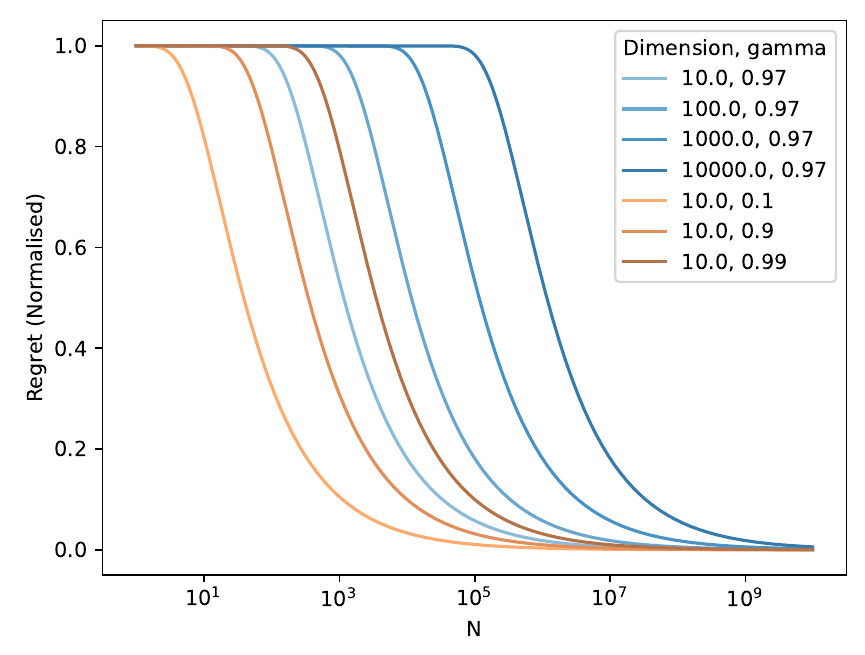}
  \end{center}\vspace{-0.2cm}
  \caption{Normalised Regret Curves for $C=1$}
  \label{fig:theoretical_regret_curves}
  \vspace{-0.2cm}
\end{wrapfigure}Using our result in \cref{proof:info_rate}, we plot the normalised regret bound (i.e. taking $\mathcal{R}_{\max}=0.5$) in Ineq.~\ref{eq_app:parametric_regret} for increasing dimensionality (blue) and decreasing $\gamma$ (copper) in \cref{fig:theoretical_regret_curves}. Our bound reveals an S-shaped curve with three distinct phases as number of data points $N$ increases: an initial plateau, a sudden decrease in regret follow by a slow exponential decay towards a regret of zero. The plateau indicates that a minimum amount of data is needed before any benefit can be realised in terms of regret. This is to be expected because initially the only information about the parameter values is given by the prior, which has no guarantee of accuracy under our analysis. Once a threshold of data points has been reached, the data can start to overwhelm the prior, resulting in a sudden decrease in regret. The higher the dimensionality of the model, the greater this data limit is - represented in \cref{fig:theoretical_regret_curves} by the plateau length increasing with greater $d$ (blue curves). Due to overspecification in models, this limit is likely to be set by the effective dimension of the problem (which may be much lower than $d$) as many parameters will be redundant, however the effective dimension is typically not possible to ascertain a priori. Finally, we observe that increasing the discount factor $\gamma$ leads to a longer regret plateau (copper curves) due to any error in the model dynamics being compounded over a longer horizon at test time. 

\subsection{Extensions for Model Misspecification and Sub-optimal Policy Learning}
\label{app:generalisation}
We now generalise our theorems to include the effects of model misspecification, that is models that cannot fully represent the true environment dynamics, and sub-optimal Bayesian policy learning, that is the effect of using a policy that does not fully optimise the Bayesian RL objective. We use the dagger notation to denote the maximum cross entropy parametrisation: 
\begin{align}
\theta^\dagger_i\in \argmax_{\theta\in\Theta}\ell(\theta)=\argmin_{\theta\in\Theta} \mathbb{E}_{s,a\sim \rho_\pi^\star,r,s'\sim P_{R,S}^\star(s,a)}\left[\textrm{KL}\left( P^\star_{R,S}(s,a)\vert P_{R,S}(s,a,\theta)\right)\right].
\end{align}
To characterise the degree of model misspecification, we use the KL divergence:
\begin{align}
    \epsilon_\textrm{miss}\coloneqq \min_{\theta\in\Theta} \mathbb{E}_{s,a\sim \rho_\pi^\star,r,s'\sim P_{R,S}^\star(s,a)}\left[\textrm{KL}\left( P^\star_{R,S}(s,a)\vert P_{R,S}(s,a,\theta)\right)\right].
\end{align}
We also introduce the following simplifying dagger notation for the expected cross entropy and corresponding gradients and Hessian under the optimal parameter:
\begin{align}
	\ell^\dagger &\coloneqq  \max_{\theta\in\Theta}\ell(\theta)=\mathbb{E}_{s,a\sim \rho_\pi^\star,r,s'\sim P_{R,S}^\star(s,a)}\left[\log p(r,s'\vert s,a,\theta_i^\dagger)\right],\\
    g^\dagger_{i,N}&\coloneqq\sqrt{N}\nabla_\theta\ell_N(\theta)\big\vert_{\theta=\theta^\dagger_i},\\
    H^\dagger_i&\coloneqq\nabla_\theta^2 \ell(\theta)\big\vert_{\theta=\theta^\dagger_i}.
\end{align}
We now relax \cref{ass:miss_model_regularity} to allow for model misspecification:
\begin{assumption}\label{ass:miss_model_regularity}
    We assume that:
    \begin{enumerate}[label=\Roman*.]
    \item The maximum likelihood is finite $\left\lvert\ell^\dagger\right\rvert<\infty$ and $\left\lvert\ell^\dagger-\ell(\theta) \right\rvert$ is bounded $P_\Theta$-almost surely.
        \item $\ell_N(\theta)$ and $\ell(\theta)$ are $C^2$-continuous in $\theta$.
    \item There are $K<\infty$  maximising points $\theta_i^\dagger$:
    \begin{align}
    	\{\theta_1^\dagger,\theta_2^\dagger,\dots \theta_K^\dagger\}=\argmax_{\theta\in\Theta}\ell(\theta).
    \end{align}
    For each maximiser $\theta^\dagger_i$, there exists a small region $\Theta^\dagger_i\coloneqq \{\theta\in\Theta\vert \lVert \theta^\dagger_i-\theta\rVert\le\epsilon \}  $ for some $\epsilon>0$ such that $\theta^\dagger_i$ is the unique maximiser in $\Theta_i^\dagger$,  $\theta^\dagger_i$ is in the interior of $\theta^\dagger_i$, $\nabla_\theta^2\ell(\theta^\dagger_i)$ is negative definite, invertible and the regions are disjoint: $\bigcap_{i=1}^K \Theta^\dagger_i=\varnothing$.

    \item The  prior $p(\theta)$ is Lipschitz continuous in $\theta$ with support over $\Theta$.
    \item The sampling regime ensures that the strong law of large numbers holds for all maximisers $\theta_i^\dagger$ for the Hessian, and uniformly for $\theta\in\Theta$ for the likelihood, that is:
    \begin{align}
      \ell_N(\theta) \xrightarrow{\textrm{Unif. } a.s.} \ell(\theta),\quad \nabla^2_\theta\ell_N(\theta^\dagger_i) \xrightarrow{a.s.} \nabla^2_\theta\ell(\theta^\dagger_i).
    \end{align}
    The central limit theorem applies to the gradient at each $\theta^\dagger_i$, that is:
    \begin{align}
    	\sqrt{N}\nabla_\theta\ell_N(\theta^\dagger_i)\xrightarrow{d}\mathcal{N}(0,\Sigma_i^g),
    \end{align}
    where $\Sigma_i^g=\mathbb{E}_{s,a\sim \rho_\pi^\star,r,s'\sim P_{R,S}^\star(s,a)}\left[\nabla_\theta\log p(r,s'\vert s,a,\theta^\dagger_i)\nabla_\theta\log p(r,s'\vert s,a,\theta^\dagger_i)^\top\right]$ with $\lVert\Sigma_i^g\rVert<\infty$.
    \end{enumerate}
\end{assumption}

Finally, we account for let the Bayes sub-optimality be defined as
\begin{align}
    \epsilon_\textrm{Bayes}\coloneqq \left\lvert J^{\pi^\star_\textrm{Bayes}}_\textrm{Bayes}(P_\Phi(\mathcal{D}_N)-J^{\hat{\pi}}_\textrm{Bayes}(P_\Phi(\mathcal{D}_N)\right\rvert. \label{eq:Bayes_suboptimality}
\end{align}

\begin{lemma} \label{proof:regret_bound_app_miss} Let $\mathcal{R}_{\max}\coloneqq \frac{(r_\textrm{max}-r_\textrm{min})}{1-\gamma}$ denote the maximum possible regret for the MDP and the Bayes sub-optimality $\epsilon_\textrm{Bayes}$ be defined as in \cref{eq:Bayes_suboptimality}. For a prior $P_\Theta(\mathcal{D}_N)$, the true regret can be bounded as:
\begin{align}
\textnormal{\textrm{Regret}}(\mathcal{M}^\star,\mathcal{D}_N)&\le2\mathcal{R}_{\max}\cdot \sup_\pi \mathbb{E}_{i\sim\mathcal{G}(\gamma)}\left[\textrm{TV}\left(P_{i+1,\pi}^\star\Vert P_{i+1}^\pi( \mathcal{D}_N)\right)\right]+\epsilon_\textnormal{\textrm{Bayes}}.
\end{align}
\begin{proof}
We start from the definition of the true regret under Bayes sub-optimality:
\begin{align}
    \textnormal{\textrm{Regret}}(\mathcal{M}^\star,\mathcal{D}_N)&\coloneqq J^{\pi^\star}(\mathcal{M}^\star)- J^{\hat{\pi}}(\mathcal{M}^\star,\mathcal{D}_N),\\
 &=J^{\pi^\star}(\mathcal{M}^\star)-J^{\pi^\star}_\textrm{Bayes}(P_\Phi(\mathcal{D}_N))+J^{\pi^\star}_\textrm{Bayes}(P_\Phi(\mathcal{D}_N))-J^{\hat{\pi}}(\mathcal{M}^\star,\mathcal{D}_N),\\
   &\le J^{\pi^\star}(\mathcal{M}^\star)-J^{\pi^\star}_\textrm{Bayes}(P_\Phi(\mathcal{D}_N))+J^{\pi^\star_\textrm{Bayes}}_\textrm{Bayes}(P_\Phi(\mathcal{D}_N))-J^{\hat{\pi}}(\mathcal{M}^\star,\mathcal{D}_N),\\
   &=J^{\pi^\star}(\mathcal{M}^\star)-J^{\pi^\star}_\textrm{Bayes}(P_\Phi(\mathcal{D}_N))+J^{\hat{\pi}}_\textrm{Bayes}(P_\Phi(\mathcal{D}_N))-J^{\hat{\pi}}(\mathcal{M}^\star,\mathcal{D}_N)\\
   &\quad+J^{\pi^\star_\textrm{Bayes}}_\textrm{Bayes}(P_\Phi(\mathcal{D}_N))-J^{\hat{\pi}}_\textrm{Bayes}(P_\Phi(\mathcal{D}_N)),\\
    &\le \sup_\pi \left\lvert J^{\pi}(\mathcal{M}^\star)-J^\pi_\textrm{Bayes}(P_\Theta(\mathcal{D}_N))\right\rvert+\sup_\pi \left\lvert J^\pi_\textrm{Bayes}(P_\Theta(\mathcal{D}_N)) -J^{\pi}(\mathcal{M}^\star)\right\rvert\\
    &\quad +\left\lvert J^{\pi^\star_\textrm{Bayes}}_\textrm{Bayes}(P_\Phi(\mathcal{D}_N)-J^{\hat{\pi}}_\textrm{Bayes}(P_\Phi(\mathcal{D}_N)\right\rvert,\\
    &=2 \sup_\pi \left\lvert  J^{\pi}(\mathcal{M}^\star)-J^\pi_\textrm{Bayes}(P_\Theta(\mathcal{D}_N))\right\rvert+\epsilon_\textrm{Bayes}, \label{eq:miss_true_regret_bayes}
\end{align}
where the second line follows from $J^{\pi^\star}_\textrm{Bayes}(P_\Phi(\mathcal{D}_N))\le J^{\pi^\star_\textrm{Bayes}}_\textrm{Bayes}(P_\Phi(\mathcal{D}_N))$ by definition. We then bound $ \left\lvert  J^{\pi}(\mathcal{M}^\star)-J^\pi_\textrm{Bayes}(P_\Theta(\mathcal{D}_N))\right\rvert$ using \cref{proof:regret_bound_app} to obtain our desired result.
\end{proof}
\end{lemma}

\begin{theorem} \label{app:miss_proof_info_rate} Let the data be drawn from the underlying true distribution $\mathcal{D}_N\sim P_\textrm{Data}^\star$.
    Under \cref{ass:miss_model_regularity}, there exists some constant $0<C<\infty$ such that for sufficiently large $N$: 
    \begin{align}
    \mathbb{E}_{\mathcal{D}_N\sim P_\textrm{Data}^\star}\left[\textnormal{\textrm{Regret}}(\mathcal{M}^\star,\mathcal{D}_N)\right] \le 2\mathcal{R}_{\max}\cdot\exp\left(1-\sqrt{C\left(\frac{d}{(1-\gamma)N}+\frac{\epsilon_\textnormal{\textrm{miss}}}{1-\gamma}\right)}\right)+\epsilon_{\textnormal{\textrm{Bayes}}}.     \label{eq_app_miss:parametric_regret}
    \end{align}

    \begin{proof}
    Starting with \cref{proof:regret_bound_app_miss}, we obtain: 
    \begin{align}
    \textnormal{\textrm{Regret}}(\mathcal{M}^\star,\mathcal{D}_N)&\le2\mathcal{R}_{\max}\cdot\sqrt{1-\exp\left(-C\left(\frac{ d^2}{(1-\gamma)N}+\frac{ \epsilon_\textrm{miss}}{(1-\gamma)}\right)\right)}+\epsilon_\textnormal{\textrm{Bayes}}.
    \end{align}
    Next we apply \cref{app:proof_PIL} to bound the first term, obtaining: 
    \begin{align}
        \textnormal{\textrm{Regret}}(\mathcal{M}^\star,\mathcal{D}_N)&\le2 \mathcal{R}_{\max}\cdot\sup_\pi\sqrt{1-\exp\left(-\frac{\mathcal{I}_N^\pi}{1-\gamma}\right)}+\epsilon_\textnormal{\textrm{Bayes}}.
    \end{align}
  We write the PIL to include misspecification  as:
    \begin{align}
 \mathcal{I}_N^\pi\coloneqq&\mathbb{E}_{s,a\sim \rho_\pi^\star}\left[ \mathbb{E}_{\theta\sim P_\Theta(\mathcal{D}_N)}\left[\textrm{KL}\left(P_{R,S}^\star(s,a)\Vert P_{R,S}(s,a,\theta) \right)\right]\right],\\
 =&\mathbb{E}_{\theta\sim P_\Theta(\mathcal{D}_N)}\left[\mathbb{E}_{s,a\sim \rho_\pi^\star,r,s'\sim P_{R,S}^\star(s,a)}\left[\log p(r,s'\vert s,a,\theta^\star)-\log p(r,s'\vert s,a,\theta)\right]\right],\\
 =&\mathbb{E}_{\theta\sim P_\Theta(\mathcal{D}_N)}\Big[\mathbb{E}_{s,a\sim \rho_\pi^\star,r,s'\sim P_{R,S}^\star(s,a)}[\log p(r,s'\vert s,a,\theta^\star)-\log p(r,s'\vert s,a,\theta^\dagger_i)\\
 &\qquad+\log p(r,s'\vert s,a,\theta^\dagger_i)-\log p(r,s'\vert s,a,\theta)]\Big],\\
 =&\mathbb{E}_{\theta\sim P_\Theta(\mathcal{D}_N)}\Big[\mathbb{E}_{s,a\sim \rho_\pi^\star,r,s'\sim P_{R,S}^\star(s,a)}\left[\log p(r,s'\vert s,a,\theta^\star)-\log p(r,s'\vert s,a,\theta^\dagger_i)\right]\\
 &\qquad+\mathbb{E}_{s,a\sim \rho_\pi^\star,r,s'\sim P_{R,S}^\star(s,a)}\left[\log p(r,s'\vert s,a,\theta^\dagger_i)-\log p(r,s'\vert s,a,\theta)]\right]\Big],\\
 =&\mathbb{E}_{s,a\sim \rho_\pi^\star,r,s'\sim P_{R,S}^\star(s,a)}\left[\log p(r,s'\vert s,a,\theta^\star)-\log p(r,s'\vert s,a,\theta^\dagger_i)\right]\\
 &\qquad+\mathbb{E}_{\theta\sim P_\Theta(\mathcal{D}_N)}\left[\mathbb{E}_{s,a\sim \rho_\pi^\star,r,s'\sim P_{R,S}^\star(s,a)}\left[\log p(r,s'\vert s,a,\theta^\dagger_i)-\log p(r,s'\vert s,a,\theta)]\right]\right],\\
  =&\mathbb{E}_{s,a\sim \rho_\pi^\star,r,s'\sim P_{R,S}^\star(s,a)}\left[\textrm{KL}\left( P^\star_{R,S}(s,a)\vert P_{R,S}(s,a,\theta_i^\dagger)\right)\right]\\
 &\qquad+\mathbb{E}_{\theta\sim P_\Theta(\mathcal{D}_N)}\left[\mathbb{E}_{s,a\sim \rho_\pi^\star,r,s'\sim P_{R,S}^\star(s,a)}\left[\log p(r,s'\vert s,a,\theta^\dagger_i)-\log p(r,s'\vert s,a,\theta)]\right]\right],\\
   =&\mathbb{E}_{\theta\sim P_\Theta(\mathcal{D}_N)}\left[\ell^\dagger-\ell(\theta)\right]+\mathbb{E}_{s,a\sim \rho_\pi^\star,r,s'\sim P_{R,S}^\star(s,a)}\left[\textrm{KL}\left( P^\star_{R,S}(s,a)\vert P_{R,S}(s,a,\theta_i^\dagger)\right)\right],\\
   =&\mathbb{E}_{\theta\sim P_\Theta(\mathcal{D}_N)}\left[\ell^\dagger-\ell(\theta)\right]+\min_\theta\mathbb{E}_{s,a\sim \rho_\pi^\star,r,s'\sim P_{R,S}^\star(s,a)}\left[\textrm{KL}\left( P^\star_{R,S}(s,a)\vert P_{R,S}(s,a,\theta)\right)\right],\\
   =&\mathbb{E}_{\theta\sim P_\Theta(\mathcal{D}_N)}\left[\ell^\dagger-\ell(\theta)\right]+\epsilon_\textrm{miss}.
    \end{align}
   To bound the first term $\mathbb{E}_{\theta\sim P_\Theta(\mathcal{D}_N)}\left[\ell^\dagger-\ell(\theta)\right]$, we follow the remainder of the proof of \cref{app:proof_info_rate} to Ineq.~\ref{eq:raw_hessian_form}, replacing $\star$ notion with $\dagger$ to yield:
\begin{align}
	\mathcal{I}_N^\pi&=\epsilon_\textrm{miss}+\mathcal{O}\left(\frac{d- \sum_{i=1}^K{g^\dagger_{i,N}}^\top{H^\dagger_i}^{-1}g^\dagger_{i,N}}{N}\right).\label{eq:miss_pil_asymptotic_bound}
\end{align}
almost surely. As $f(x  )\coloneqq2\mathcal{R}_{\max}\cdot\sqrt{1-\exp\left(-\frac{x}{(1-\gamma)}\right)}$ is monotonic in $x$ and $\frac{d- \sum_{i=1}^K{g^\dagger_{i,N}}^\top{H^\dagger_i}^{-1}g^\dagger_{i,N}}{N}+\epsilon_\textrm{miss}\ge0$, \cref{eq:miss_pil_asymptotic_bound} implies there exists some positive $0<C<\infty$ such that:
\begin{align}
    \textnormal{\textrm{Regret}}(\mathcal{M}^\dagger,\mathcal{D}_N)\le 2\mathcal{R}_{\max}\cdot\sqrt{1-\exp\left(-C\left(\frac{ d- \sum_{i=1}^K{g^\dagger_{i,N}}^\top{H^\dagger_i}^{-1}g^\dagger_{i,N}}{(1-\gamma)N}+\frac{ \epsilon_\textrm{miss}}{(1-\gamma)}\right)\right)},
\end{align}
almost surely for large enough $N$. Under \cref{ass:miss_model_regularity}, $g^\dagger_{i,N}\xrightarrow{d}\mathcal{N}(0,\Sigma_i^g)$. As $f(x)$ is also a bounded, continuous function and concave, we can apply the Portmanteau Theorem (see for example \citet[Chapter 21.7]{Bass13}) followed by Jensen's inequality to yield:
       \begin{align}
       	&\mathbb{E}_{\mathcal{D}_N\sim P_\textrm{Data}^\dagger}\left[\textnormal{\textrm{Regret}}(\mathcal{M}^\dagger,\mathcal{D}_N)\right]\\
        &\le2\mathcal{R}_{\max}\cdot\mathbb{E}_{g_i\sim \mathcal{N}(0,\Sigma_i^g)}\left[\sqrt{1-\exp\left(-C\left(\frac{d- \sum_{i=1}^K{g_i}^\top{H^\dagger_i}^{-1}g_i}{(1-\gamma)N}++\frac{ \epsilon_\textrm{miss}}{(1-\gamma)}\right)\right)}\right],\\
        \le&2\mathcal{R}_{\max}\cdot\sqrt{1-\exp\left(-C\left(\frac{ d-\sum_{i=1}^K\mathbb{E}_{g_i\sim \mathcal{N}(0,\Sigma_i^g)}\left[{g}^\top{H^\dagger_i}^{-1}g\right]}{(1-\gamma)N}+\frac{ \epsilon_\textrm{miss}}{(1-\gamma)}\right)\right)},\\
        =&2\mathcal{R}_{\max}\cdot\sqrt{1-\exp\left(-C\left(\frac{ d-\sum_{i=1}^K\textrm{Tr}\left(\Sigma_i^g{H^\dagger_i}^{-1}\right)}{(1-\gamma)N}+\frac{ \epsilon_\textrm{miss}}{(1-\gamma)}\right)\right)}.
       \end{align}
        Now $\textrm{Tr}\left(\Sigma_i^g{H^\star_i}^{-1}\right)=\mathcal{O}(d^2)$, hence
        \begin{align}
            \mathbb{E}_{\mathcal{D}_N\sim P_\textrm{Data}^\dagger}\left[\textnormal{\textrm{Regret}}(\mathcal{M}^\dagger,\mathcal{D}_N)\right]\le2\mathcal{R}_{\max}\cdot\sqrt{1-\exp\left(-C\left(\frac{ (k+1)d^2}{(1-\gamma)N}+\frac{ \epsilon_\textrm{miss}}{(1-\gamma)}\right)\right)},\\
            \le2\mathcal{R}_{\max}\cdot\sqrt{1-\exp\left(-C'\left(\frac{ d^2}{(1-\gamma)N}+\frac{ \epsilon_\textrm{miss}}{(1-\gamma)}\right)\right)},
        \end{align}
        for some $0<C'<\infty$ and sufficiently large $N$, as required.
    \end{proof}
\end{theorem}

Taking the limit $N\rightarrow \infty$, we see the residual term due to misspecification and sub-optimal policy learning is: 
\begin{align}
   \mathbb{E}_{\mathcal{D}_N\sim P_\textrm{Data}^\dagger}\left[\textnormal{\textrm{Regret}}(\mathcal{M}^\dagger,\mathcal{D}_N)\right]\le 2\mathcal{R}_{\max}\cdot\sqrt{1-\exp\left(-\frac{\epsilon_\textnormal{\textrm{miss}}C}{1-\gamma}\right)}+\epsilon_{\textnormal{\textrm{Bayes}}}.
\end{align}
We compare against prior work such as MOReL \citep[Corollary 2]{Kidambi20}, where the residual misspecification term is:
\begin{align}
  \frac{4\gamma\mathcal{R}_{\max}}{1-\gamma}\epsilon_{TV}
\end{align}
where $\epsilon_{TV}$ characterises the misspecification in terms of total variational distance instead of KL divergence of our method. Crucially, we see that our bound is much less sensitive to $\gamma$; our bound is $\mathcal{O}(\frac{1}{\sqrt{1-\gamma}})$ in comparison to $\mathcal{O}(\frac{\gamma}{1-\gamma})$ of MOReL meaing our bound is tighter as $\gamma\rightarrow 1$.

\newpage
\section{Further Results}\vspace{-0.25em}\label{app:further_results}
\subsection{SOReL}
\label{app:sorel_further_results}
\begin{figure}[H]
    \centering
    \includegraphics[width=0.8\linewidth]{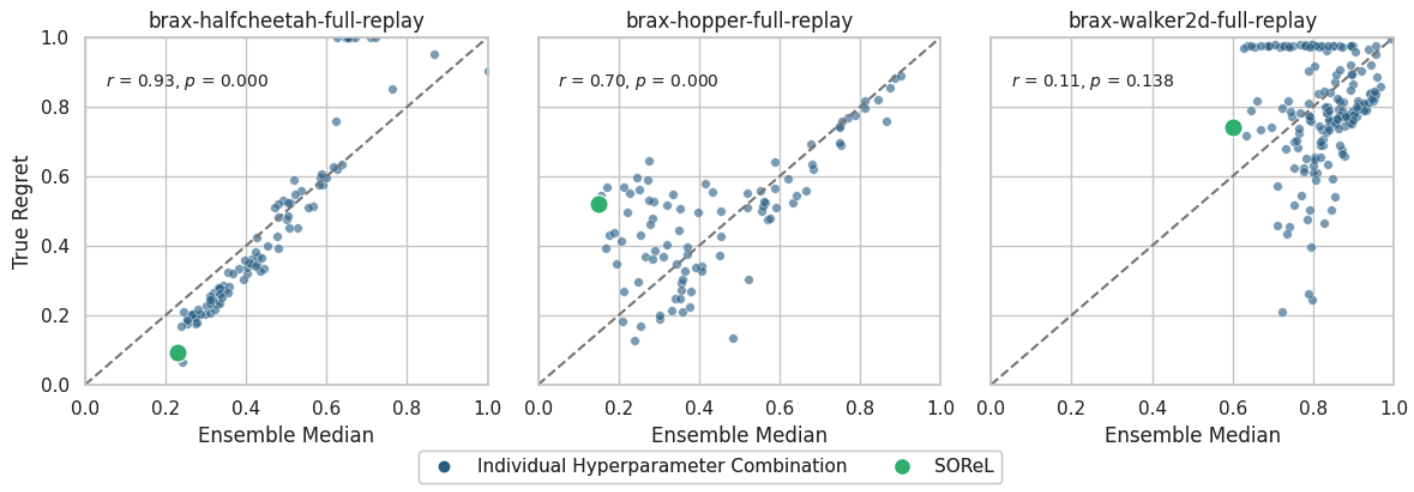}
    \caption{\textbf{SOReL BAMDP hyperparameter tuning} for 200,000 randomly sampled transitions of the Brax datasets. SOReL selects the BAMDP hyperparameters that yield the lowest approximate regret (green). Figure corresponds to tuning set \textcolor{RedOrange}{$\phi_{III}$}) in \cref{alg:sorel}.}
    \label{fig:sorel_hyper_sweep}
\end{figure}

In practice, each time additional offline data is incorporated, the model, approximate inference and BAMDP hyperparameters should be newly tuned. To avoid too high a computational burden in our experiments, we use fixed model and approximate inference hyperparameters, highlighting in red the region where the approximate regret may be unreliable, and only tune the BAMDP hyperparameters using the approximate regret for one seed and offline dataset size (\cref{fig:sorel_hyper_sweep} in \cref{app:sorel_further_results}).

\begin{figure}[H]
    \centering
    \includegraphics[width=\linewidth]{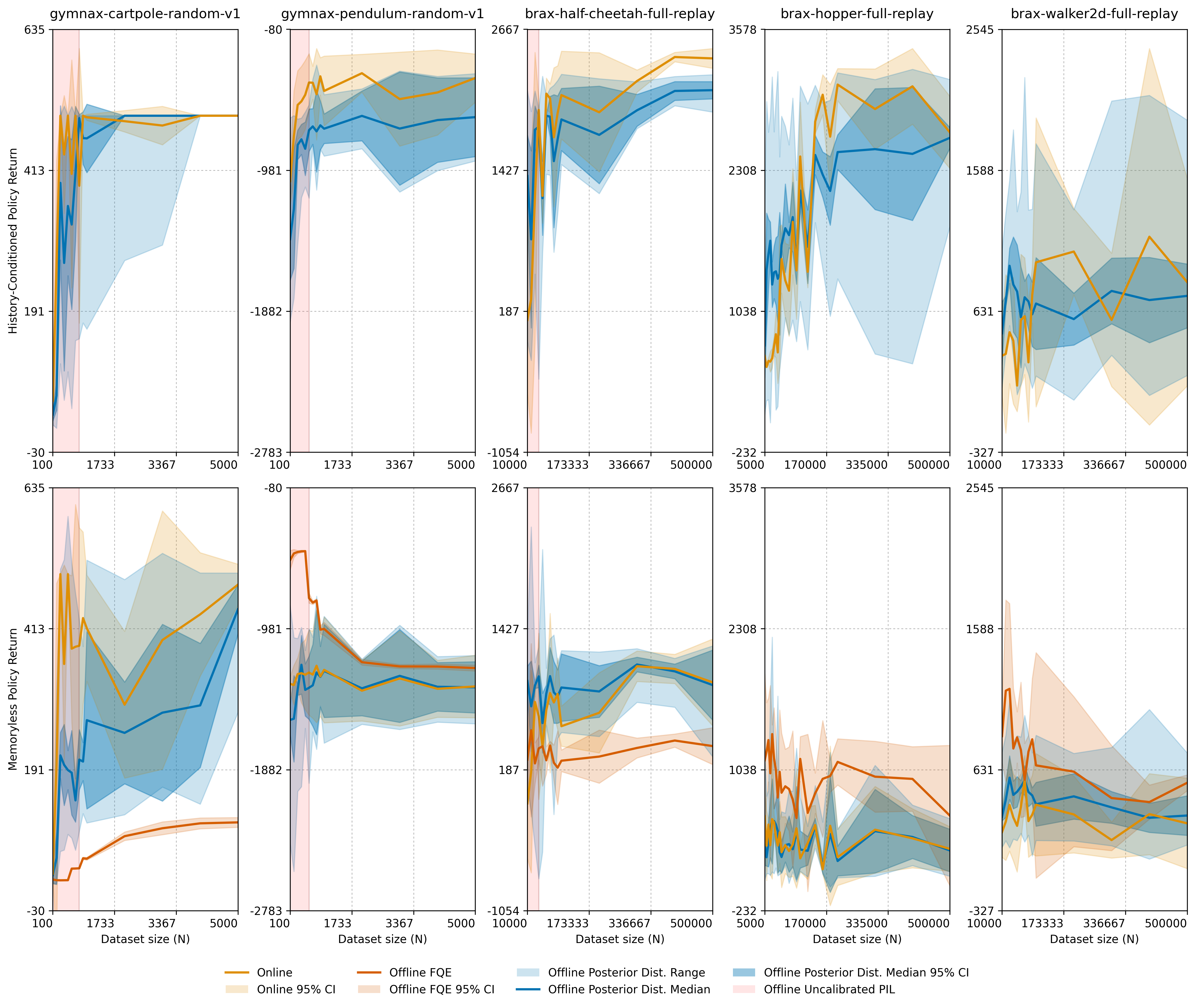}
    \caption{SOReL applied to various tasks (\textbf{top}). Approximating the value of a policy for various tasks (\textbf{bottom}) (we use the recurrent policy trained using SOReL, but with the history set to 0 across all timesteps (to allow comparison with FQE, since the offline dataset does not contain trajectories)). Light shaded blue shows the \textbf{offline approximate return range} across rollouts with environments sampled from the posterior. Dark shaded blue shows the \textbf{offline standard deviation of the median rollout} sampled from the posterior. Shaded red indicates where the \textbf{PIL is poorly calibrated} $(\mathcal{D}_N,\mathcal{M}^\star)\not\approx\mathcal{V}(\mathcal{D}_N)$ (for a threshold of $0.25$), and hence the approximate return may be unreliable. Importantly, a practitioner is aware of this offline. 95\% CI over 3 seeds.}
    \label{fig:sorel_5_panel}
\end{figure}

\newpage
\subsection{TOReL}\label{app:torel_further_results}
\begin{table}[H]
\centering
\small
\renewcommand{\arraystretch}{1.15}
\begin{tabular}{llccccc}
\toprule
\textbf{Dataset} & \textbf{Alg.} 
& \textbf{TOReL} 
& \textbf{FQE} 
& \textbf{Dual DICE} 
& \textbf{DR} 
& \textbf{IS} \\
\midrule
\multirow{4}{*}{\makecell[l]{\textbf{brax-}\\\textbf{half-cheetah-}\\\textbf{full-}\\\textbf{replay}}}
& IQL 
& 0.29 (0.45)
& 
& 
& 
&  \\
& ReBRAC 
& \textbf{0.92 (0.00)}
&  
& 
& 
& \\
& MOPO 
& \textbf{0.98 (0.00)}
& 
& 
& 
&  \\
& MOReL
& \textbf{0.93 (0.00)}
&  
& 
& 
& \\
\multirow{4}{*}{\makecell[l]{\textbf{brax-}\\\textbf{hopper-}\\\textbf{full-}\\\textbf{replay}}}
& IQL 
& 0.18 (0.635)
& 
& 
& 
&  \\
& ReBRAC 
& \textbf{0.98 (0.00)}
&  
& 
& 
& \\
& MOPO 
& \textbf{1.00 (0.00)}
& 
& 
& 
&  \\
& MOReL
& \textbf{0.98 (0.00)}
&  
& 
& 
& \\
\multirow{4}{*}{\makecell[l]{\textbf{brax-}\\\textbf{walker2d-}\\\textbf{full-}\\\textbf{replay}}}
& IQL 
& 0.32 (0.41)
& 
& 
& 
&  \\
& ReBRAC 
& \textbf{0.81 (0.02)}
&  
& 
& 
& \\
& MOPO 
& -0.68 (0.13)
& 
& 
& 
&  \\
& MOReL
& \textbf{0.99 (0.00)}
&  
& 
& 
& \\
\midrule
\multirow{4}{*}{\makecell[l]{\textbf{d4rl-}\\\textbf{half-cheetah-}\\\textbf{medium-}\\\textbf{expert-v2}}}
& IQL 
& 0.29 (0.44)
& -0.21 (0.59) 
& -0.47 (0.20) 
& -0.24 (0.54) 
& -0.33 (0.39) \\
& ReBRAC 
& \textbf{0.90 (0.00)} 
& \textbf{0.66 (0.00)} 
& 0.19 (0.17) 
& 0.32 (0.02) 
& -0.44 (0.01) \\
& MOPO 
& 0.02 (0.98)
& 
& 
& 
&  \\
& MOReL
& \textbf{0.98 (0.00)}
&  
& 
& 
& \\
\addlinespace
\multirow{4}{*}{\makecell[l]{\textbf{d4rl-}\\\textbf{hopper-}\\\textbf{medium-}\\\textbf{v2}}}
& IQL 
& 0.29 (0.44) 
& 0.16 (0.67) 
& 0.19 (0.62) 
& 0.26 (0.50) 
& -0.39 (0.30) \\
& ReBRAC 
& \textbf{0.98 (0.00)} 
& 0.27 (0.04) 
& 0.34 (0.01) 
& 0.20 (0.16) 
& 0.27 (0.04) \\
& MOPO 
& \textbf{0.98 (0.00)}
& 
& 
& 
&  \\
& MOReL
& \textbf{0.94 (0.00)}
&  
& 
& 
& \\
\addlinespace
\multirow{4}{*}{\makecell[l]{\textbf{d4rl-}\\\textbf{walker2d-}\\\textbf{medium-}\\\textbf{replay-v2}}}
& IQL 
& \textbf{0.65 (0.05)}
& 0.41 (0.28) 
& 0.52 (0.15) 
& 0.19 (0.63) 
& -0.06 (0.88) \\
& ReBRAC 
& \textbf{0.78 (0.00)} 
& \textbf{0.89 (0.00)} 
& -0.04 (0.700) 
& \textbf{0.82 (0.00)}
& \textbf{0.86 (0.00)} \\
& MOPO 
& \textbf{1.00 (0.00)}
& 
& 
& 
&  \\
& MOReL
& -0.53 (0.01)
&  
& 
& 
& \\
\addlinespace
\multirow{2}{*}{\makecell[l]{\textbf{d4rl-pen-}\\\textbf{expert-v1}}}
& IQL 
& 0.00 (0.30) 
& -0.00 (0.40) 
& 0.49 (0.18) 
& 0.51 (0.23)
& 0.26 (0.50) \\
& ReBRAC 
& \textbf{0.89 (0.01)} 
& \textbf{0.91 (0.00)}  
& -0.01 (0.98) 
& 0.26 (0.06) 
& -0.66 (0.00) \\
\addlinespace
\multirow{2}{*}{\makecell[l]{\textbf{d4rl-hammer-}\\\textbf{expert-v1}}}
& IQL 
& \textbf{0.70 (0.04)} 
& \textbf{0.74 (0.02)} 
& -0.12 (0.76) 
& -0.35 (0.35) 
& -0.40 (0.29) \\
& ReBRAC 
& 0.30 (0.02) 
& 0.11 (0.42) 
& -0.14 (0.31) 
& -0.04 (0.75) 
& -0.66 (0.00) \\
\bottomrule
\end{tabular}
\caption{Pearson correlation $r$ (with $p$-value in parentheses) between approximate offline and true online returns. 
\textbf{Bold} indicates \textbf{strong} ($|r|>0.50$) and \textbf{statistically significant} ($p<0.05$) correlation.}
\label{tab:fpe_main}
\end{table}

\newpage
\begin{figure}
    \centering
    \begin{subfigure}{\linewidth}
        \includegraphics[width=\linewidth]{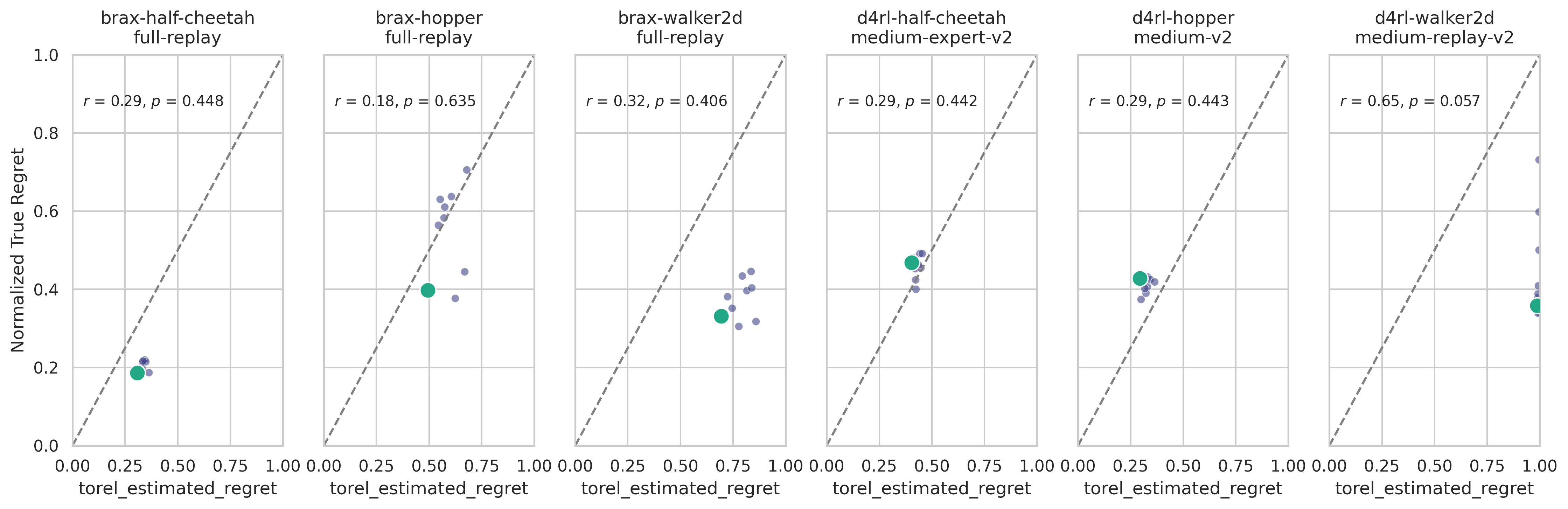}
    \end{subfigure}
    \begin{subfigure}{\linewidth}
        \includegraphics[width=\linewidth]{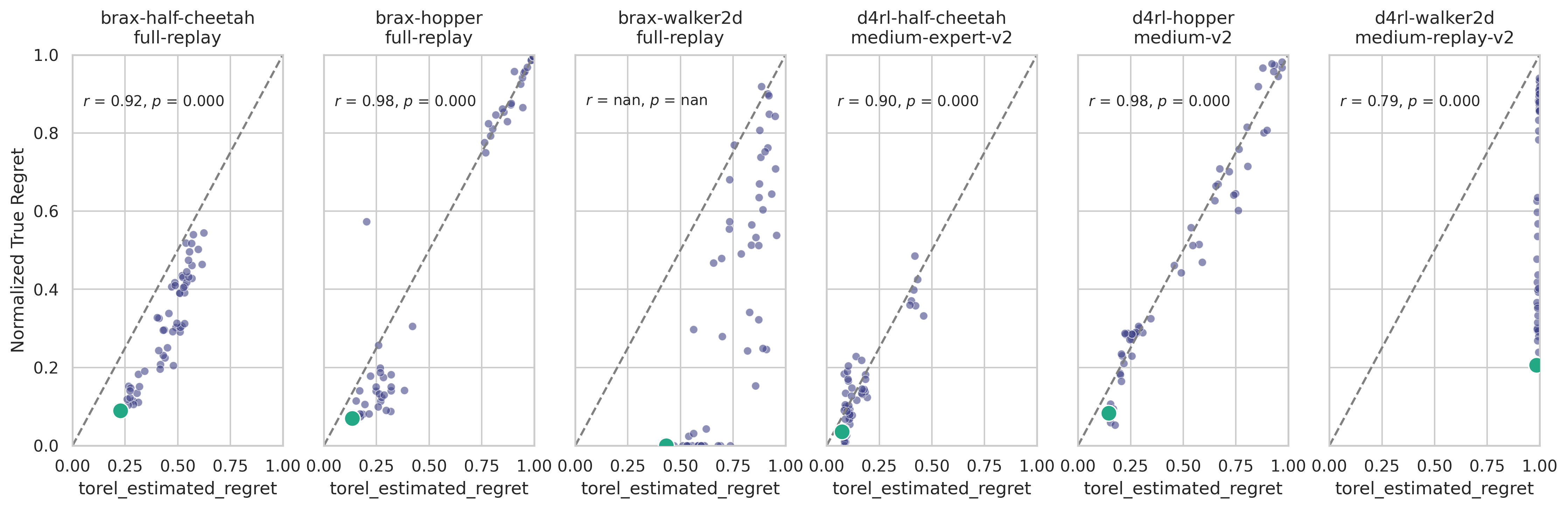}
    \end{subfigure}
    \begin{subfigure}
        {\linewidth}
        \includegraphics[width=\linewidth]{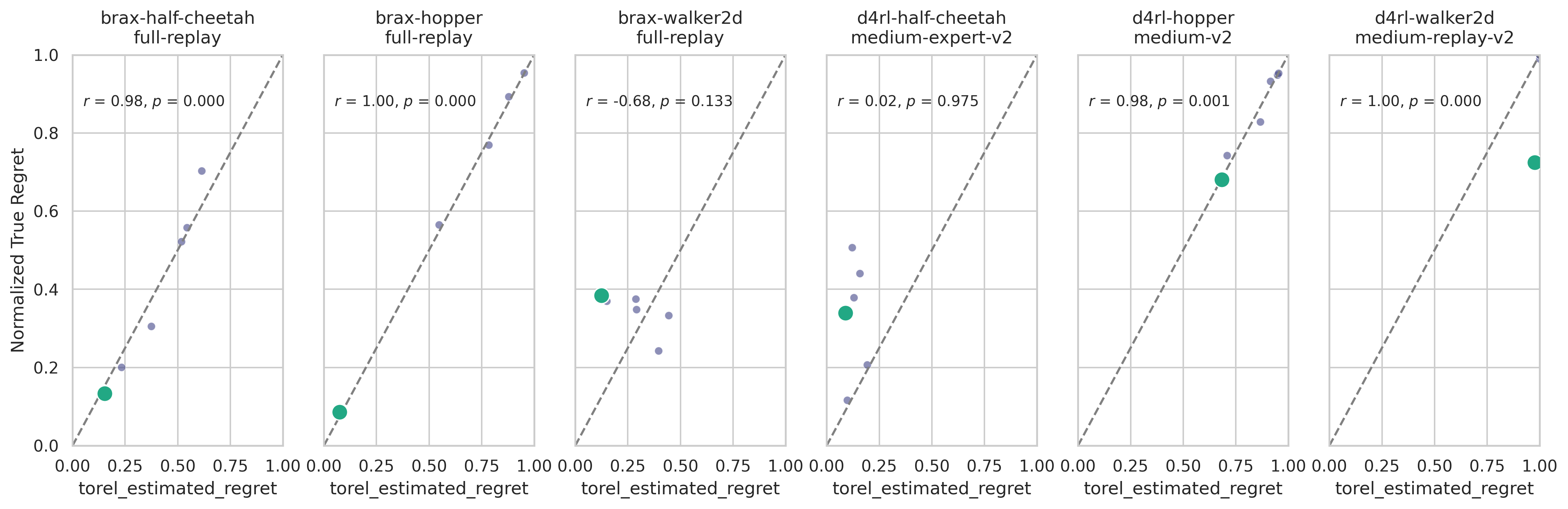}
    \end{subfigure}
    \begin{subfigure}
        {\linewidth}
        \includegraphics[width=\linewidth]{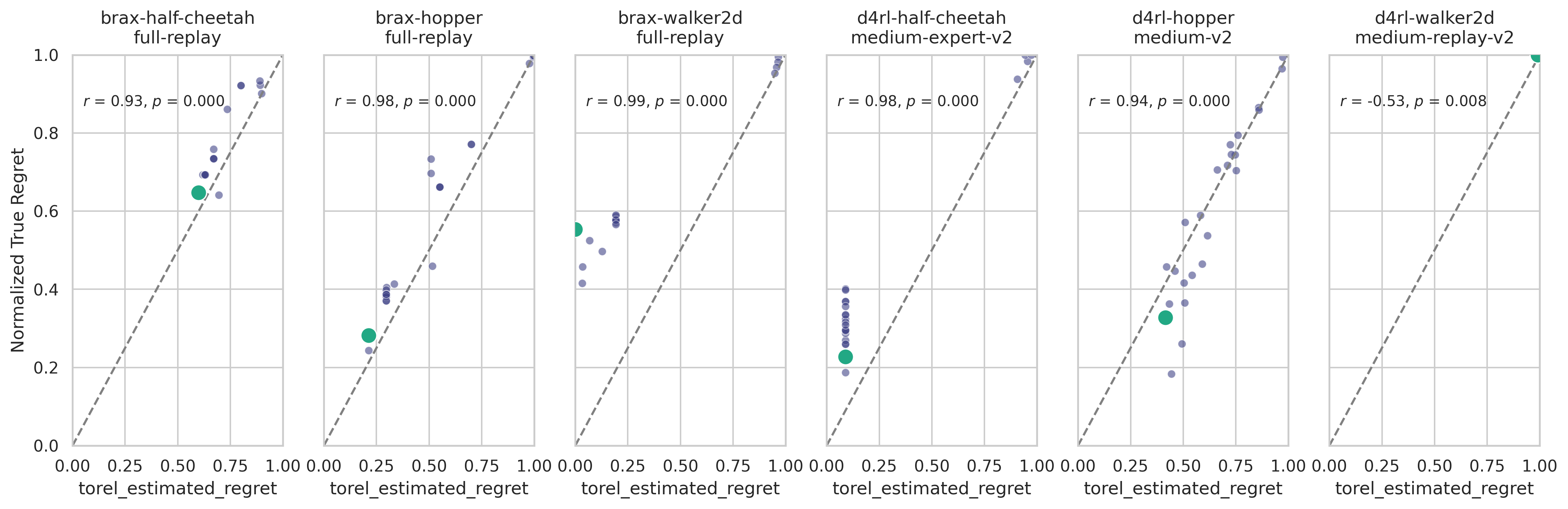}
    \end{subfigure}
    \caption{\textbf{IQL}, \textbf{ReBRAC}, \textbf{MOPO} and \textbf{MOReL} (\textbf{left} to \textbf{right}) for brax-half-cheetah-full-replay, brax-hopper-full-replay, brax-walker2d-full-replay, d4rl-half-cheetah-medium-expert-v2, d4rl-hopper-medium-v2, d4rl-walker2d-medium-replay-v2, (\textbf{left} to \textbf{right}). Scatter plots (1) comparing TOReL's \textbf{offline estimated performance} and \textbf{online true performance} across datasets and algorithms.}
    \label{fig:placeholder}
\end{figure}

\clearpage

\begin{figure}
    \centering
    \begin{subfigure}{\linewidth}
        \includegraphics[width=\linewidth]{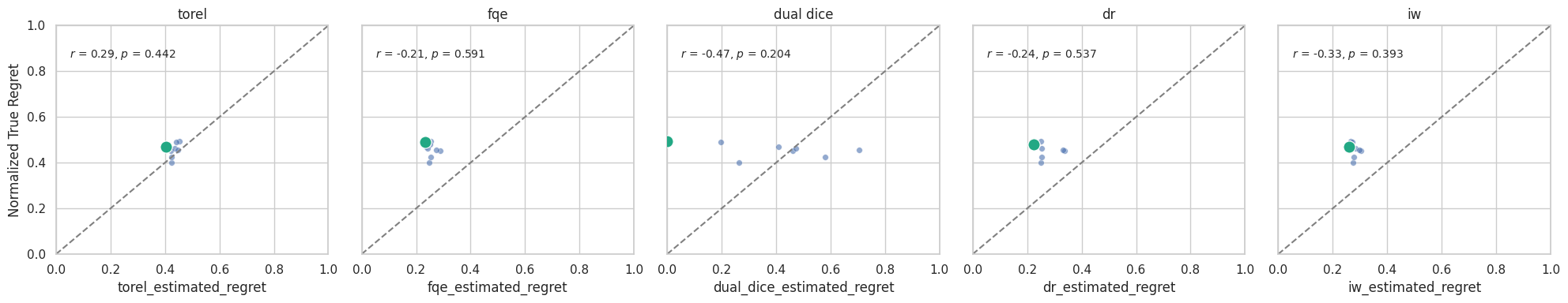}
    \end{subfigure}
    \begin{subfigure}{\linewidth}
        \includegraphics[width=\linewidth]{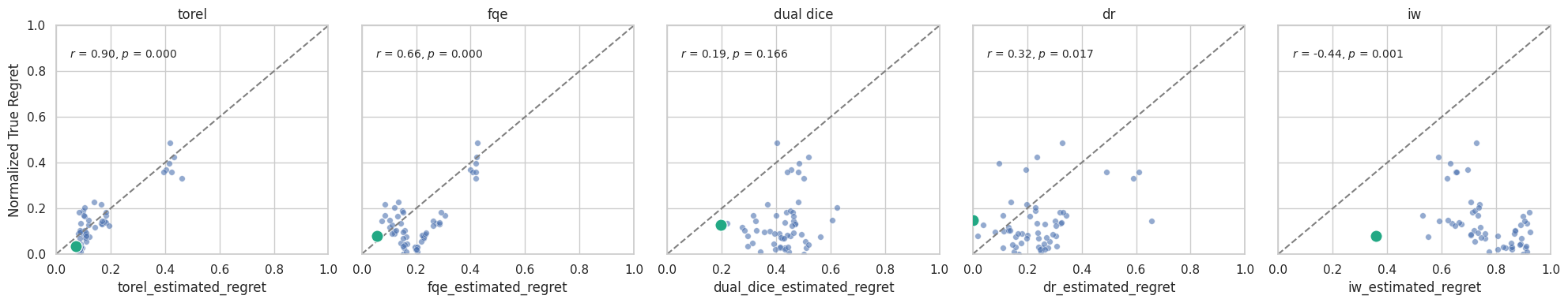}
        \caption{\textbf{IQL} (\textbf{top}) and \textbf{ReBRAC} (\textbf{bottom}), d4rl-half-cheetah-medium-expert-v2}
    \end{subfigure}
    \vspace{0.5em}
    \begin{subfigure}{\linewidth}
        \includegraphics[width=\linewidth]{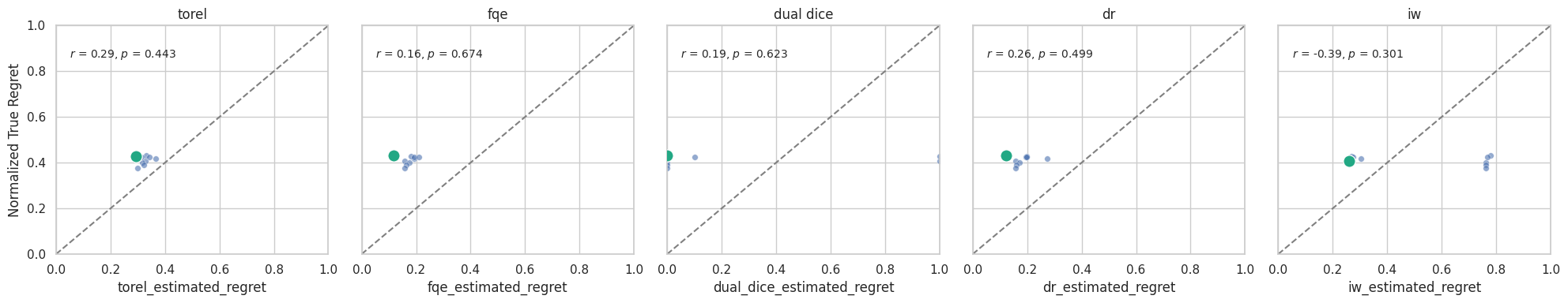}
    \end{subfigure}
    \begin{subfigure}{\linewidth}
        \includegraphics[width=\linewidth]{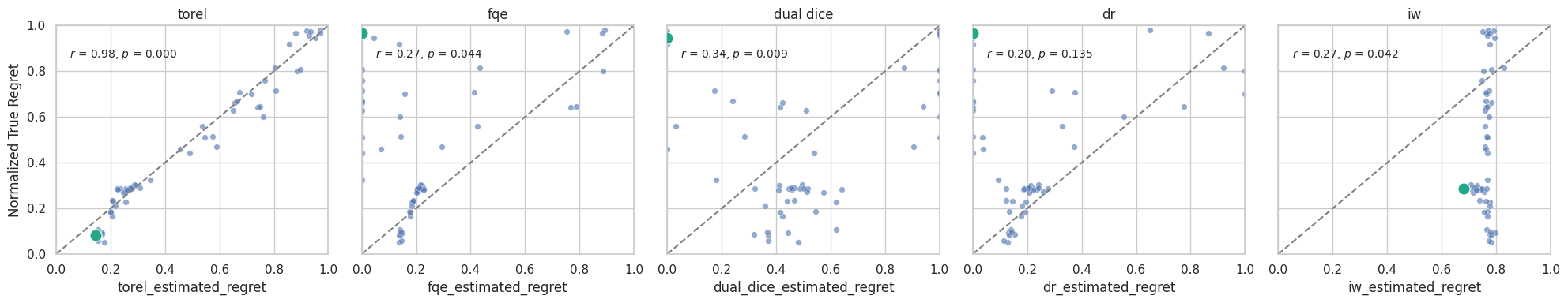}
        \caption{\textbf{IQL} (\textbf{top}) and \textbf{ReBRAC} (\textbf{bottom}), d4rl-hopper-medium-v2}
    \end{subfigure}
    \vspace{0.5em}
    \begin{subfigure}{\linewidth}
        \includegraphics[width=\linewidth]{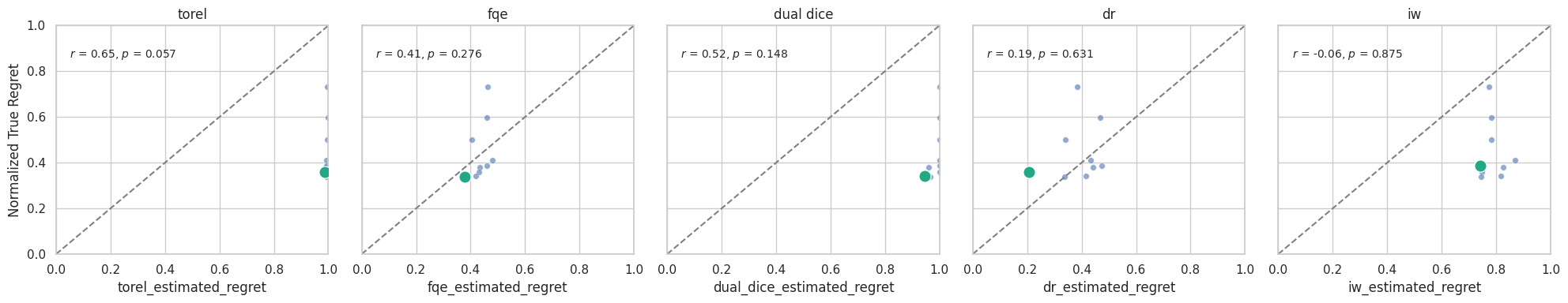}
    \end{subfigure}
    \begin{subfigure}{\linewidth}
        \includegraphics[width=\linewidth]{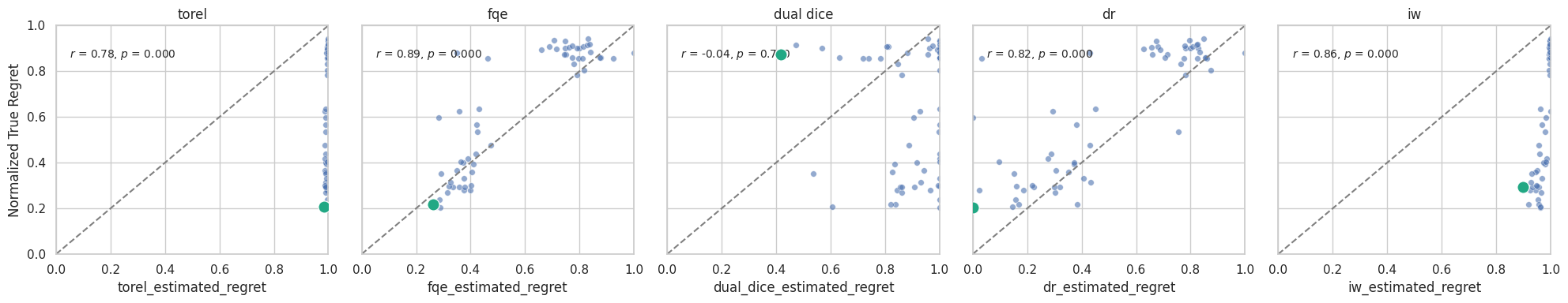}
        \caption{\textbf{IQL} (\textbf{top}) and \textbf{ReBRAC} (\textbf{bottom}), d4rl-walker2d-medium-replay-v2}
    \end{subfigure}
    \caption{Scatter plots (2) comparing different OPE algorithms' \textbf{offline estimated performance} and \textbf{online true performance} across datasets and algorithms.}
    \label{fig:ope_all_vertical_1}
\end{figure}

\newpage

\begin{figure}
    \centering
    \begin{subfigure}{\linewidth}
        \includegraphics[width=\linewidth]{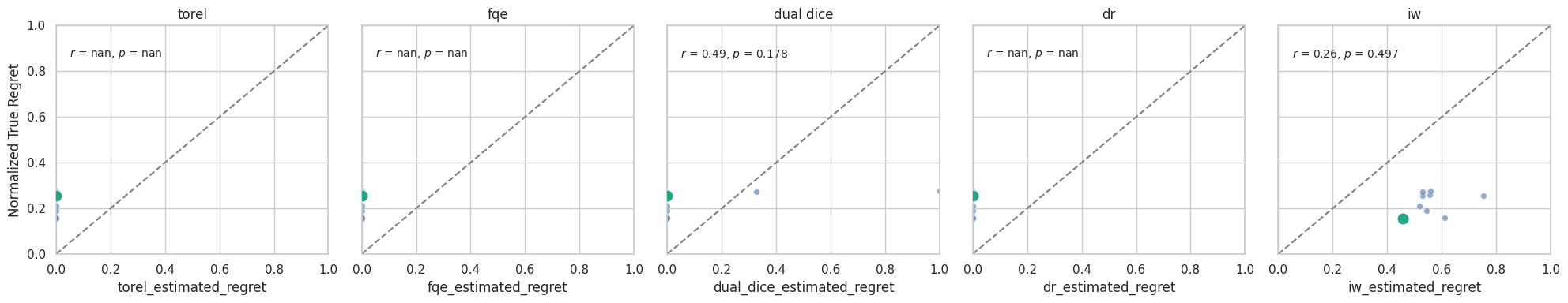}
    \end{subfigure}
    \begin{subfigure}{\linewidth}
        \includegraphics[width=\linewidth]{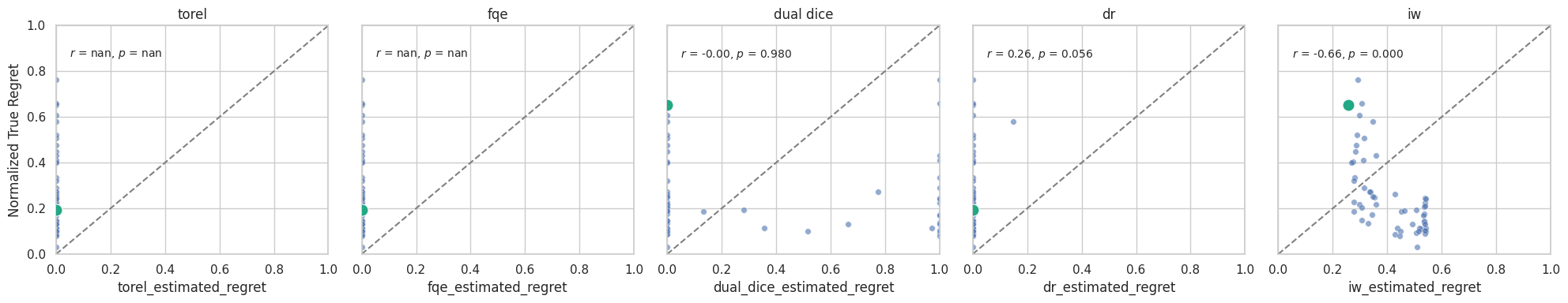}
        \caption{\textbf{IQL} (\textbf{top}) and \textbf{ReBRAC} (\textbf{bottom}), d4rl-pen-expert-v1}
    \end{subfigure}
    \begin{subfigure}{\linewidth}
        \includegraphics[width=\linewidth]{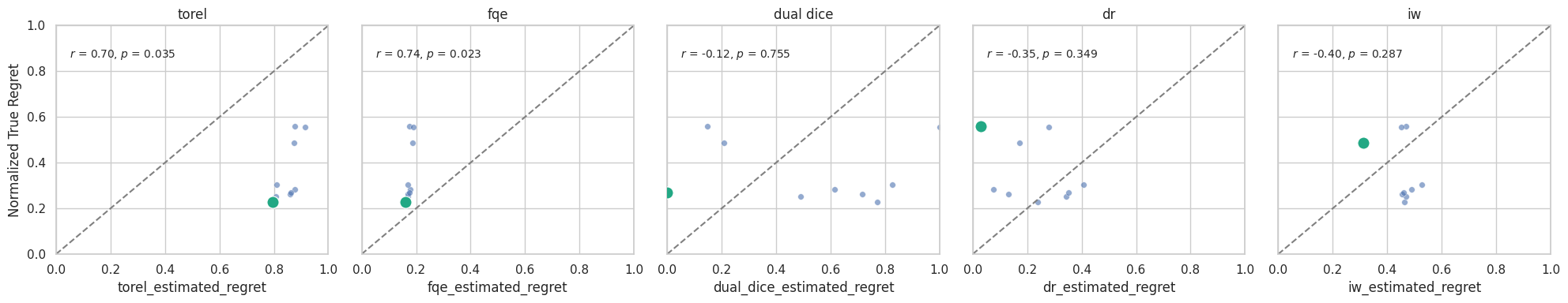}
    \end{subfigure}
    \begin{subfigure}{\linewidth}
        \includegraphics[width=\linewidth]{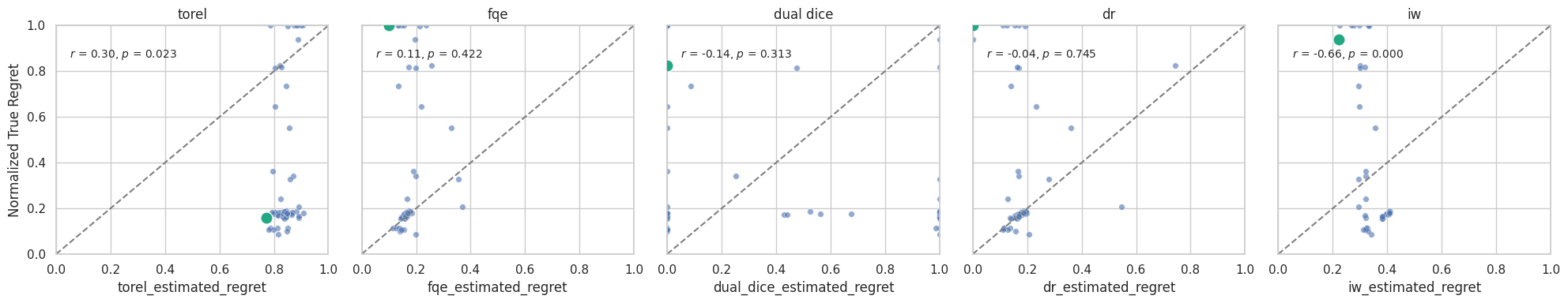}
        \caption{\textbf{IQL} (\textbf{top}) and \textbf{ReBRAC} (\textbf{bottom}), d4rl-hammer-expert-v1}
    \end{subfigure}
    \caption{Scatter plots (3) comparing different OPE algorithms' \textbf{offline estimated performance} and \textbf{online true performance} across datasets and algorithms.}
    \label{fig:ope_all_vertical_2}
\end{figure}

\clearpage

\section{Implementation Details}
Our implementations of SOReL and TOReL are publicly available.  
\label{app:implementation}
\subsection{Diverse full-replay datasets}
\label{app:sorel_datasets}
\begin{figure}[H]
    \centering
    \includegraphics[width=\linewidth]{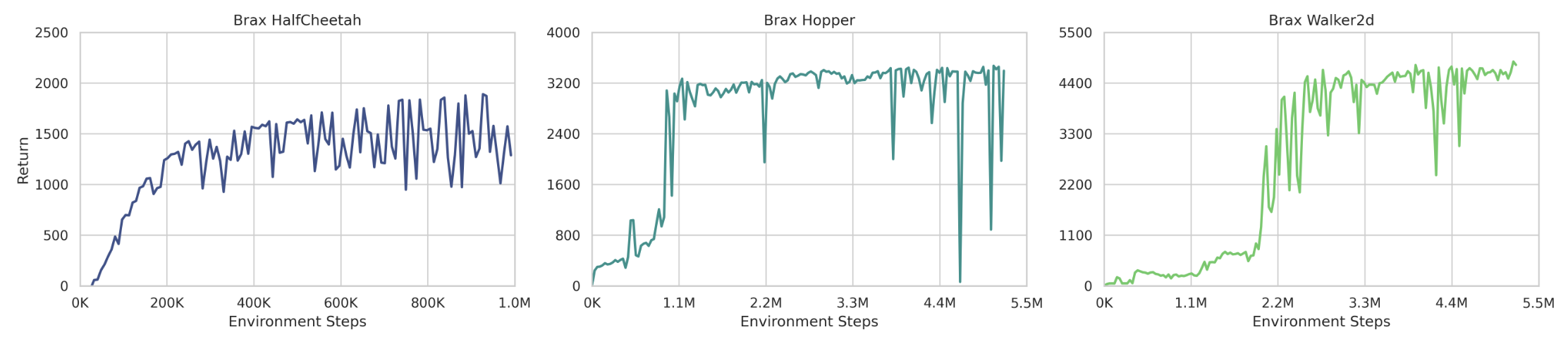}
    \caption{Training curves while \textbf{collecting the brax-full-replay} offline datasets. We ensure that the agent spend sufficient time in poor, medium and expert regions of performance such that the offline dataset captures diverse transitions.}
    \label{fig:brax_datasets}
\end{figure}

Without a model prior, the offline dataset must include transitions from poor, medium and expert regions of performance. For the brax environments, we collect our own full-replay datasets to ensure that this is the case. We arbitrarily choose the hyperparameters, simply requiring that the agent spends sufficient time in all three regions of performance. The training curves obtained while collecting the offline datasets are given in Figure \ref{fig:brax_datasets}. The Gymnax environments are simple enough that collecting a dataset using an ensemble of randomly initialised policies leads to sufficient coverage across all three regions of performance. 

\subsection{World Model and RP Approximate Inference}\label{app:world_model_implementation}
To ensure compatibility with Unifloral implementations \citep{unifloral2025}, our world model is a variation of the Gaussian World Model presented in \ref{sec:gauss_world}, but amended to predict the change in state $\Delta \coloneq s'-s$ rather than the absolute next state $s'$.  

When rolling out sequences of trajectories on which to train our (Bayes-optimal) policy, we uniformly sample a model from the ensemble of elite models and then sample the transition from the corresponding Gaussian output distribution.

Our ensemble consists of multilayer perceptrons (MLPs) with ReLU activation, which we train using negative log-likelihood loss derived in \cref{app:log_likelihood_derivation}. Training the models in parallel allows us to simultaneously optimise maximum and minimum (log) variance parameters for each dimension across the model ensemble, which we use to soft-clamp the (log) variances output by the individual models. This prevents any individual model becoming overly confident or too uncertain in one dimension. All models in our ensemble have identical structure, but are initialised differently. The maximum and minimum log-variance terms are initialised at constants. The exact loss function and ensemble dynamics model are the same as the one implemented by~\citet{unifloral2025}, but we use an Adam optimiser \citep{kingma2014adam} with cosine learning rate schedule rather than constant learning rate.
A percentage of the available offline dataset is used as a validation set to calculate the PIL.
At the end of training, only a subset of elite models are retained, based on their validation MSE.
Although the current implementation uses hard-coded reset and termination conditions during model rollouts, the dynamics model could naturally be extended to learn reset and termination heads.
When sampling transitions, we conservatively clip the rewards to remain within the support of the offline dataset distribution.
\begin{table}[htbp]
\centering
\begin{tabular}{lccccc}
\toprule
\textbf{Hyperparameter} &
\textbf{Gymnax} &
\textbf{Brax/D4RL}\\
\midrule
Num. layers & 3 & 3\\
Layer Size & 200 & 200\\
Activation & ReLU & ReLU\\
Num. Ensemble Models & 7 & 10\\
Num. Elite Models & 5 & 8\\
Log Var. Diff. Coeff. & 0.01 & 0.01\\
Batch Size & 64 & 256\\
Num. Epochs & 400 & 400\\
Learning Rate & 0.001 & 0.001\\
Learning Rate Schedule & cosine & cosine \\
Final Learning Rate \% & 10 & 10\\
Weight Decay & 2.5e-05 & 2.5e-05\\
Validation Split & 0.1 & 0.1\\
\bottomrule
\end{tabular}
\vspace{0.75em}
\caption{World Model Ensemble Dynamics Hyperparameters.}
\label{tab:ensemble_dynamics_hyperparams}
\end{table}

\newpage
\subsection{BAMDP Solver}\label{bamdp_implementation}
We use the RNN-PPO implementation of~\citet{lu2022discovered}, which we amend to be compatible with continuous action spaces. We sweep over the hyperparameters given in \cref{tab:sorel_ppo_rnn_hyperparameter_sweep}.
\begin{table}[H]
\centering
\begin{tabular}{lc}
\toprule
\textbf{Hyperparameter} & \textbf{Value / Sweep Values} \\
\midrule
Learning rate & [0.0001, 0.0003] \\
Anneal learning rate & True\\
Number of environments & [4, 64, 128, 256, 512]\\
Steps per environment & [32, 64, 128]\\
Total timesteps & Set to 500,000, 1,000,000 or 50,000,000 \\
Update epochs & [2, 4, 8]\\
Number of minibatches & [2, 4, 8, 16]\\
Discount factor ($\gamma$) & [0.99, 0.995, 0.998]\\
GAE lambda & [0.8, 0.9, 0.95]\\
Clip $\epsilon$ & [0.2, 0.3]\\
Entropy coefficient & [0.000, 0.001, 0.010]\\
Value function coefficient & 0.5\\
Max gradient norm & [0.5, 1.0]\\
Layer Size & 256 \\
Activation function & tanh \\
RNN size & Set to 64, 128 or 256\\
Burn-in Percentage & 25\\
\bottomrule
\end{tabular}
\vspace{0.75em}
\caption{RNN-PPO hyperparameters swept over.}
\label{tab:sorel_ppo_rnn_hyperparameter_sweep}
\end{table}
\begin{table}[H]
\centering
\begin{tabular}{lccccc}
\toprule
\textbf{Hyperparameter} &
\makecell[c]{\textbf{gymnax-}\\\textbf{cartpole-}\\\textbf{random-v1}} &
\makecell[c]{\textbf{gymnax-}\\\textbf{pendulum-}\\\textbf{random-v1}} &
\makecell[c]{\textbf{brax-}\\\textbf{halfcheetah-}\\\textbf{full-replay}} &
\makecell[c]{\textbf{brax-}\\\textbf{hopper-}\\\textbf{full-replay}} &
\makecell[c]{\textbf{brax-}\\\textbf{walker2d-}\\\textbf{full-replay}} \\
\midrule
Learning rate & 0.0003 & 0.0003 & 0.0003 & 0.0003 & 0.0003\\
Anneal learning rate & True & True & True & True & True\\
Number of environments & 4 & 128 & 512 & 512 & 512 \\
Steps per environment & 128 & 64 & 64 & 32 & 64\\
Total timesteps & 500,000 & 1,000,000 & 50,000,000 & 50,000,000 & 50,000,000\\
Update epochs & 4 & 8 & 8 & 2 & 4\\
Number of minibatches & 4 & 16 & 16 & 8 & 8 \\
Gamma & 0.99 & 0.99 & 0.99 & 0.998 & 0.995\\
GAE lambda & 0.95 & 0.95 & 0.95 & 0.8 & 0.95\\
Clip $\epsilon$ & 0.2 & 0.2 & 0.2 & 0.3 & 0.2\\
Entropy coefficient & 0.01 & 0.003 & 0.003 & 0.001 & 0.001 \\
Value function coefficient & 0.5 & 0.5 & 0.5 & 0.5 & 0.5\\
Max gradient norm & 0.5 & 0.5 & 0.5 & 1.0 & 0.5\\
Layer Size & 256 & 256 & 256 & 256 & 256 \\
Activation function & tanh & tanh & tanh & tanh & tanh\\
RNN size & 64 & 128 & 256 & 256 & 256 \\
Burn-in Percentage & 25 & 25 & 25 & 25 & 25 \\
\bottomrule
\end{tabular}
\vspace{0.75em}
\caption{RNN-PPO hyperparameters for Gymnax and Brax environments. For computational reasons, we sweep over the hyperparameters of each task once, for a fixed dataset size (1000 datapoints for the Gymnax tasks and 200,000 datapoints for Brax tasks) and choose the hyperparameters corresponding to the lowest approximate regret, which we then use to train the policy of all other dataset sizes. Ideally, we would sweep over all hyperparameters for each dataset size.}
\label{tab:sorel_ppo_rnn_hyperparameters}
\end{table}
\newpage
\subsection{ORL Implementations}\label{app:orl_implementation}
We use~\citet{unifloral2025}'s implementations of the ORL algorithms. We use their default hyperparameters and sweep over their suggested hyperparameters, which we summarise in \cref{tab:unifloral_orl_hyperparameters}.
\begin{table}[H]
\centering
\begin{tabular}{llc}
\toprule
& \textbf{Hyperparameter} & \textbf{Value / Sweep Values} \\
\midrule
\multirow{2}{*}{\makecell{\textbf{Generic}\\ \textbf{Optimisation}}}
& Discount factor $\gamma$ & 0.99 \\
& Polyak averaging coefficient & 0.005 \\
\midrule
\multirow{5}{*}{\textbf{IQL}}
& Learning rate & 0.0003 \\
& Batch size & 256 \\
& Beta & [0.5, 3.0, 10.0]\\
& $\tau$ (expectile) & [0.5, 0.7, 0.9]\\
& Advantage clip & 100.0 \\
\midrule
\multirow{10}{*}{\textbf{ReBRAC}}
& Learning rate & 0.001 \\
& Batch size & 1024 \\
& Critic BC coefficient & [0, 0.0001, 0.0005, 0.001, 0.005, 0.01, 0.1]\\
& Actor BC coefficient & [0.0005, 0.001, 0.002, 0.003, 0.03, 0.1, 0.3, 1.0]\\
& Critic layer norm & true \\
& Actor layer norm & false \\
& Observation normalization & false\\
& Noise clip & 0.5 \\
& Policy noise & 0.2 \\
& Num critic updates per step & 2 \\
\midrule
\multirow{9}{*}{\textbf{MOPO}}
& Learning rate & 0.0001 \\
& Batch size & 256 \\
& Model retain epochs & 5 \\
& Number of critics & 10 \\
& Rollout batch size & 50000 \\
& Rollout interval & 1000 \\
& Rollout length & [1, 3, 5]\\
& Dataset sample ratio & 0.05\\
& Step penalty coefficient & [1.0, 5.0]\\
\midrule
\multirow{10}{*}{\textbf{MOReL}}
& Learning rate & 0.0001 \\
& Batch size & 256 \\
& Model retain epochs & 5 \\
& Number of critics & 10 \\
& Rollout batch size & 50000 \\
& Rollout interval & 1000 \\
& Rollout length & 5 \\
& Dataset sample ratio & 0.01 \\
& Threshold coefficient & [0, 5, 10, 15, 20, 25] \\
& Termination penalty offset & [-30, -50, -100, -200] \\
\bottomrule
\end{tabular}
\vspace{0.75em}
\caption{Hyperparameters and sweep ranges for IQL, ReBRAC, MOPO, and MOReL.}
\label{tab:unifloral_orl_hyperparameters}
\end{table}
For the sample efficiency experiments in \cref{fig:d4rl_torel_regret_sample_efficiency}, we use the default UCB bandit-based hyperparameter-tuning algorithm hyperparameters.

\subsection{Experiment Compute Resources}
All of our experiments were run within a week using  four L40S GPUs. 

\end{document}